\documentclass[10pt,journal,cspaper,compsoc]{IEEEtran}

\usepackage{graphicx}
\usepackage{caption}
\usepackage{subcaption}

\usepackage{amsmath}
\usepackage{amssymb}
\usepackage{epsfig}

\usepackage{algpseudocode}
\usepackage{algorithm}

\usepackage{times}
\usepackage{epsfig}
\usepackage{graphicx}
\usepackage{tikz}
\usepackage{mathrsfs}

\usepackage{mathtools}
\usepackage{amsmath}
\usepackage[top=0.5in, left=.5in, bottom=.5in, right=.5in]{geometry}

\usepackage{slashbox}

\renewcommand{\a}{\boldsymbol{a}}
\newcommand{\bdelta}{\boldsymbol{\delta}}

\newcommand{\A}{\boldsymbol{A}}
\newcommand{\B}{\boldsymbol{B}}
\newcommand{\q}{\boldsymbol{q}}
\newcommand{\bd}{\bar{\boldsymbol{d}}}
\newcommand{\bz}{\bar{\boldsymbol{z}}}
\renewcommand{\d}{\boldsymbol{d}}
\newcommand{\C}{\boldsymbol{C}}

\newcommand{\G}{\mathcal{G}}
\newcommand{\D}{\boldsymbol{D}}

\renewcommand{\v}{\boldsymbol{v}}
\renewcommand{\u}{\boldsymbol{u}}
\newcommand{\e}{\boldsymbol{e}}
\newcommand{\g}{\boldsymbol{g}}
\newcommand{\h}{\boldsymbol{h}}

\newcommand{\z}{\boldsymbol{z}}
\newcommand{\w}{\boldsymbol{w}}
\newcommand{\x}{\boldsymbol{x}}
\newcommand{\y}{\boldsymbol{y}}
\renewcommand{\r}{\boldsymbol{r}}
\renewcommand{\L}{\mathcal{L}}

\newcommand{\Z}{\boldsymbol{Z}}
\newcommand{\X}{\mathbb{X}}
\newcommand{\Y}{\mathbb{Y}}
\renewcommand{\Re}{\mathbb{R}}

\newcommand{\bLambda}{\boldsymbol{\Lambda}}

\newcommand{\btheta}{\boldsymbol{\theta}}

\newcommand{\I}{\boldsymbol{I}}
\newcommand{\Io}{\operatorname{I}}
\newcommand{\0}{\boldsymbol{0}}
\newcommand{\1}{\boldsymbol{1}}
\newcommand{\hbbeta}{\hat{\boldsymbol{\beta}}}
\newcommand{\bbeta}{\boldsymbol{\beta}}

\newcommand{\bvarep}{\boldsymbol{\varepsilon}}

\newcommand{\st}{\operatorname{s.t.}}

\newcommand{\tr}{\operatorname{tr}}
\newcommand{\argmin}{\operatorname{argmin}}

\newcommand{\ie}{\text{i.e.}}
\newcommand{\eg}{\text{e.g.}}

\newtheorem{theorem}{Theorem}

\newtheorem{lemma}{Lemma}
\newtheorem{corollary}{Corollary}
\newtheorem{definition}{Definition}
\newtheorem{remark}{Remark}
\newtheorem{example}{Example}
\newtheorem{assumption}{Assumption}

\def\QED{~\rule[-1pt]{5pt}{5pt}\par\medskip}
\newenvironment{proof}{\emph{Proof.}}{\hfill\QED}

\ifCLASSOPTIONcompsoc
\else
\fi
\ifCLASSINFOpdf
\else
\fi

\begin{document}
%
\title{Dissimilarity-based Sparse Subset Selection}
%
%
%
%

\author{Ehsan~Elhamifar,~\IEEEmembership{Member,~IEEE,}
        Guillermo Sapiro,~\IEEEmembership{Fellow,~IEEE,}
        \\and~S.~Shankar~Sastry,~\IEEEmembership{Fellow,~IEEE}
\IEEEcompsocitemizethanks{\IEEEcompsocthanksitem E. Elhamifar is with the College of Computer and Information Science and the Department of Electrical and Computer Engineering, Northeastern University, USA. E-mail: eelhami@ccs.neu.edu.
\IEEEcompsocthanksitem G. Sapiro is with the Department
of Electrical and Computer Engineering, Duke University, USA. E-mail: guillermo.sapiro@duke.edu.
\IEEEcompsocthanksitem S. Shankar Sastry is with the Department
of Electrical Engineering and Computer Sciences, UC Berkeley, USA. E-mail: sastry@eecs.berkeley.edu.}
\thanks{}}

\IEEEcompsoctitleabstractindextext{%
\begin{abstract}
Finding an informative subset of a large collection of data points or models is at the center of many problems in computer vision, recommender systems, bio/health informatics as well as image and natural language processing. Given pairwise dissimilarities between the elements of a `source set' and a `target set,' we consider the problem of finding a subset of the source set, called \emph{representatives} or \emph{exemplars}, that can efficiently describe the target set. We formulate the problem as a row-sparsity regularized trace minimization problem. Since the proposed formulation is, in general, NP-hard, we consider a convex relaxation. The solution of our optimization finds representatives and the assignment of each element of the target set to each representative, hence, obtaining a clustering. We analyze the solution of our proposed optimization as a function of the regularization parameter. We show that when the two sets jointly partition into multiple groups, our algorithm finds representatives from all groups and reveals clustering of the sets. In addition, we show that the proposed framework can effectively deal with outliers. Our algorithm works with arbitrary dissimilarities, which can be asymmetric or violate the triangle inequality. To efficiently implement our algorithm, we consider an Alternating Direction Method of Multipliers (ADMM) framework, which results in quadratic complexity in the problem size. We show that the ADMM implementation allows to parallelize the algorithm, hence further reducing the computational time. Finally, by experiments on real-world datasets, we show that our proposed algorithm improves the state of the art on the two problems of scene categorization using representative images and time-series modeling and segmentation using representative~models.

\end{abstract}

}

\maketitle

\IEEEdisplaynotcompsoctitleabstractindextext

%
\IEEEpeerreviewmaketitle

\section{Introduction}
\begin{figure*}[t!]
\centering
\includegraphics[width=0.114\linewidth, trim = 75 75 75 43, clip]{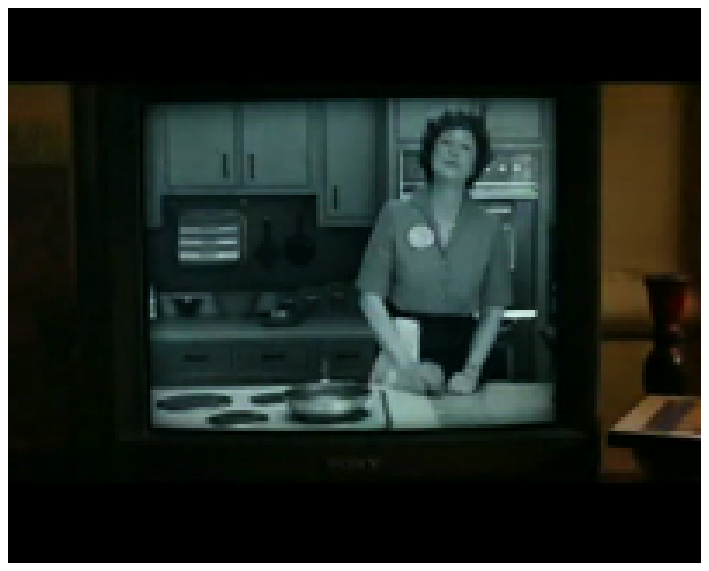}\
\includegraphics[width=0.114\linewidth, trim = 75 75 75 43, clip]{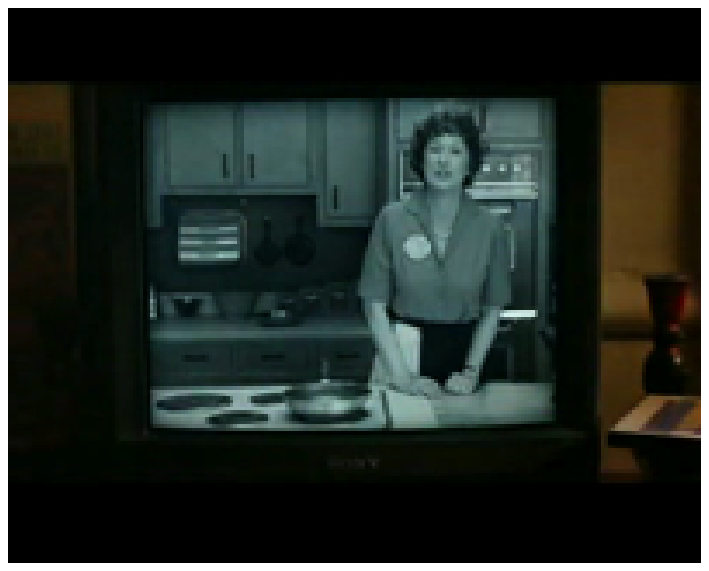}\
\colorbox{red}{\includegraphics[width=0.114\linewidth, trim = 75 75 75 43, clip]{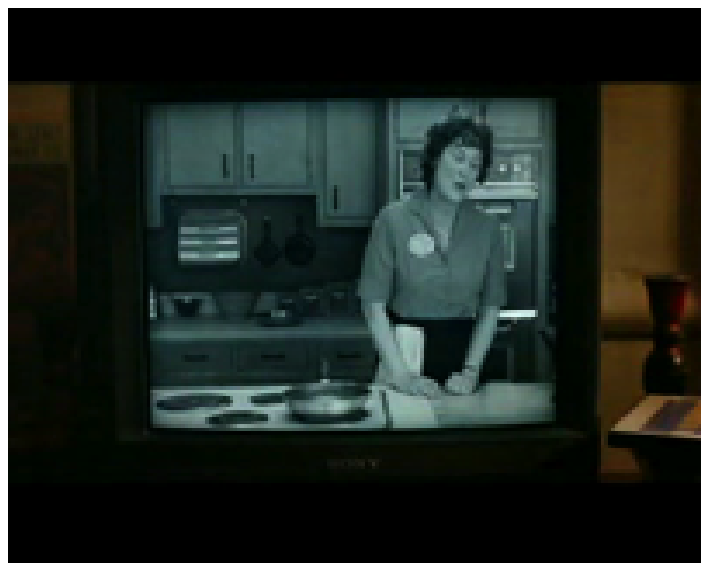}}\
\includegraphics[width=0.114\linewidth, trim = 75 75 75 43, clip]{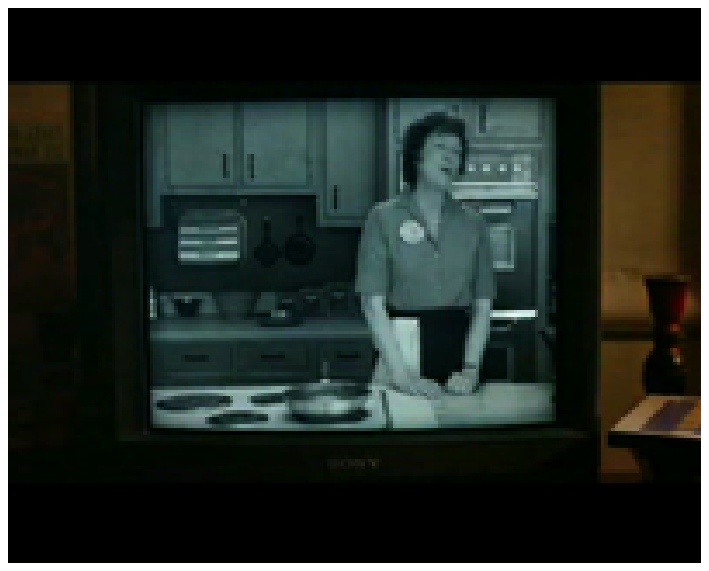}\
\includegraphics[width=0.114\linewidth, trim = 75 75 75 43, clip]{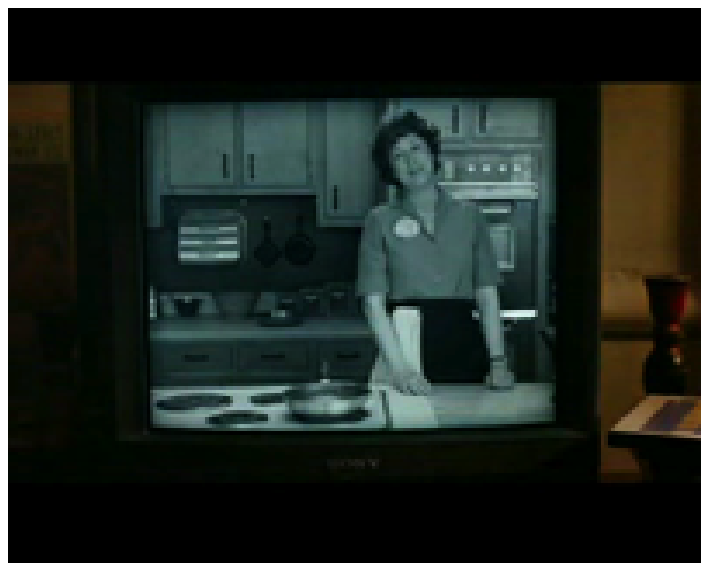}\
\includegraphics[width=0.114\linewidth, trim = 75 75 75 43, clip]{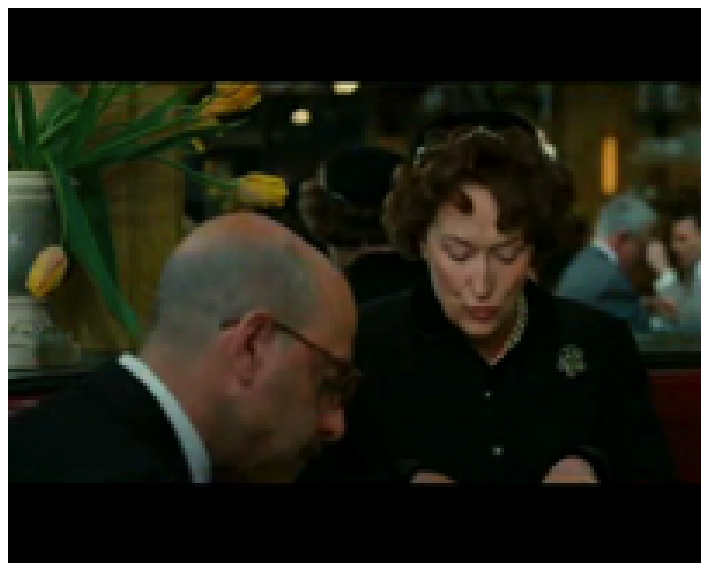}\
\colorbox{red}{\includegraphics[width=0.114\linewidth, trim = 75 75 75 43, clip]{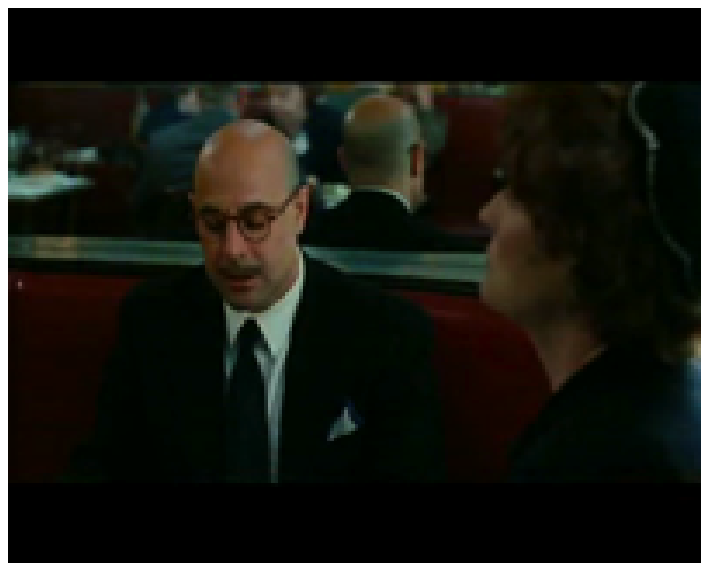}}\
\includegraphics[width=0.114\linewidth, trim = 75 75 75 43, clip]{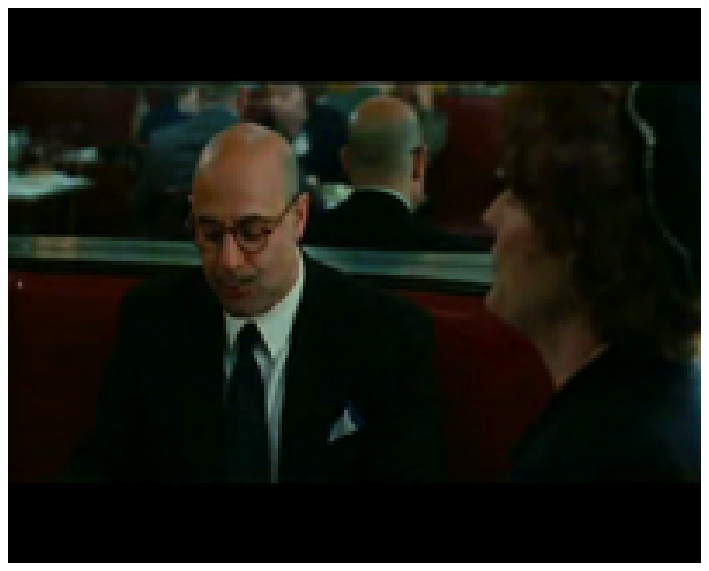}\
\includegraphics[width=0.114\linewidth, trim = 75 75 75 43, clip]{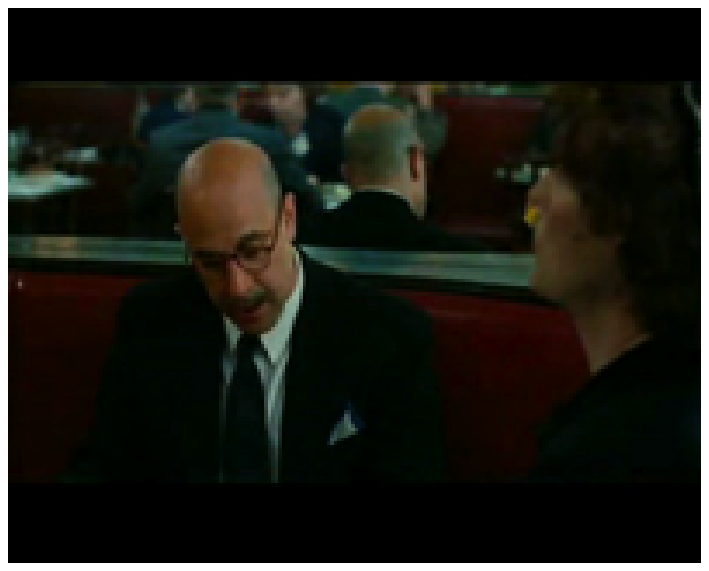}\
\includegraphics[width=0.114\linewidth, trim = 75 75 75 43, clip]{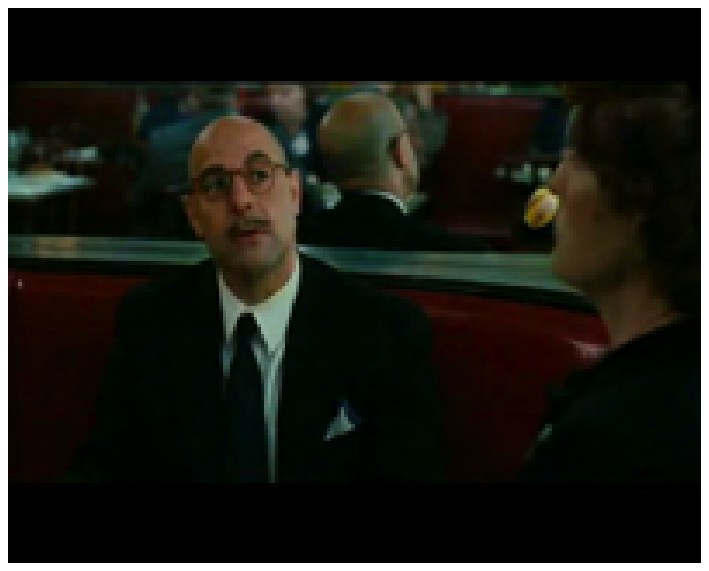}\
\colorbox{red}{\includegraphics[width=0.114\linewidth, trim = 75 75 75 43, clip]{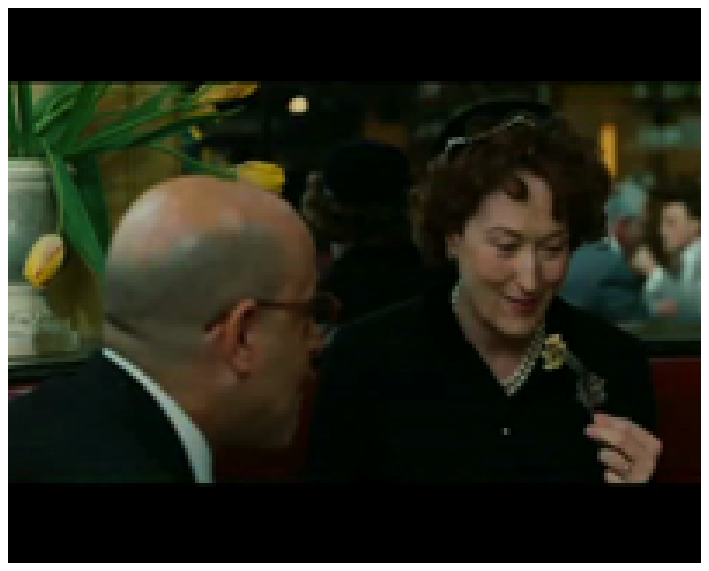}}\
\includegraphics[width=0.114\linewidth, trim = 75 75 75 43, clip]{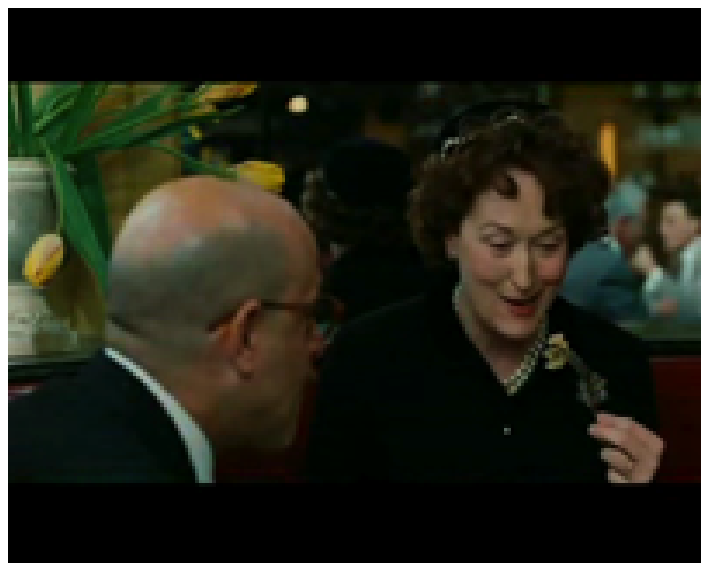}\
\includegraphics[width=0.114\linewidth, trim = 75 75 75 43, clip]{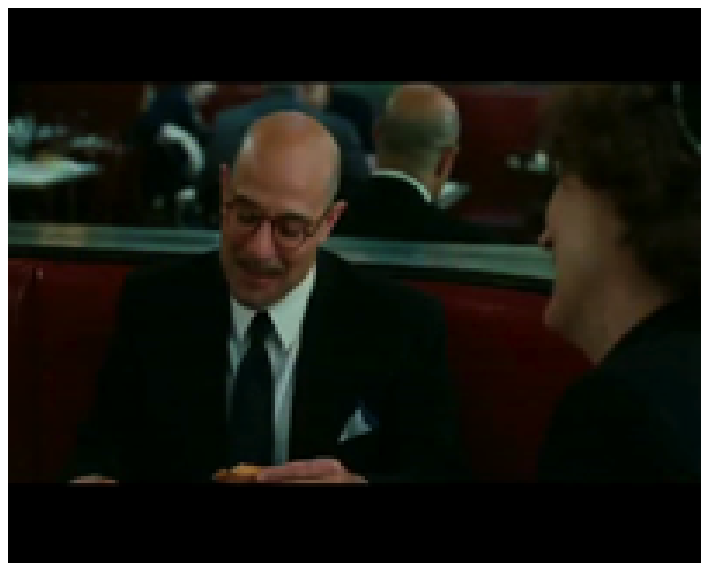}\
\includegraphics[width=0.114\linewidth, trim = 75 75 75 43, clip]{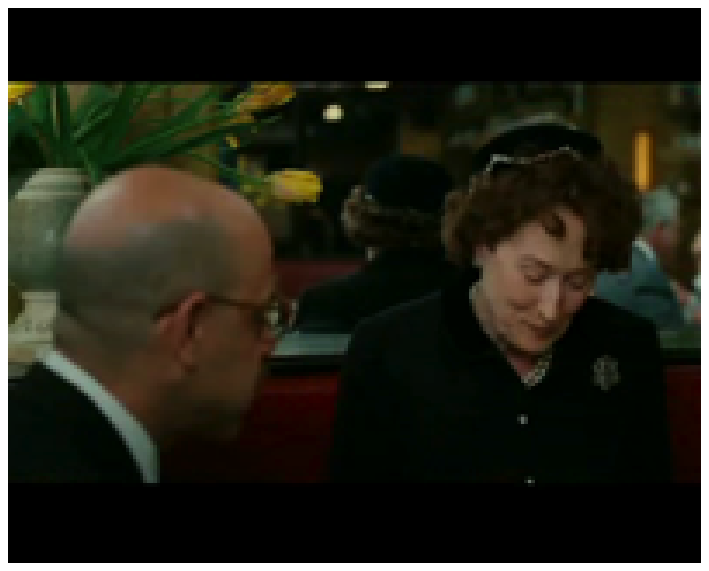}\
\includegraphics[width=0.114\linewidth, trim = 75 75 75 43, clip]{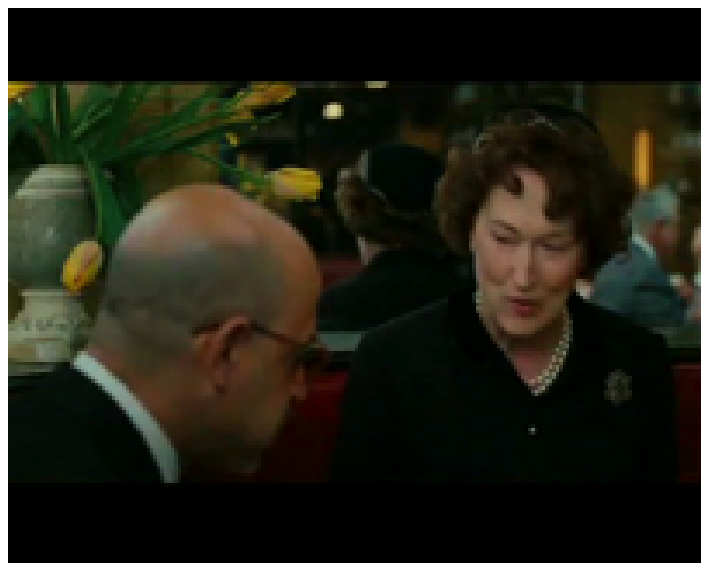}\
\includegraphics[width=0.114\linewidth, trim = 75 75 75 43, clip]{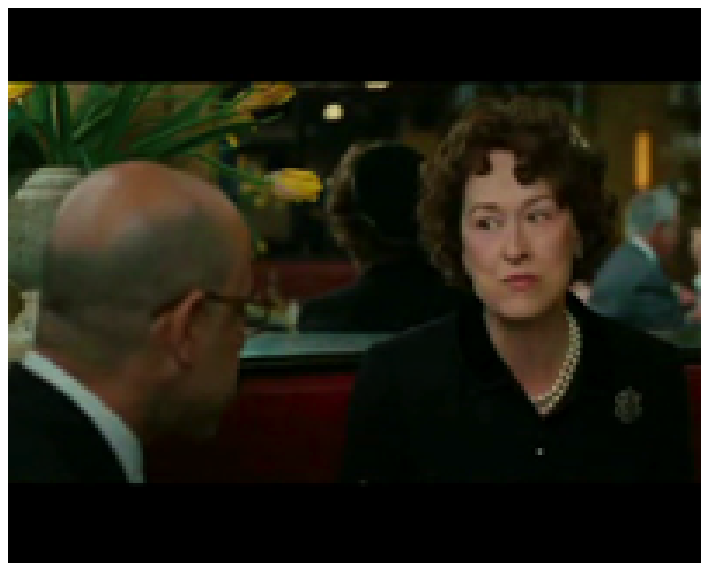}\
\colorbox{red}{\includegraphics[width=0.114\linewidth, trim = 75 75 75 43, clip]{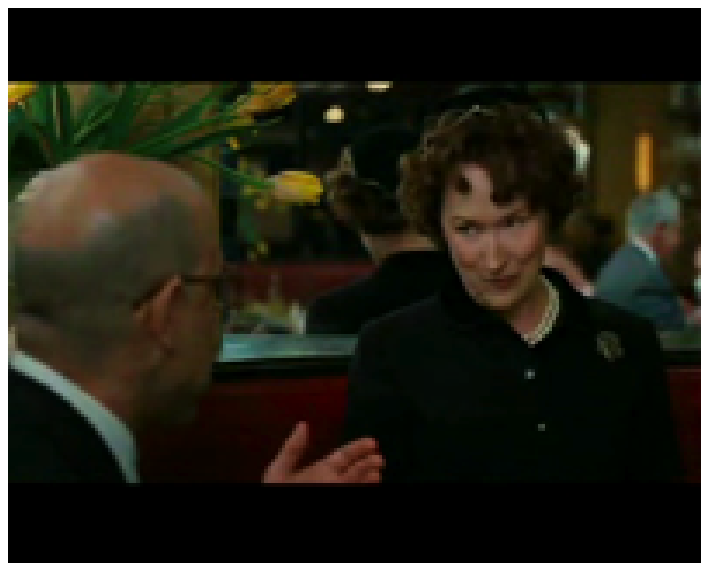}}\
\includegraphics[width=0.114\linewidth, trim = 75 75 75 43, clip]{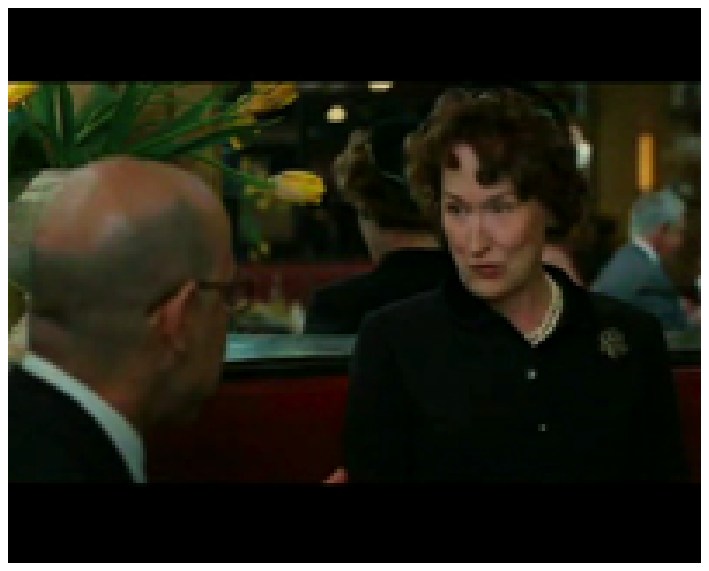}\
\includegraphics[width=0.114\linewidth, trim = 75 75 75 43, clip]{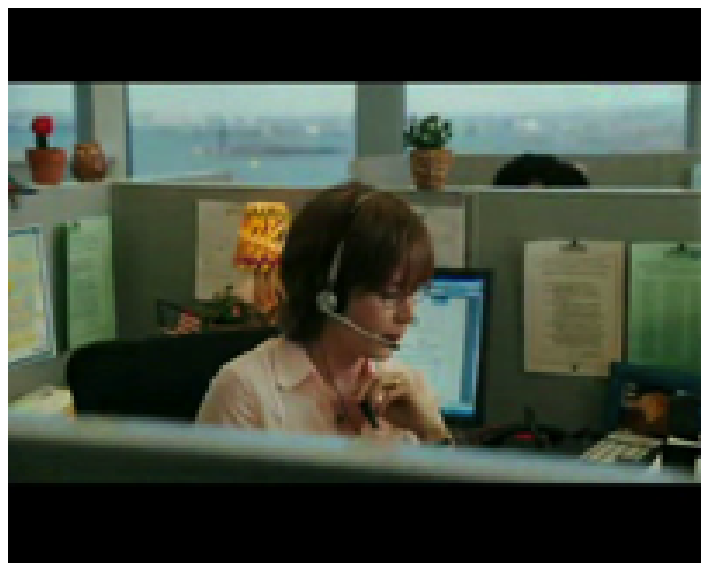}\
\colorbox{red}{\includegraphics[width=0.114\linewidth, trim = 75 75 75 43, clip]{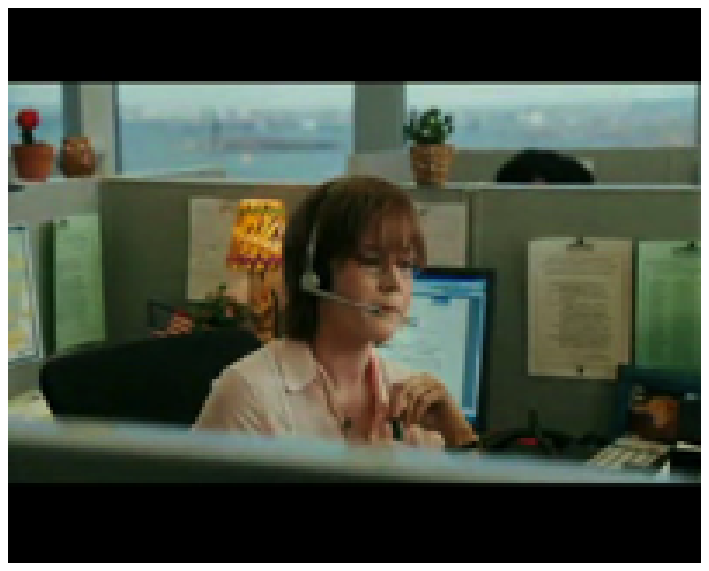}}\
\includegraphics[width=0.114\linewidth, trim = 75 75 75 43, clip]{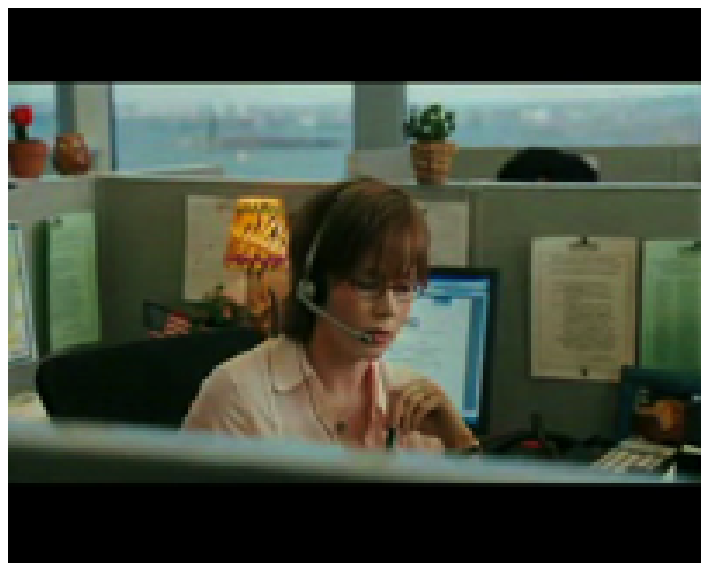}\
\includegraphics[width=0.114\linewidth, trim = 75 75 75 43, clip]{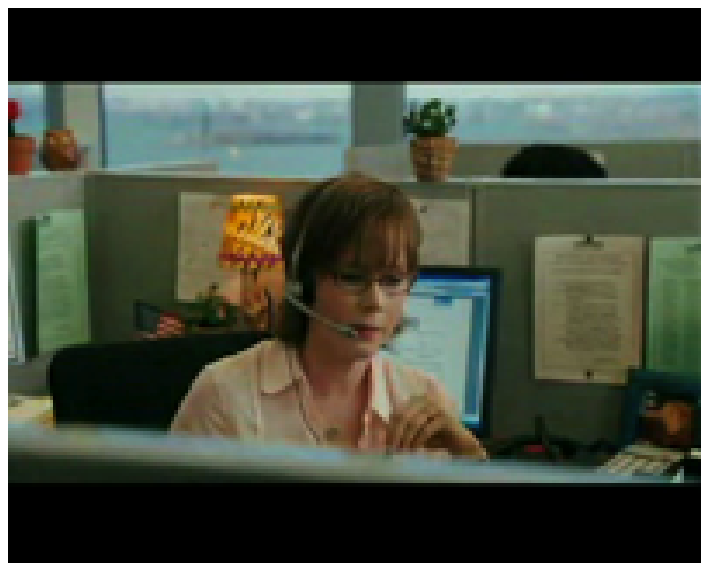}\
\includegraphics[width=0.114\linewidth, trim = 75 75 75 43, clip]{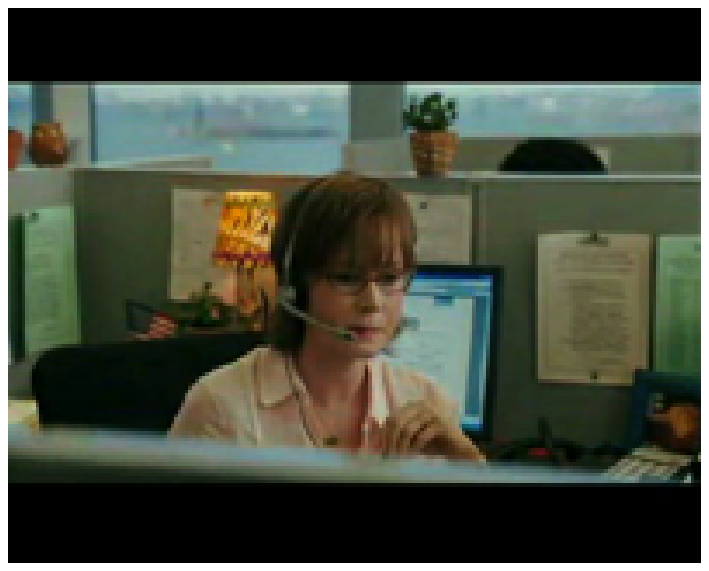}\
\colorbox{red}{\includegraphics[width=0.114\linewidth, trim = 75 75 75 43, clip]{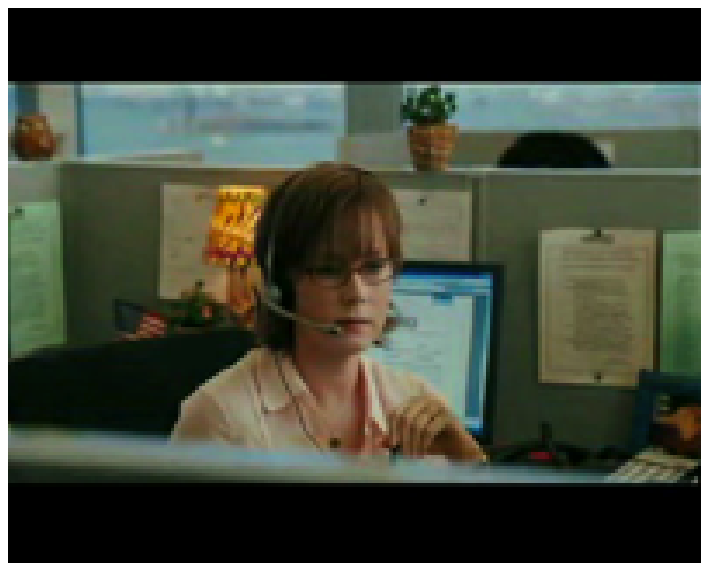}}\ 
\includegraphics[width=0.114\linewidth, trim = 75 75 75 43, clip]{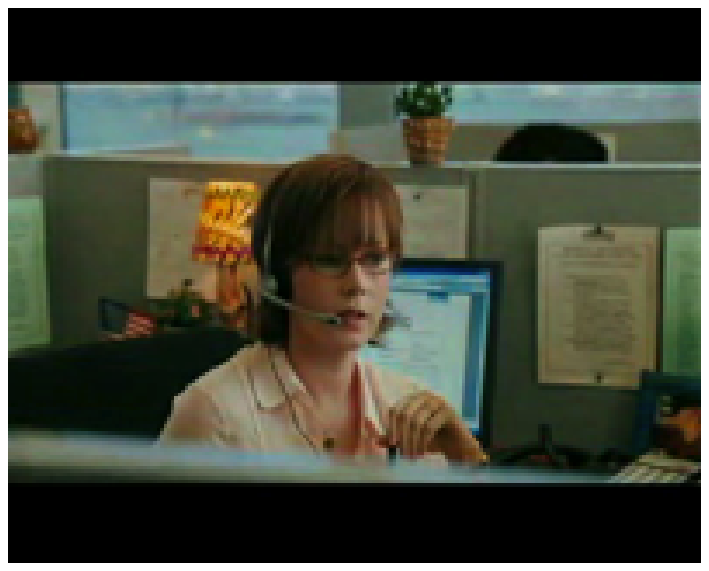}\
\includegraphics[width=0.114\linewidth, trim = 75 75 75 43, clip]{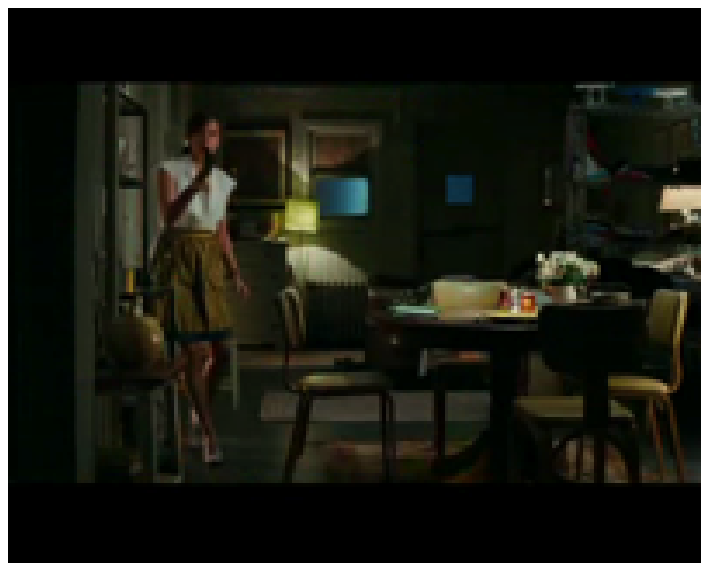}\
\includegraphics[width=0.114\linewidth, trim = 75 75 75 43, clip]{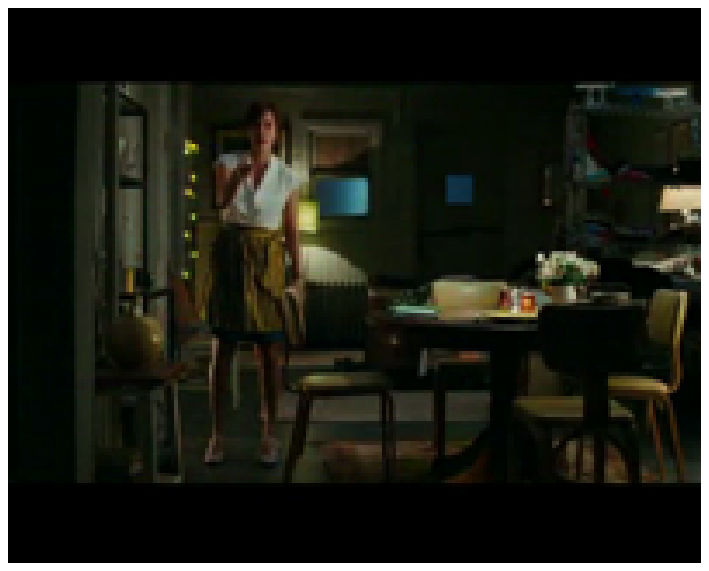}\
\colorbox{red}{\includegraphics[width=0.114\linewidth, trim = 75 75 75 43, clip]{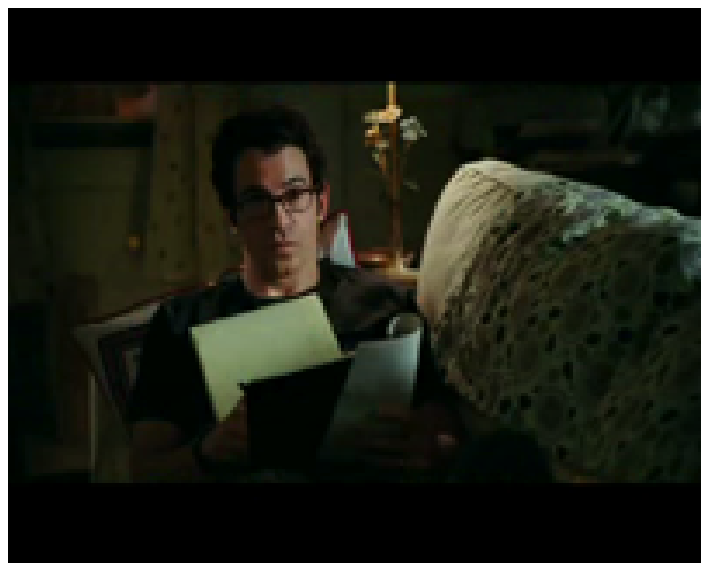}}\
\colorbox{red}{\includegraphics[width=0.114\linewidth, trim = 75 75 75 43, clip]{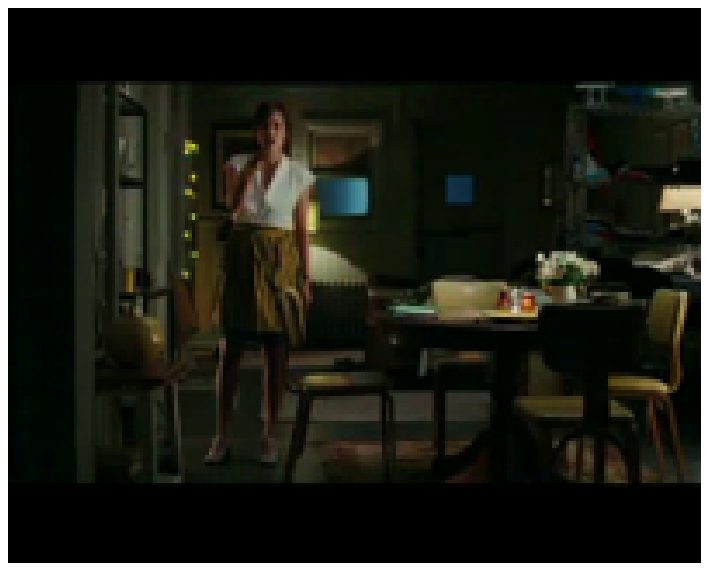}}\
\includegraphics[width=0.114\linewidth, trim = 75 75 75 43, clip]{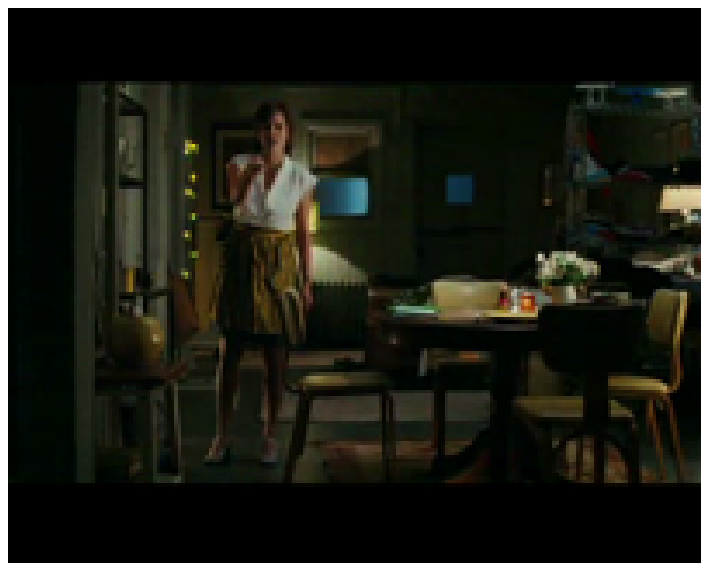}\
\includegraphics[width=0.114\linewidth, trim = 75 75 75 43, clip]{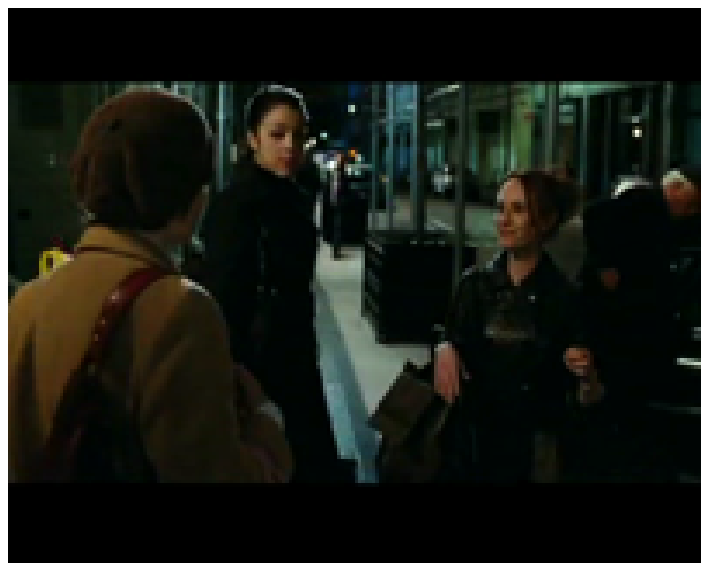}\
\includegraphics[width=0.114\linewidth, trim = 75 75 75 43, clip]{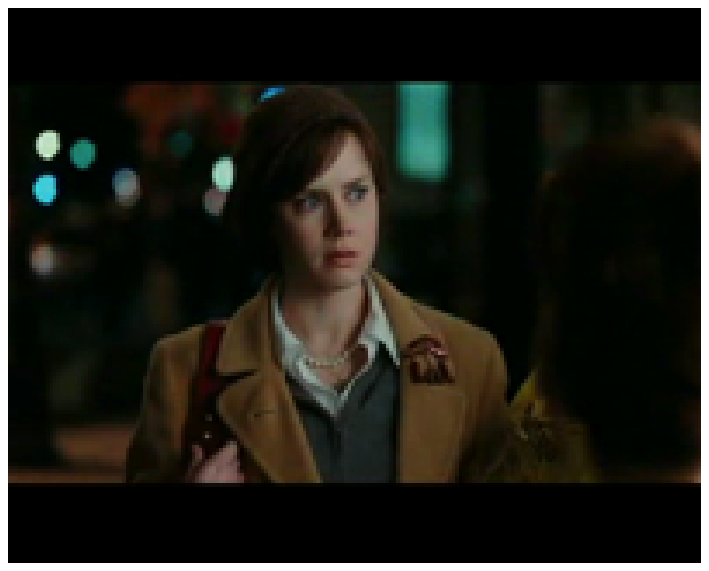}\
\colorbox{red}{\includegraphics[width=0.114\linewidth, trim = 75 75 75 43, clip]{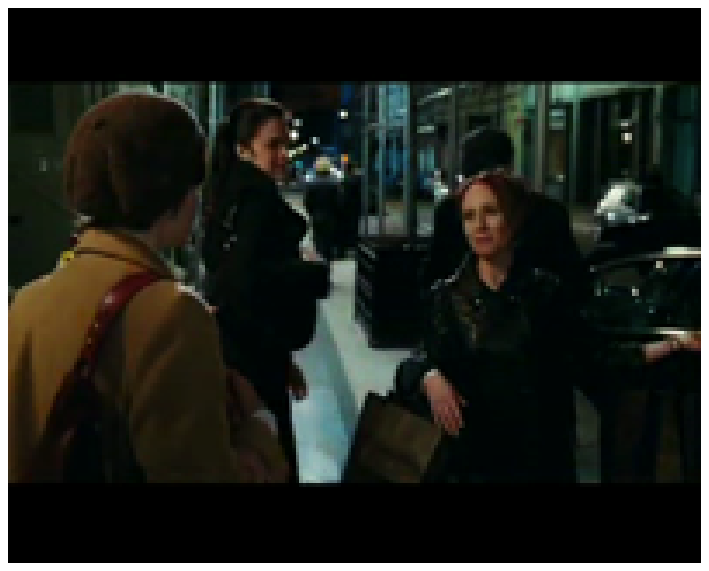}}\
\includegraphics[width=0.114\linewidth, trim = 75 75 75 43, clip]{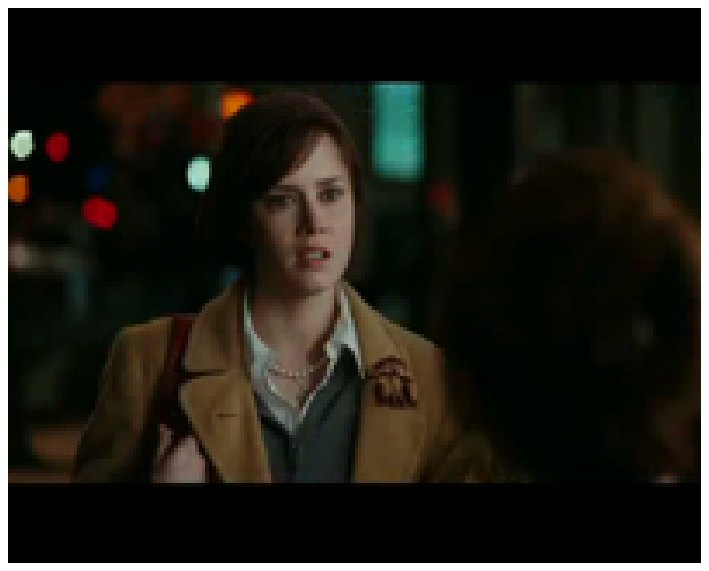}\
\colorbox{red}{\includegraphics[width=0.114\linewidth, trim = 75 75 75 43, clip]{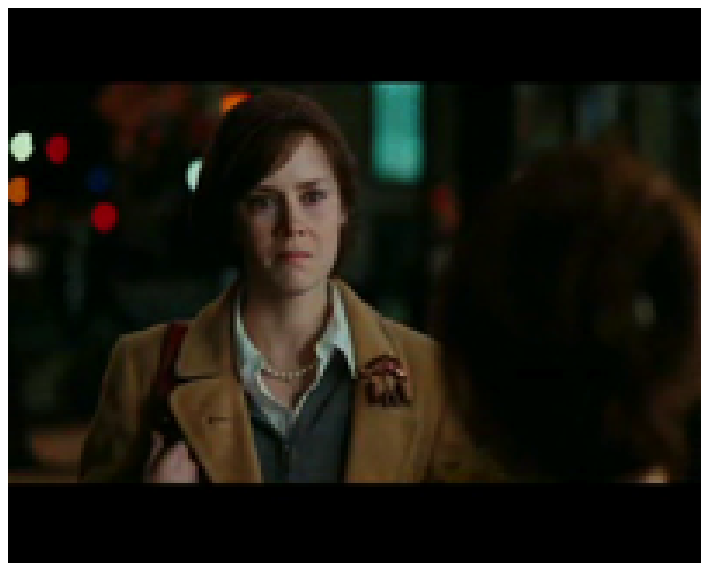}}\
\includegraphics[width=0.114\linewidth, trim = 75 75 75 43, clip]{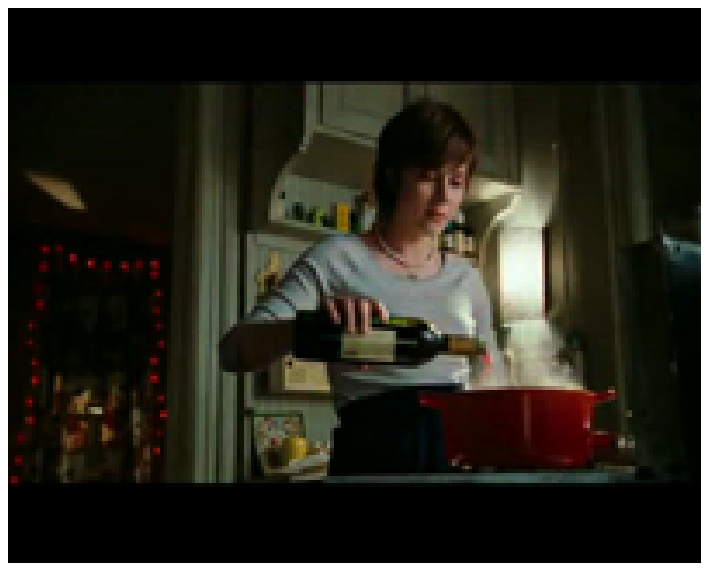}\
\includegraphics[width=0.114\linewidth, trim = 75 75 75 43, clip]{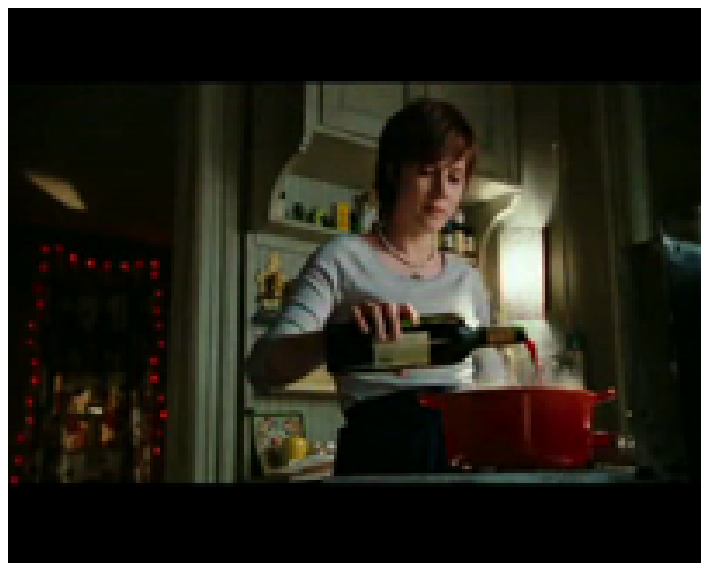}\
\colorbox{red}{\includegraphics[width=0.114\linewidth, trim = 75 75 75 43, clip]{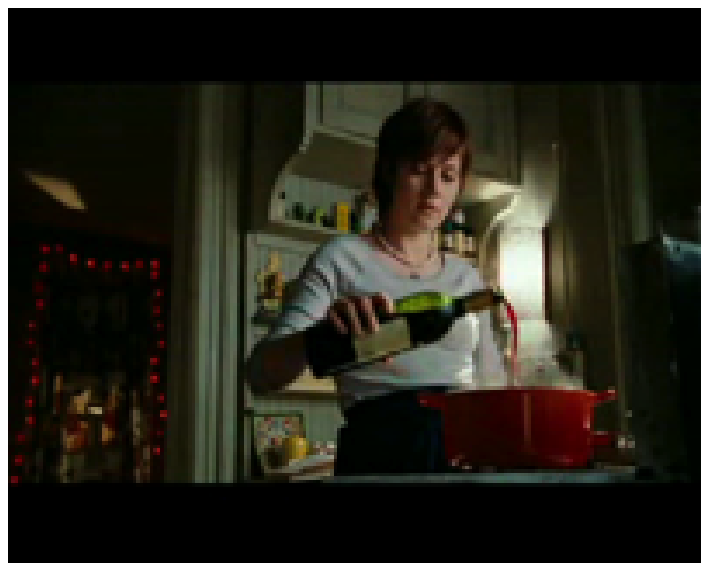}}
\includegraphics[width=0.114\linewidth, trim = 75 75 75 43, clip]{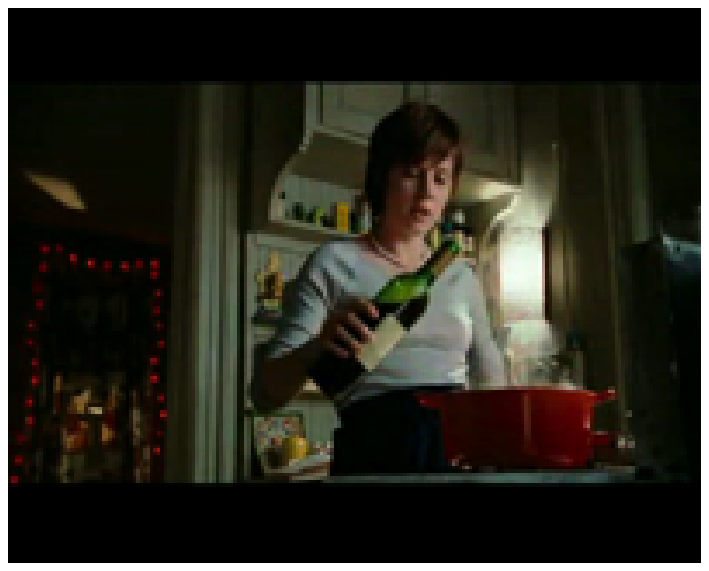}\
\includegraphics[width=0.114\linewidth, trim = 75 75 75 43, clip]{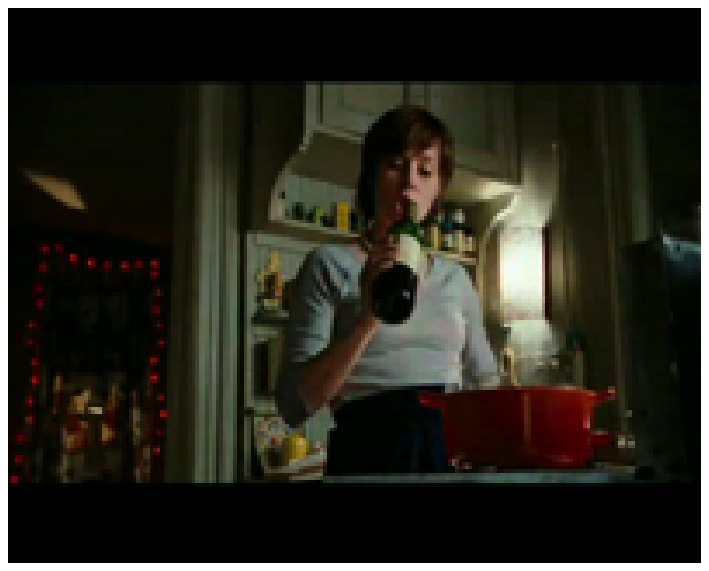}\
\vspace{0mm} 
\caption{\footnotesize{Video summarization using representatives: some video frames of a movie trailer, which consists of multiple shots, and the automatically computed representatives (inside red rectangles) of the whole video sequence using our proposed algorithm. We use the Bag of Features (BoF) approach \cite{Li:CVPR05} by extracting SIFT features \cite{Lowe:IJCV04} from all frames of the video and forming a histogram with $b=100$ bins for each frame. We apply the DS3 algorithm to the dissimilarity matrix computed by using the $\chi^2$ distance between pairs of histograms.}} 
\label{fig:videosummary}
\end{figure*}

\IEEEPARstart{F}{inding} a subset of a large number of models or data points, which preserves the characteristics of the entire set, is an important problem in machine learning and data analysis with applications in computer vision \cite{Simon:ICCV07, Elhamifar:CVPR12, Elhamifar:NIPS12, Taskar:ICML11, Yang:CVPR13, Sha:NIPS14}, image and natural language processing \cite{Bilmes:ARSU09, Esser:TIP12}, bio/health informatics \cite{Frey:Science07, Bien:AAStats11}, recommender systems \cite{Gillenwater:NIPS14, Hartline:WWW08} and more \cite{Elhamifar:IFAC14, Mahoney:NAS09, Garcia:PAMI12, Elhamifar:ICCV13}. Such informative elements are referred to as \emph{representatives} or \emph{exemplars}. Data representatives help to summarize and visualize datasets of text/web documents, images and videos (see Figure \ref{fig:videosummary}), hence, increase the interpretability of large-scale datasets for data analysts and domain experts \cite{Simon:ICCV07, Elhamifar:CVPR12, Elhamifar:NIPS12, Frey:Science07, Grauman:CVPR13}. {Model representatives help to efficiently describe complex phenomena or events using a small number of models or can be used for model compression in ensemble models \cite{Elhamifar:IFAC14, Hebert:WACV14}.} More importantly, the computational time and memory requirements of learning and inference algorithms, such as the Nearest Neighbor (NN) classifier, improve by working on representatives, which contain much of the information of the original set \cite{Garcia:PAMI12}. Selecting a good subset of products to recommend to costumers helps to not only boost revenue of retailers, but also save customer time \cite{Gillenwater:NIPS14, Hartline:WWW08}. 
Moreover, representatives help in clustering of datasets, and, as the most prototypical elements, can be used for efficient synthesis/generation of new data points. Last but not least, representatives can be used to obtain high performance classifiers using very few samples selected and annotated from a large pool of unlabeled samples \cite{Elhamifar:ICCV13, Grauman:ECCV12}.

%

\subsection{Prior Work on Subset Selection}

The problem of finding data representatives has been well-studied in the literature \cite{Elhamifar:CVPR12, Taskar:ICML11, Frey:Science07, Mahoney:NAS09, Kaufman:TR87, Gu:SIAM96, Tropp:SODA09, Boutsidis:SODA09, Golland:NIPS07}. Depending on the type of information that should be preserved by the representatives, algorithms can be divided into two categories. 

The first group of algorithms finds representatives of data that lie in one or multiple low-dimensional subspaces \cite{Elhamifar:CVPR12, Esser:TIP12, Mahoney:NAS09, Tropp:SODA09, Boutsidis:SODA09, Chan:LAA87, Balzano:WNIPS10}. Data in such cases are typically embedded in a vector space. The Rank Revealing QR (RRQR) algorithm assumes that the data come from a low-rank model and tries to find a subset of columns of the data matrix that corresponds to the best conditioned submatrix \cite{Chan:LAA87}. Randomized and greedy algorithms have also been proposed to find a subset of the columns of a low-rank matrix \cite{Tropp:SODA09, Boutsidis:SODA09, Balzano:WNIPS10, Bien:NIPS10}. CUR approximates a large data matrix by using a few of its rows and columns \cite{Mahoney:NAS09}. Assuming that the data can be expressed as a linear combination of the representatives, \cite{Elhamifar:CVPR12} and \cite{Esser:TIP12} formulate the problem of finding representatives as a joint-sparse recovery problem, \cite{Elhamifar:CVPR12} showing that when data lie in a union of low-rank models, the algorithm finds representatives from each model. While such methods work well for data lying in low-dimensional linear models, they cannot be applied to the more general case where data do not lie in subspaces, \eg, when data lie in nonlinear manifolds or do not live in a vector space.

The second group of algorithms uses similarities/dissimilarities between pairs of data points instead of measurement vectors \cite{Elhamifar:NIPS12, Frey:Science07, Kaufman:TR87, Charikar:JCSS02, Frey:NIPS06, Frey:UAI11}. Working on pairwise relationships has several advantages. First, for high-dimensional datasets, where the ambient space dimension is much higher than the cardinality of the dataset, working on pairwise relationships is more efficient than working on high-dimensional measurement vectors. Second, while some real datasets do not live in a vector space, \eg, social network data or proteomics data \cite{Bien:AAStats11}, pairwise relationships are already available or can be computed efficiently. More importantly, working on pairwise similarities/dissimilarities allows one to consider models beyond linear subspaces. However, existing algorithms suffer from dependence on the initialization, finding approximate solutions for the original problem, or imposing restrictions on the type of pairwise relationships.

The Kmedoids algorithm \cite{Kaufman:TR87} tries to find $K$ representatives from pairwise dissimilarities between data points. As solving the corresponding optimization program is, in general, NP-hard \cite{Charikar:JCSS02}, an iterative approach is employed. Therefore, the performance of Kmedoids, similar to Kmeans \cite{Duda:04}, depends on the initialization and decreases as the number of representatives, $K$, increases. The Affinity Propagation (AP) algorithm \cite{Frey:Science07, Frey:NIPS06, Frey:UAI11} tries to find representatives from pairwise similarities between data points by using an approximate message passing algorithm. While it has been shown empirically that AP performs well in problems such as unsupervised image categorization \cite{Frey:ICCV07}, there is no guarantee for AP to find the desired solution and, in addition, it works only with a single dataset. Determinantal Point Processes (DPPs) \cite{Gillenwater:NIPS14, Macchi:AAP75, Borodin:Arxiv09} and its fixed-size variant, kDPPs, \cite{Taskar:ICML11, Taskar:AISTATS13} find representatives by sampling from a probability distribution, defined on all subsets of the given set, using a positive semidefinite kernel matrix. While DPPs and kDPPs promote diversity among representatives, they cannot work with arbitrary similarities, only work with a single dataset and are computationally expensive in general, since they require to compute the eigen-decomposition of the kernel matrix. Using submodular selection methods, \cite{Bilmes:ARSU09, Bilmes:INTERSPEECH09, Krause:JMLR08} propose algorithms with approximate solutions for the problem of subset selection. Moreover, in the operations research literature, subset selection has been studied under the name of facility location problem for which, under the assumption of metric dissimilarities, algorithms with approximate solutions have been proposed \cite{Shmoys:TheoryComp12, Li:InfoComp12, Li:TheoryComp12}.

{Finally, it is important to note that while using dissimilarities has several advantages, a limitation of algorithms that require working with all pairwise dissimilarities \cite{Taskar:ICML11, Kaufman:TR87, Taskar:AISTATS13} is that they do not scale well, in general, in the size of datasets.}

\subsection{Paper Contributions}
{{In this paper, we consider the problem of finding representatives, given pairwise dissimilarities between the elements of a source set, $\X$, and a target set, $\Y$, in an unsupervised framework.}} In order to find \emph{a few representatives} of $\X$ that \emph{well encode} the collection of elements of $\Y$, we propose an optimization algorithm based on simultaneous sparse recovery \cite{Tropp:SP06, Jenatton:JMLR11}. We formulate the problem as a row-sparsity regularized trace minimization program, where the regularization parameter puts a trade-off between the number of representatives and the encoding cost of $\Y$ via representatives. The solution of our algorithm finds representatives and the probability that each element in the target set is associated with each representative. We also consider an alternative optimization, which is closely related to our original formulation, and establish relationships to Kmedoids.

Our proposed algorithm has several advantages with respect to the state of the art:

\smallskip\noindent\textbf{--} While AP \cite{Frey:Science07}, DPPs \cite{Macchi:AAP75} and kDPPs \cite{Taskar:ICML11} work with a single set, we consider the more general setting of having dissimilarities between two different sets. This is particularly important when computing pairwise dissimilarities in a given set is difficult while dissimilarities to a different set can be constructed efficiently. For instance, while computing distances between dynamical models is, in general, a difficult problem \cite{Afsari:CVPR12}, one can easily compute dissimilarities between models and data, \eg, using representation or encoding error. {In addition, our method works in situations where only a subset of pairwise dissimilarities are provided.}

\smallskip\noindent\textbf{--} Unlike DPPs \cite{Macchi:AAP75}, kDPPs \cite{Taskar:ICML11} and metric-based methods \cite{Duda:04, Shmoys:TheoryComp12, Li:InfoComp12}, our algorithm works with arbitrary dissimilarities. We do not require that dissimilarities come from a metric, \ie, they can be asymmetric or violate the triangle inequality.

\smallskip\noindent\textbf{--} 
Our algorithm has sampling and clustering theoretical guarantees. More specifically, when there is a grouping of points, defined based on dissimilarities, we show that our method selects representatives from all groups and reveals the clustering of sets. We also obtain the range of the regularization parameter for which the solution of our algorithm changes from selecting a single representative to selecting the maximum number~of~representatives.

\begin{figure}[t!]
\centering
\includegraphics[width=0.48\linewidth, trim = 0 0 0 0 , clip]{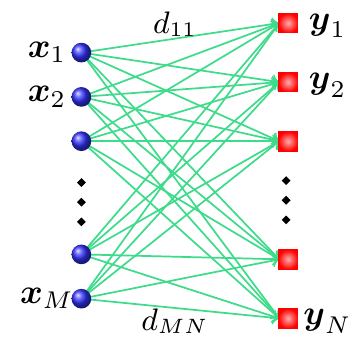}\hspace{1mm}
\includegraphics[width=0.48\linewidth, trim = 0 0 0 0 , clip]{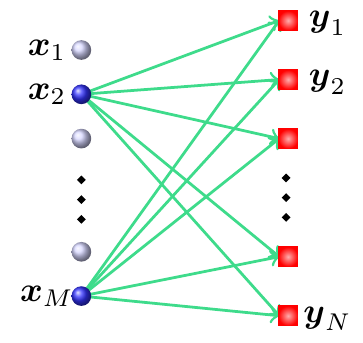}
\caption{\small{Left: The DS3 algorithm takes pairwise dissimilarities between a source set $\X = \{\x_1, \ldots, \x_M\}$ and a target set $\Y = \{\y_1, \ldots, \y_N\}$. The dissimilarity $d_{ij}$ indicates how well $\x_i$ represents $\y_j$. Right: The DS3 algorithm finds a few representative elements of $\X$ that, based on the provided dissimilarities, well represent the set $\Y$.}}
\label{fig:dsrs}
\end{figure}

\smallskip\noindent\textbf{--} Our algorithm can effectively deal with outliers: it does not select outliers in the source set and rejects outliers in the target~set. 

\smallskip\noindent\textbf{--} Our proposed algorithm is based on convex programming, hence, unlike algorithms such as Kmedoids, does not depend on initialization. Since standard convex solvers such as CVX \cite{cvx} do not scale well with increasing the problem size, we consider a computationally efficient implementation of the proposed algorithm using the Alternating Direction Method of Multipliers (ADMM) framework \cite{Boyd:FTML10, Gabay:CMA76}, which results in quadratic complexity in the problem size. We show that our ADMM implementation allows to parallelize the algorithm, hence further reducing the computational~time. 

\smallskip\noindent\textbf{--} Finally, by experiments on real-world datasets, we show that our algorithm improves the state of the art on two problems of categorization using representative images and time-series modeling and segmentation using representative~models.

\section{Dissimilarity-based Sparse Subset Selection (DS3)}
In this section, we consider the problem of finding representatives of a `source set', $\X$, given its pairwise relationships to a `target set', $\Y$. We formulate the problem as a trace minimization problem regularized by a row-sparsity term. The solution of our algorithm finds representatives from $\X$ along with the membership of each element of $\Y$ to each representative. We also show that our algorithm can deal with outliers in both sets effectively.

\begin{figure*}[t!]
\centering
\begin{subfigure}[b]{0.25\textwidth}
\includegraphics[width=.9\textwidth, trim = 88 68 65 43 , clip]{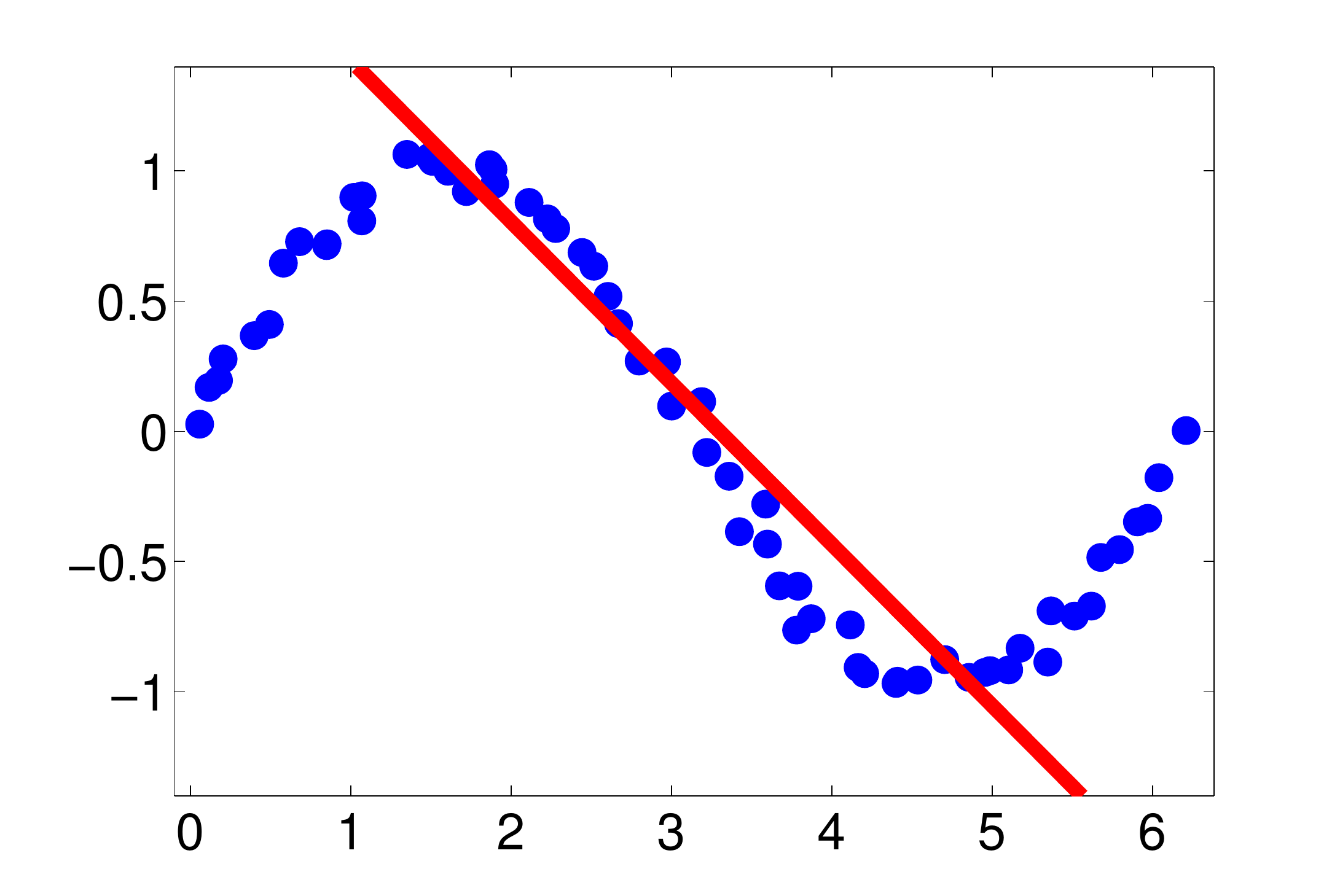}
\vspace{-1mm}
\caption{$\lambda = \lambda_{\max,\infty}$}
\end{subfigure}
\hspace{8mm}
\begin{subfigure}[b]{0.25\textwidth}
\includegraphics[width=.9\textwidth, trim = 88 68 65 43 , clip]{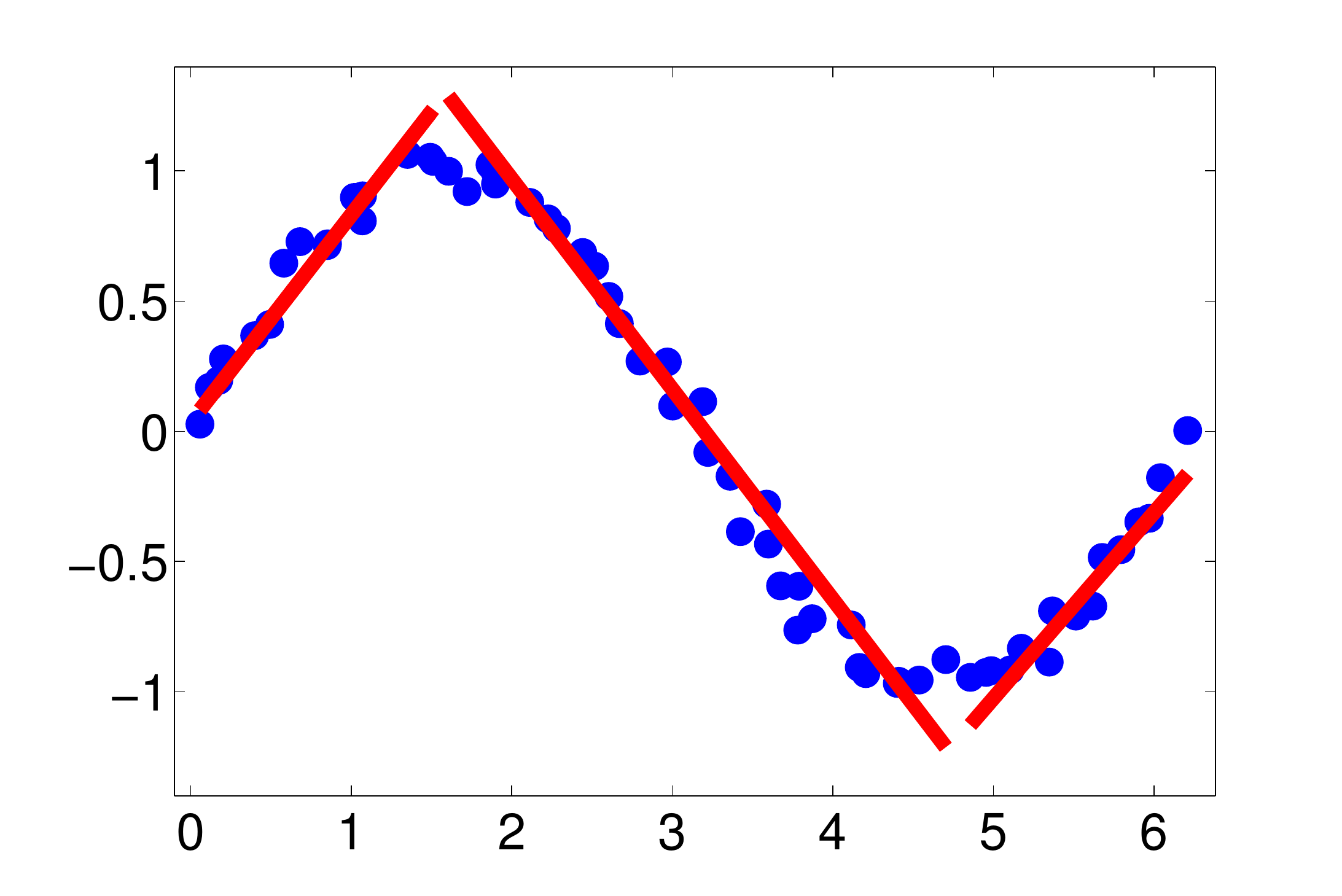}
\vspace{-1mm}
\caption{$\lambda = 0.1 \, \lambda_{\max,\infty}$}
\end{subfigure}
\hspace{8mm}
\begin{subfigure}[b]{0.25\textwidth}
\includegraphics[width=.9\textwidth, trim = 88 68 65 43 , clip]{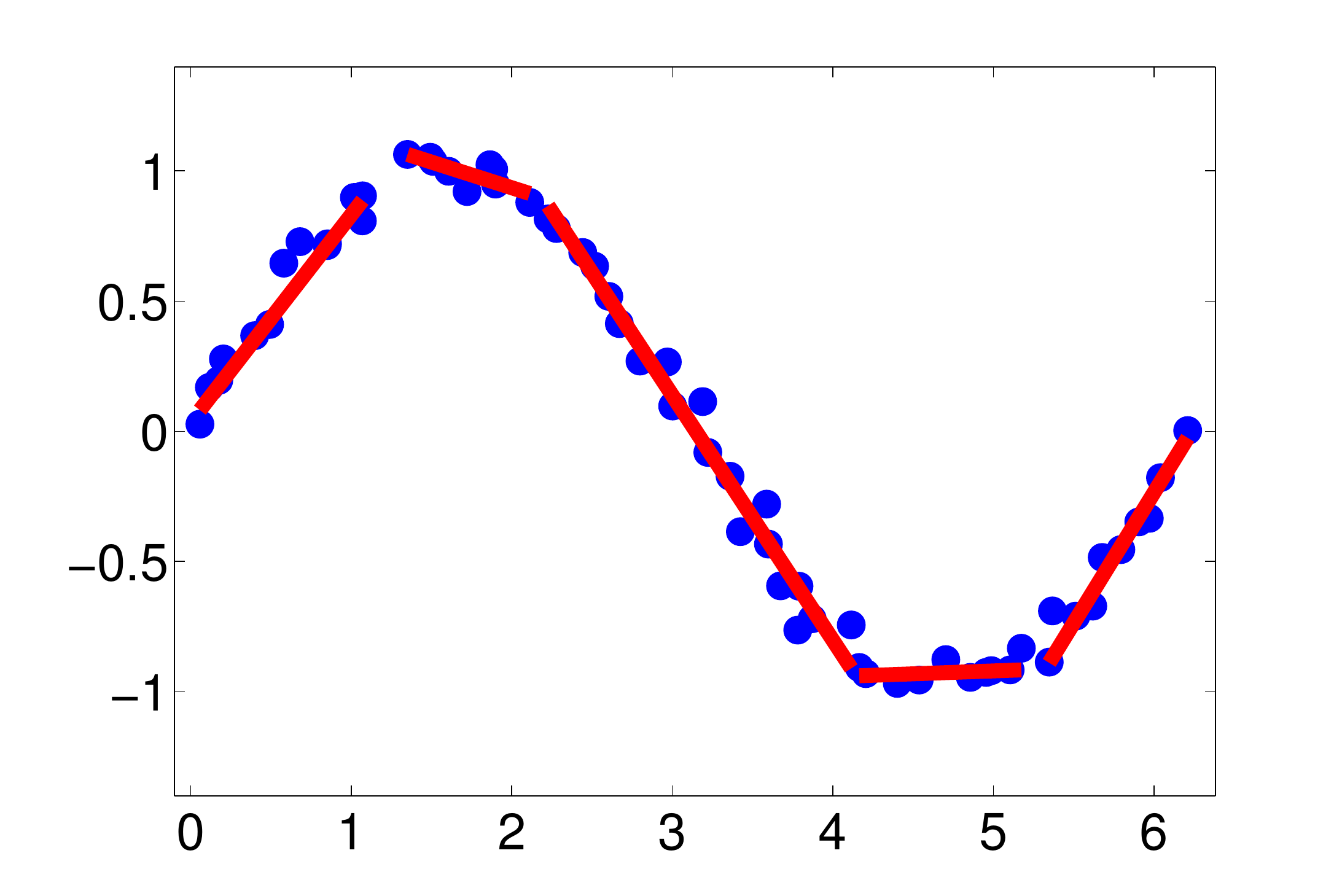}
\vspace{-1mm}
\caption{$\lambda = 0.01 \, \lambda_{\max,\infty}$}
\end{subfigure}
\caption{\small{Finding representative models for noisy data $\{ \y_j \}_{j=1}^{N}$ on a nonlinear manifold. For each data point $\y_j$ and its $K = 4$ nearest neighbors, we learn a one-dimensional affine model with parameters $\btheta_j = (\a_j,b_j)$ so as to minimize the loss $\ell_{\theta}(\y) = | \a^\top \y - b  |$ for the $K+1$ points. We set $\X = \{ \btheta_i \}_{i=1}^{N}$ and $\Y = \{ \y_j \}_{j=1}^{N}$ and compute the dissimilarity between each estimated model $\btheta_i$ and each data point $\y_j$ as $d_{ij} = \ell_{\btheta_i}(\y_j)$. Representative models found by our proposed optimization in \eqref{eq:tracerow-matrix-1} for several values of $\lambda$, with $\lambda_{\max,\infty}$ defined in \eqref{eq:max-lambda}, are shown by red lines. Notice that as we decrease $\lambda$, we obtain a larger number of representative models, which more accurately approximate the nonlinear manifold.}}
\label{fig:RepModel}
\end{figure*}
%

\subsection{Problem Statement}
Assume we have a source set $\X = \{\x_1, \ldots, \x_M\}$ and a target set $\Y = \{\y_1, \ldots, \y_N\}$, which consist of $M$ and $N$ elements, respectively. Assume that we are given pairwise dissimilarities $\{ d_{ij} \}_{i=1,\ldots, M}^{j = 1, \ldots, N}$ between the elements of $\X$ and $\Y$. Each $d_{ij}$ indicates how well $\x_i$ represents $\y_j$, \ie, the smaller the value of $d_{ij}$ is, the better $\x_i$ represents $\y_j$. We can arrange the dissimilarities into a matrix of the form
\begin{equation}
\D \triangleq \begin{bmatrix} \d_1^{\top} \\ \vdots \\ \d_M^{\top}  \end{bmatrix} = \begin{bmatrix} d_{11} & d_{12} & \cdots & d_{1N} \\ \vdots & \vdots & & \vdots \\ d_{M1} & d_{M2} & \cdots & d_{MN}  \end{bmatrix} \in \Re^{M \times N},
\end{equation}
where $\d_i \in \Re^{N}$ denotes the $i$-th row of $\D$. Given $\D$, our goal is to find a small subset of $\X$ that well represents the collection of the elements of $\Y$, as shown in Figure \ref{fig:dsrs}.

In contrast to the state-of-the-art algorithms \cite{Taskar:ICML11, Frey:Science07, Taskar:AISTATS13}, we do not restrict $\X$ and $\Y$ to consist of same type of elements or be identical. For example, $\X$ can be a set of models and $\Y$ be a set of data points, in which case we select a few models that well represent the collection of data points, see Figure \ref{fig:RepModel}. Dissimilarities in this case, can be representation or coding errors of data via models. On the other hand, $\X$ and $\Y$ can consist of the same type of elements or be identical. For example, $\X$ and $\Y$ may correspond to collection of models, hence our goal would be to select representative models. Examples of dissimilarities in this case are distances between dynamical systems and KL divergence between probability distributions. Also, when $\X$ and $\Y$ correspond to data points, our goal would be to select representative data points, see Figure \ref{fig:2G-PZ}. Examples of dissimilarities in this case are Hamming, Euclidean, or geodesic distances between data points.

\subsection{Dissimilarities}
It is important to note that we can work with both similarities $\{ s_{ij} \}$ and dissimilarities $\{ d_{ij} \}$, simply by setting $d_{ij} = - s_{ij}$ in our formulation. For example, when $\X = \Y$, we can set $d_{ij} = - K_{ij}$, where $K$ denotes a kernel matrix on the dataset.

When appropriate vector-space representations of elements of $\X$ and $\Y$ are given, we can compute dissimilarities using a predefined function, such as the encoding error, e.g., $d_{ij} = \| \x_i - \A \y_j \|$ for an appropriate $\A$, Euclidean distance, $d_{ij} = \| \x_i - \y_j \|_2$, or truncated quadratic, $d_{ij} = \min \{ \beta , \| \x_i - \y_j \|_2^2 \}$ where $\beta$ is some constant. However, we may be given or can compute (dis)similarities without having access to vector-space representations, \eg, as edges in a social network graph, as subjective pairwise comparisons between images, or as similarities between sentences computed via a string kernel. Finally, we may learn (dis)similarities, \eg, using metric learning methods \cite{Xing:NIPS02, Dhillon:ICML07}.

\begin{remark}
When the source and target sets are identical, \ie, $\X  = \Y$, we do not require dissimilarities to come from a metric, \ie, they can be asymmetric or violate the triangle inequality. For example, the set of features in an image, containing a scene or an object, can well encode the set of features in another image, containing part of the scene or the object, while the converse is not necessarily true, hence asymmetry of dissimilarities. Also, in document analysis using a Bag of Features (BoF) framework, a long sentence can well represent a short sentence while the converse is not necessarily true \cite{Elhamifar:NIPS12, Frey:Science07}.
\end{remark}
\begin{remark}
Our generalization to work with two sets, \ie, source and target sets, allows to reduce the cost of computing and storing dissimilarities. For instance, when dealing with a large dataset, we can select a small random subset of the dataset as the source set with the target set being the rest or the entire dataset. 
\end{remark}
%

\subsection{DS3 Algorithm}

Given $\D$, our goal is to select a subset of $\X$, called \emph{representatives} or \emph{exemplars}, that efficiently represent $\Y$. To do so, we consider an optimization program on unknown variables $z_{ij}$ associated with dissimilarities $d_{ij}$. We denote the matrix of all variables~by
\begin{figure*}[t!]
\centering
\begin{subfigure}[b]{0.23\textwidth}
\includegraphics[width=\textwidth, trim = 56 23 46 40 , clip]{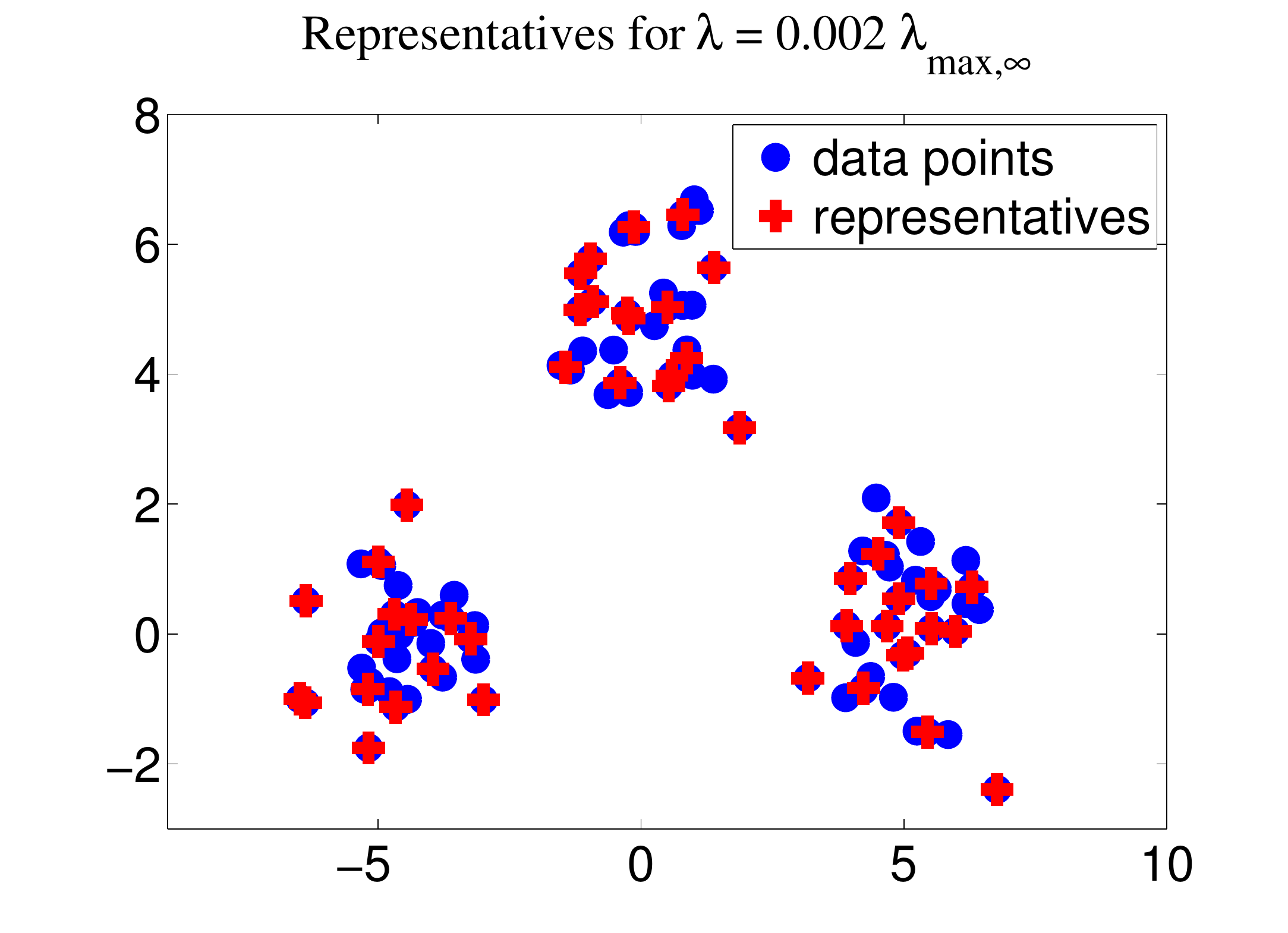}\\
\vspace{-1mm}
\includegraphics[width=\textwidth, trim = 33 20 34 40 , clip]{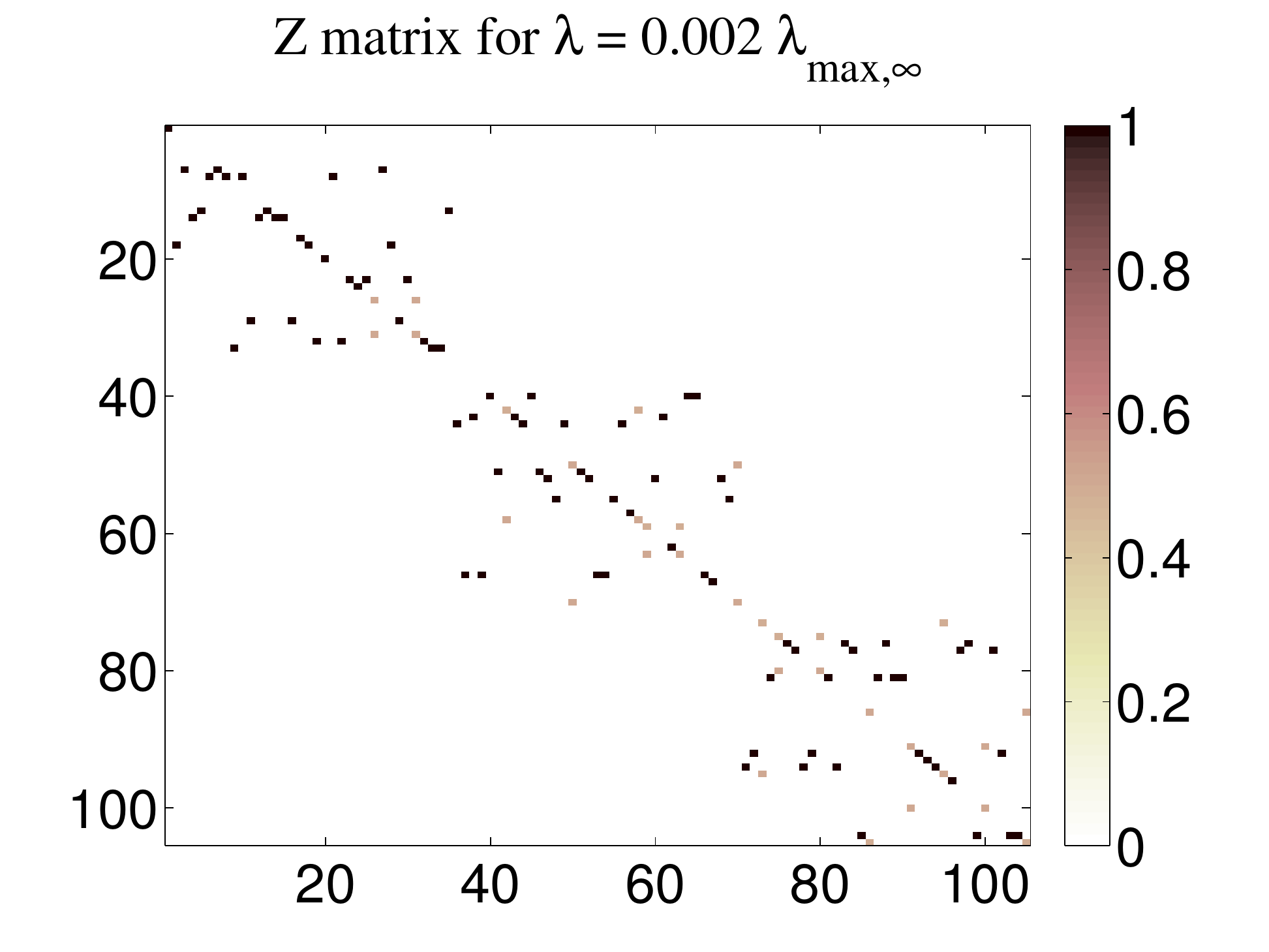}
\caption{$\lambda = 0.002 \, \lambda_{\max,\infty}$}
\end{subfigure}
\hspace{1mm}
\begin{subfigure}[b]{0.23\textwidth}
\includegraphics[width=\textwidth, trim = 56 23 46 40 , clip]{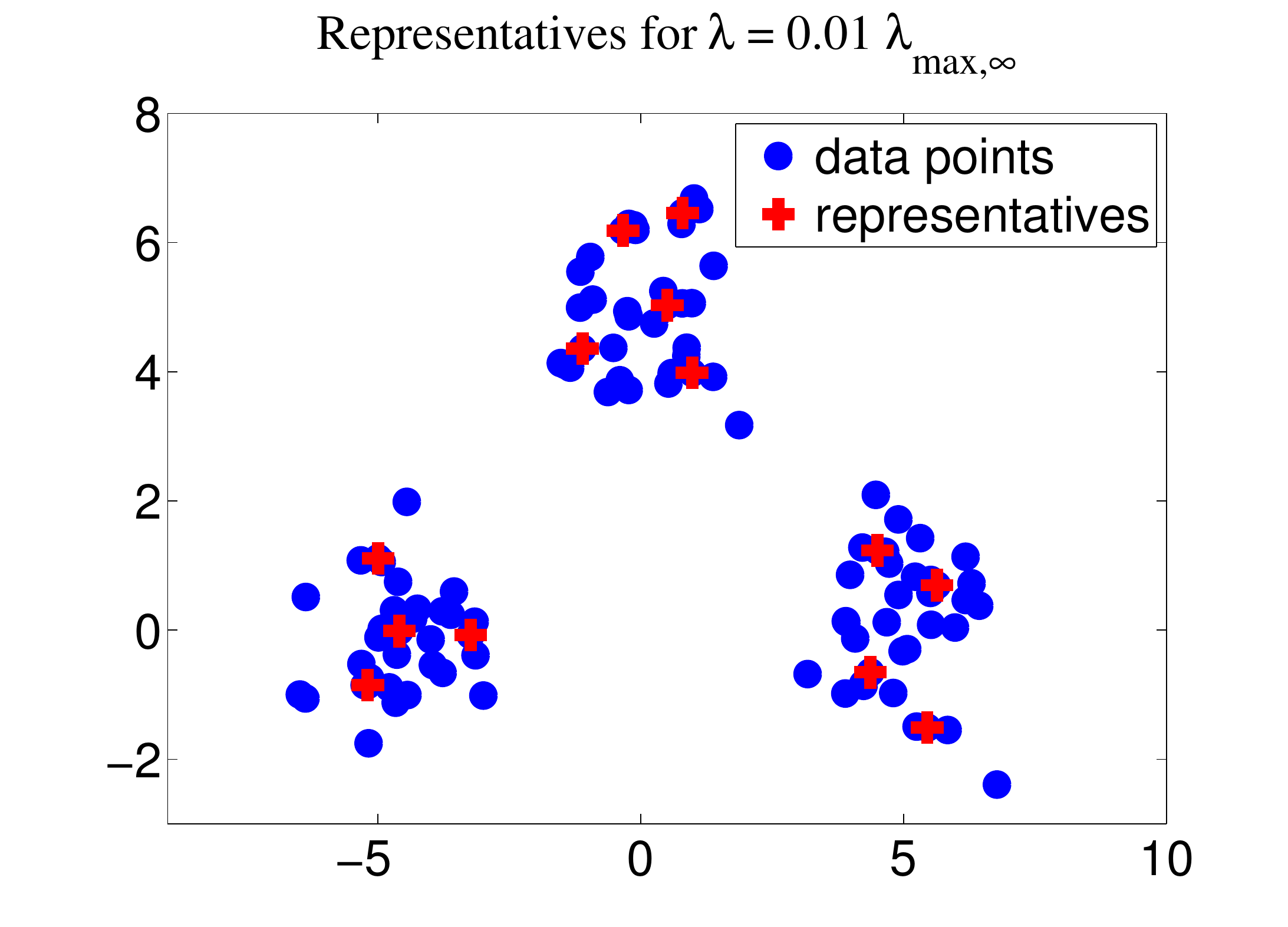}\\
\vspace{-1mm}
\includegraphics[width=\textwidth, trim = 33 20 34 40 , clip]{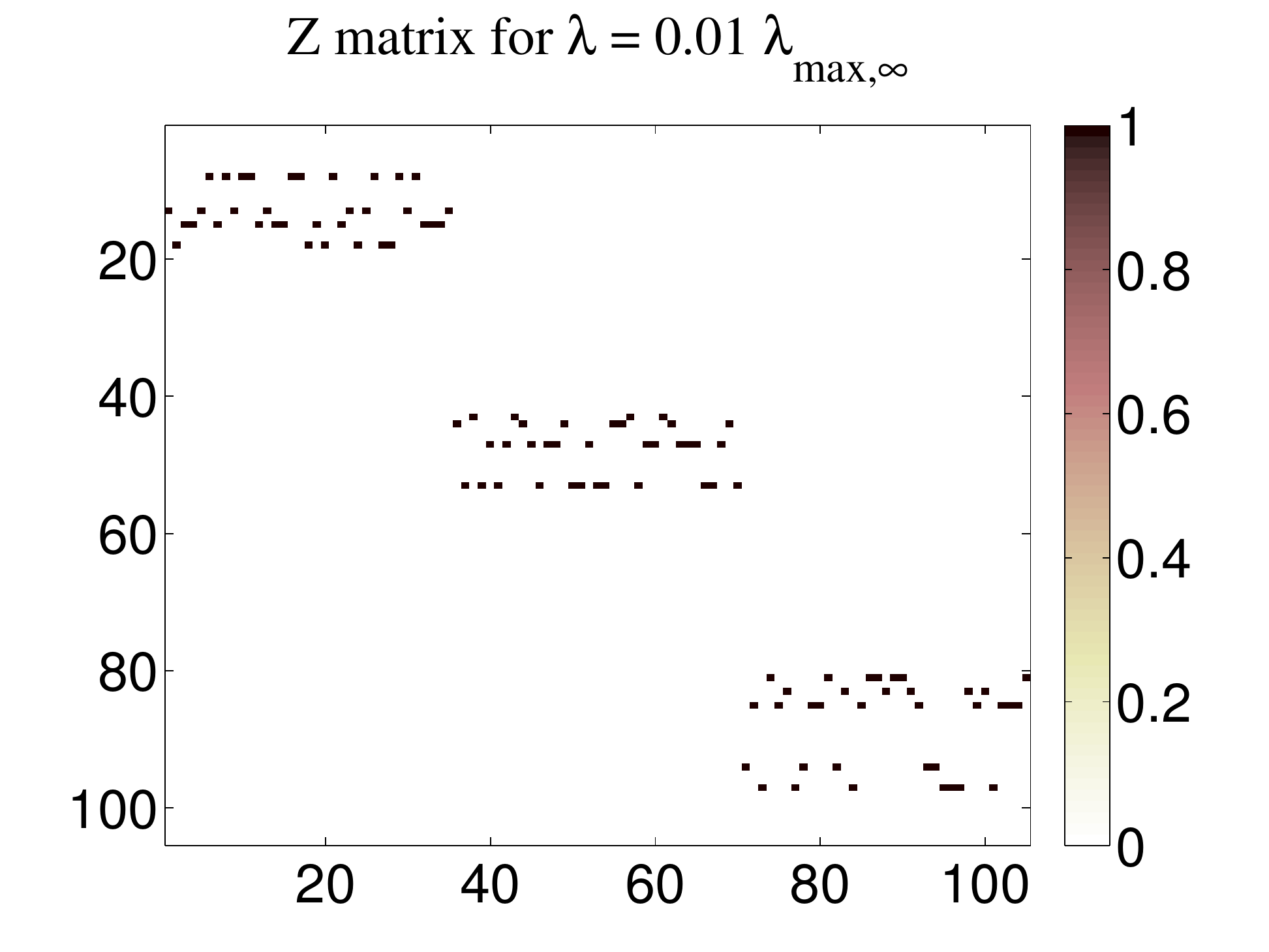}
\caption{$\lambda = 0.01 \, \lambda_{\max,\infty}$}
\end{subfigure}
\hspace{1mm}
\begin{subfigure}[b]{0.23\textwidth}
\includegraphics[width=\textwidth, trim = 56 23 46 40 , clip]{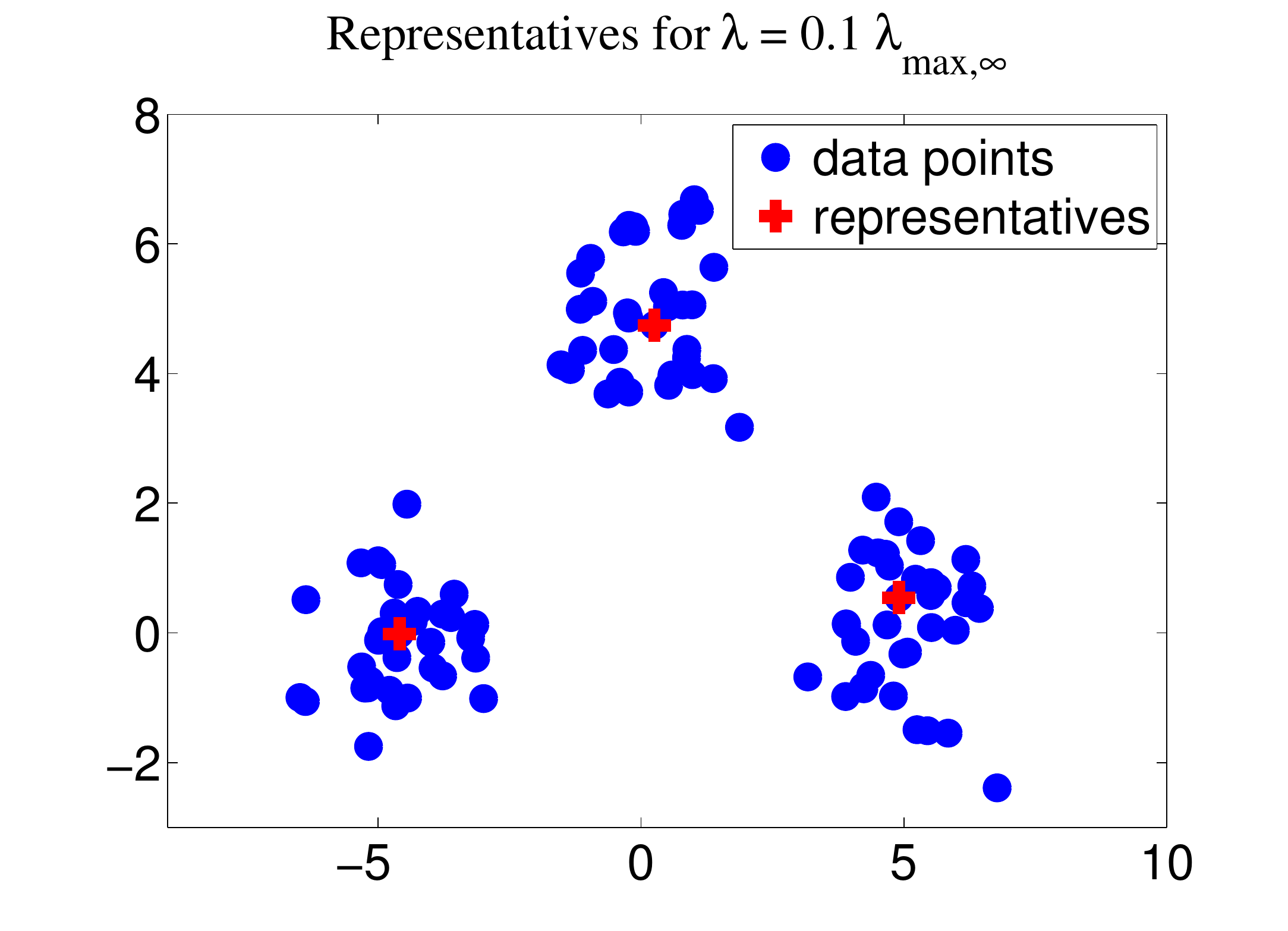}\\
\vspace{-1mm}
\includegraphics[width=\textwidth, trim = 33 20 34 40 , clip]{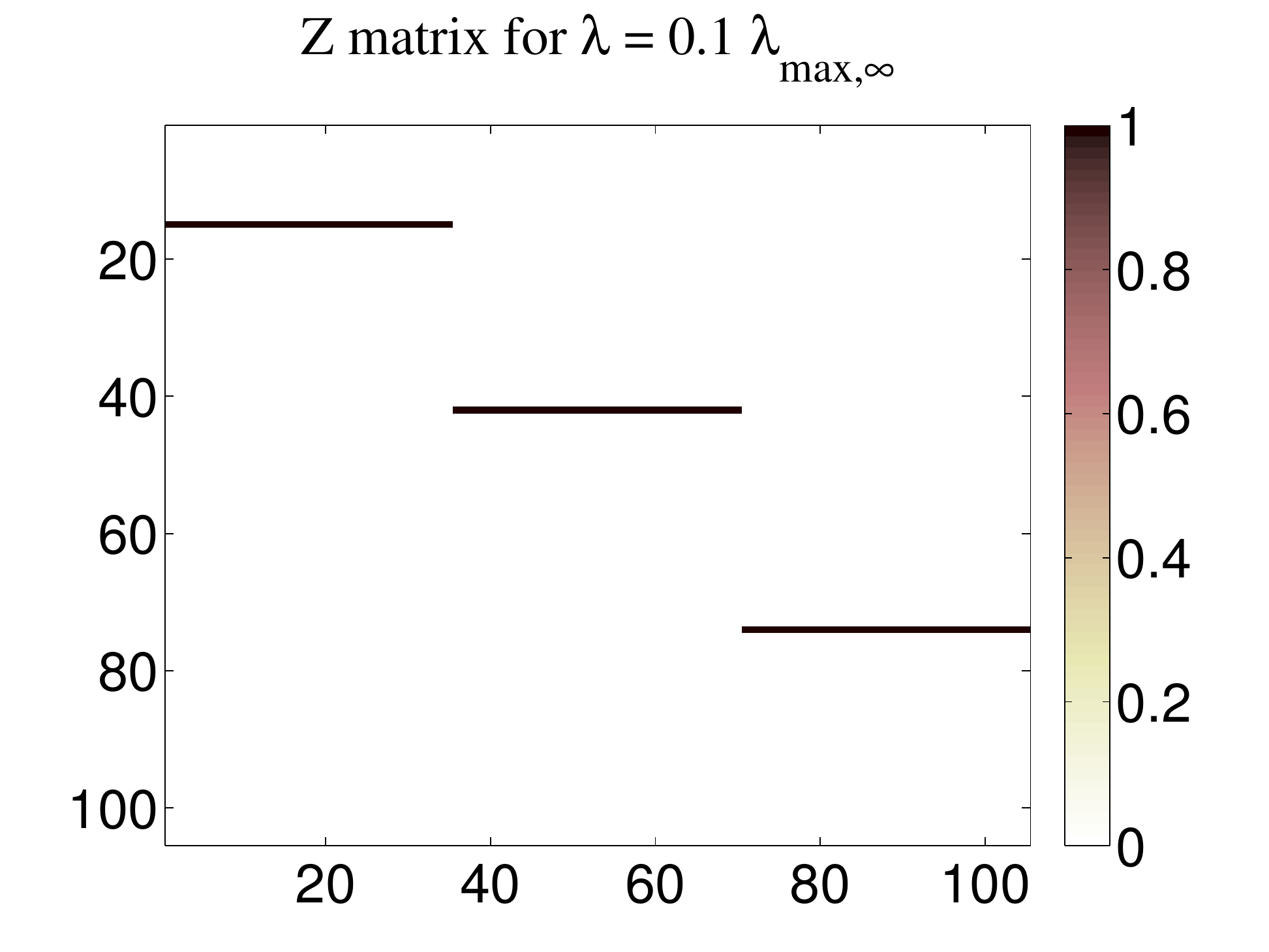}
\caption{$\lambda = 0.1 \, \lambda_{\max,\infty}$}
\end{subfigure}
\hspace{1mm}
\begin{subfigure}[b]{0.23\textwidth}
\includegraphics[width=\textwidth, trim = 56 23 46 40 , clip]{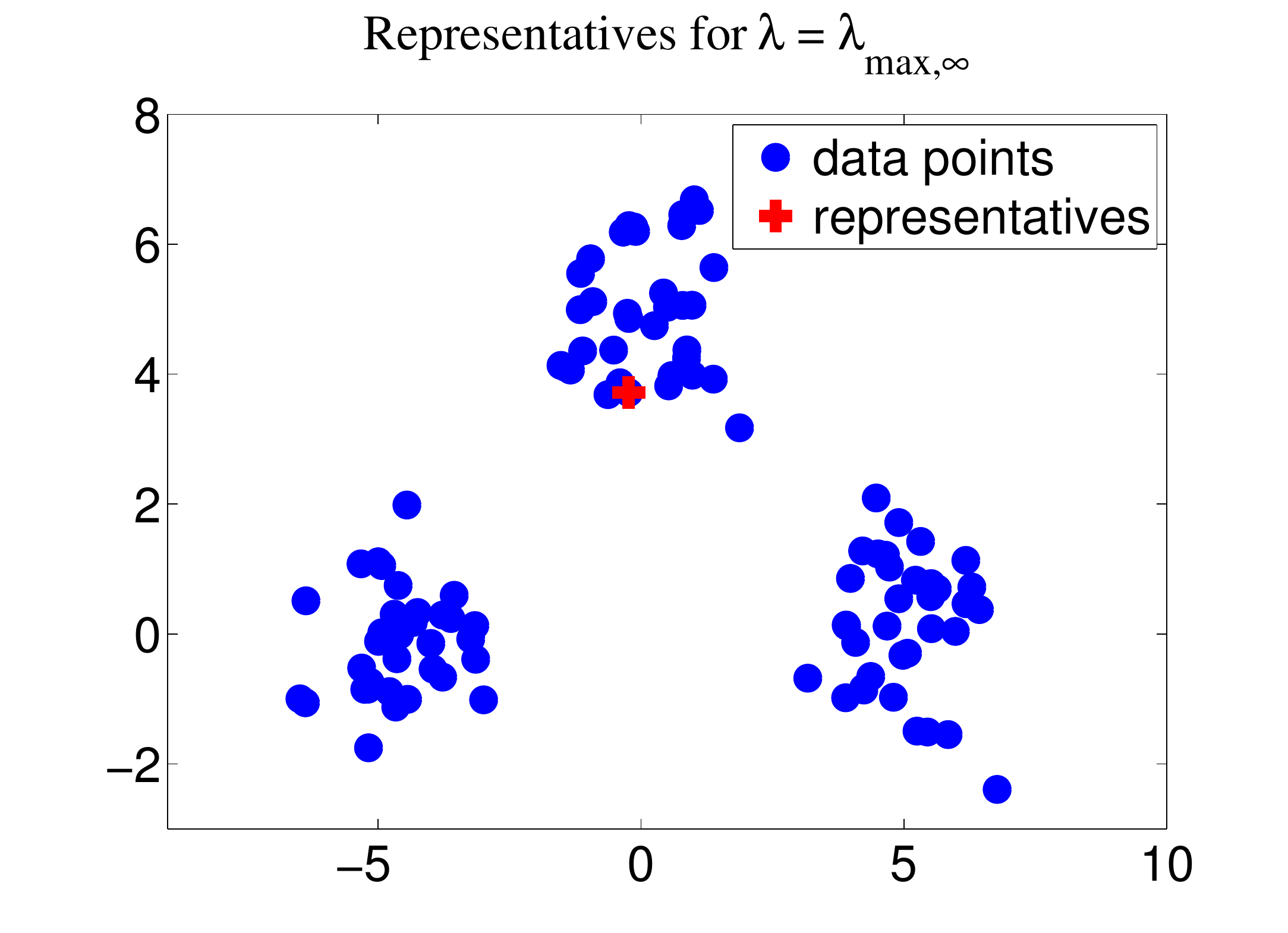}\\
\vspace{-1mm}
\includegraphics[width=\textwidth, trim = 33 20 34 40 , clip]{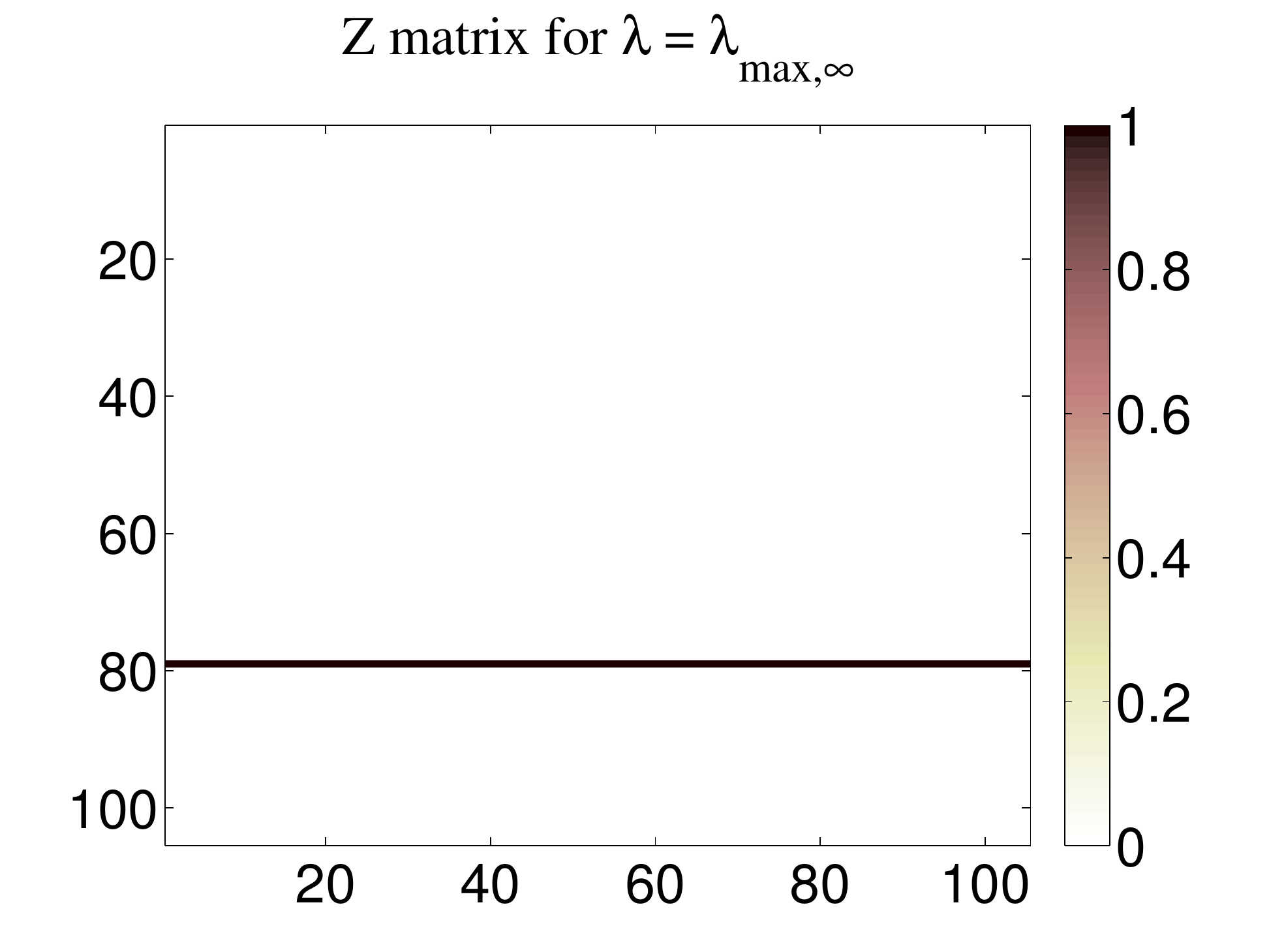}
\caption{$\lambda = \lambda_{\max,\infty}$}
\end{subfigure}
\caption{\small{Top: Data points (blue circles) drawn from a mixture of three Gaussians and the representatives (red pluses) found by our proposed optimization program in \eqref{eq:tracerow-matrix-1} for several values of $\lambda$, with $\lambda_{\max,\infty}$ defined in \eqref{eq:max-lambda}. Dissimilarity is chosen to be the Euclidean distance between each pair of data points. As we increase $\lambda$, the number of representatives decreases. Bottom: the matrix $\Z$ obtained by our proposed optimization program in \eqref{eq:tracerow-matrix-1} for several values of $\lambda$. The nonzero rows of $\Z$ indicate indices of the representatives. In addition, entries of $\Z$ provide information about the association probability of each data point with each representative.}}
\label{fig:2G-PZ}
\end{figure*}
\begin{equation}
\Z \triangleq \begin{bmatrix} \z_1^{\top} \\ \vdots \\ \z_M^{\top}  \end{bmatrix} = \begin{bmatrix} z_{11} & z_{12} & \cdots & z_{1N} \\  \vdots & \vdots & & \vdots \\ z_{M1} & z_{M2} & \cdots & z_{MN}  \end{bmatrix} \in \Re^{M \times N},
\end{equation}
where $\z_i \in \Re^N$ is the $i$-th row of $\Z$. We interpret $z_{ij} \in \{0, 1\}$ as the indicator of $\x_i$ representing $\y_j$, which is one when $\x_i$ is the representative of $\y_j$ and is zero otherwise. To ensure that each $\y_j$ is represented by one representative, we must have $\sum_{i=1}^{N}{z_{ij}} = 1$.

\subsubsection{Simultaneous Sparse Recovery-Based Optimization}
To select \emph{a few elements} of $\X$ that \emph{well encode} $\Y$ according to dissimilarities, we propose a row-sparsity regularized trace minimization program on $\Z$, that pursues two goals. First, we want representatives to well encode $\Y$. If $\x_i$ is chosen to be a representative of $\y_j$, the cost of encoding $\y_j$ via $\x_i$ is $d_{ij} z_{ij} \in \{0,d_{ij}\}$. Hence, the cost of encoding $\y_j$ using $\X$ is $\sum_{i=1}^{N}{d_{ij} z_{ij}}$ and the cost of encoding $\Y$ via $\X$ is $\sum_ {j=1}^{N}{\sum_{i=1}^{M}{d_{ij} z_{ij}}}$. Second, we would like to have as few representatives as possible. Notice that when $\x_i$ is a representative of some of the elements of $\Y$, we have $\z_i \neq \0$, \ie, the $i$-th row of $\Z$ is nonzero. Thus, having a few representatives  corresponds to having a few nonzero rows in the matrix $\Z$.

Putting these two goals together, we consider the following optimization program
\begin{equation}
\label{eq:tracerow-elements-0}
\begin{split}
&\min_{\{ z_{ij} \}} ~ \lambda \sum_{i=1}^{M}{\Io(\| \z_i \|_p)} + \sum_ {j=1}^{N}{\sum_{i=1}^{M}{d_{ij} z_{ij}}} \\ 
&\st~~ \sum_{i=1}^{M}{z_{ij}} = 1 ,~\forall j; ~~z_{ij} \in \{0,1\} ,~\forall i,j,
\end{split}
\end{equation}
where $\| \cdot \|_p$ denotes the $\ell_p$-norm and $\Io(\cdot)$ denotes the indicator function, which is zero when its argument is zero and is one otherwise. The first term in the objective function corresponds to the number of representatives and the second term corresponds to the total cost of encoding $\Y$ via representatives. The regularization parameter $\lambda > 0$ sets the trade-off between the two terms. Since the minimization in \eqref{eq:tracerow-elements-0}, which involves counting the number of nonzero rows of $\Z$ and binary constraints $z_{ij} \in \{0, 1\}$ is non-convex and, in general, NP-hard, we consider the following convex relaxation
\begin{equation}
\label{eq:tracerow-elements-1}
\begin{split}
&\min_{\{ z_{ij} \}} ~ \lambda \sum_{i=1}^{M}{\| \z_i \|_p} + \sum_ {j=1}^{N}{\sum_{i=1}^{M}{d_{ij} z_{ij}}} \\
&\st~~ \sum_{i=1}^{M}{z_{ij}} = 1 ,~\forall j; ~~z_{ij} \geq 0 ,~\forall i,j,
\end{split}
\end{equation}
where, instead of counting the number of nonzero rows of $\Z$, we use the sum of $\ell_p$-norms of rows of $\Z$. In addition, we use the relaxation $z_{ij} \in [0,1]$, hence, $z_{ij}$ acts as the probability of $\x_i$ representing $\y_j$. Notice that for $p \geq 1$, the optimization above is convex. We choose $p \in \{ 2, \infty \}$, where for $p = 2$, we typically obtain a soft assignment of representatives, \ie, $\{z_{ij}\}$ are in the range $[0,1]$, while for $p = \infty$, we typically obtain a hard assignment of representatives, \ie, $\{z_{ij}\}$ are in $\{0,1\}$.\footnote{Notice that $p = 1$ also imposes \emph{sparsity} of the elements of the nonzero \emph{rows} of $\Z$, which is not desirable since it promotes only a few points in $\Y$ to be associated with each representative in $\X$.} 
We can rewrite the optimization program \eqref{eq:tracerow-elements-1} in the matrix form as
\begin{equation}
\label{eq:tracerow-matrix-1}
\begin{split}
&\min_{\Z} \; \lambda \| \Z \|_{1,p} + \tr(\D^{\top} \Z) \\
&\st ~~ \1^{\top} \Z = \1^{\top}, ~ \Z \geq \0,
\end{split}
\end{equation}
where $\| \Z \|_{1,p} \triangleq \sum_{i=1}^{M}{\| \z_i \|_p}$ and $\1$ denotes a vector, of appropriate dimension, whose elements are all equal to one. In addition, $\tr(\cdot)$ denotes the trace operator. We also write $\tr(\D^\top \Z) = \langle \D , \Z \rangle$, \ie, the inner product of $\D$ and $\Z$. Once we solve the optimization program \eqref{eq:tracerow-matrix-1}, we can find representative indices from the nonzero rows of the solution, $\Z^*$. 
\begin{remark}
We can deal with the case where only a subset of entries of $\D$ are given. More specifically, let $\Omega$ and $\Omega^c$ denote indices of observed and missing entries of $\D$, respectively. We can find representatives by replacing $\tr(\D^\top \Z)$ with $\langle \D_{\Omega} , \Z_{\Omega} \rangle$ and adding the constraint $\Z_{\Omega^c} = \0$ in~\eqref{eq:tracerow-matrix-1}.
\end{remark}

Later in the section, we highlight connections of our proposed formulation to integer programming-based formulations and facility location algorithms.

\subsubsection{Regularization Parameter Effect}
As we change the regularization parameter $\lambda$ in \eqref{eq:tracerow-matrix-1}, the number of representatives found by our algorithm changes. For small values of $\lambda$, where we put more emphasis on better encoding of $\Y$ via $\X$, we obtain more representatives. In the limiting case of $\lambda \rightarrow 0$ each element of $\Y$ selects its closest element from $\X$ as its representative, \ie, $z_{i^*_j j} = 1$, where, $i^*_j \triangleq \argmin_{i}{d_{ij}}$. On the other hand, for large values of $\lambda$, where we put more emphasis on the row-sparsity of $\Z$, we select a small number of representatives. For a sufficiently large $\lambda$, we select only one representative from $\X$. In Section \ref{sec:theory}, we compute the range of $\lambda$ for which the solution of \eqref{eq:tracerow-matrix-1} changes from one representative to the largest possible number of representatives.

Figure \ref{fig:RepModel} demonstrates an example of approximating a nonlinear manifold using representative affine models learned from noisy data by solving \eqref{eq:tracerow-matrix-1} with $p = \infty$. Notice that as we decrease $\lambda$, we select a larger number of affine models, which better approximate the manifold. Figure \ref{fig:2G-PZ}  illustrates the representatives (top row) and the matrix $\Z$ (bottom row), for $p = \infty$ and several values of $\lambda$, for a dataset drawn from a mixture of three Gaussians with dissimilarities being Euclidean distances between points (see the supplementary materials for similar results with $p = 2$).

\subsection{Dealing with Outliers}
\label{sec:outlier}
\begin{figure}[t!]
\centering
\begin{subfigure}[b]{0.225\textwidth}
\includegraphics[width=\textwidth, trim = 44 20 40 30 , clip]{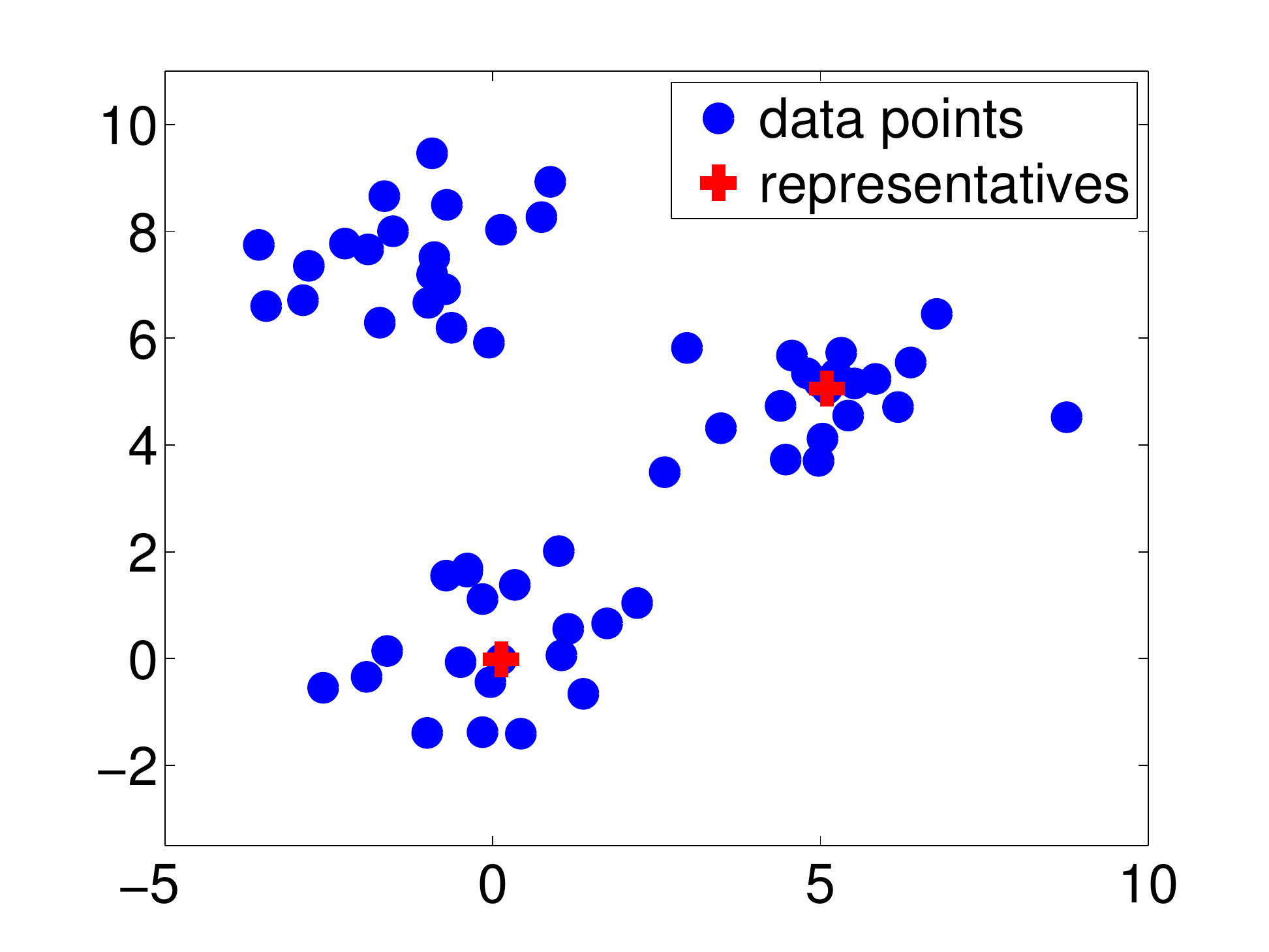}
\vspace{-4mm}
\caption{Source set}
\label{fig:outliers-s}
\end{subfigure}
\hspace{0mm}
\begin{subfigure}[b]{0.225\textwidth}
\includegraphics[width=\textwidth, trim = 44 20 40 30 , clip]{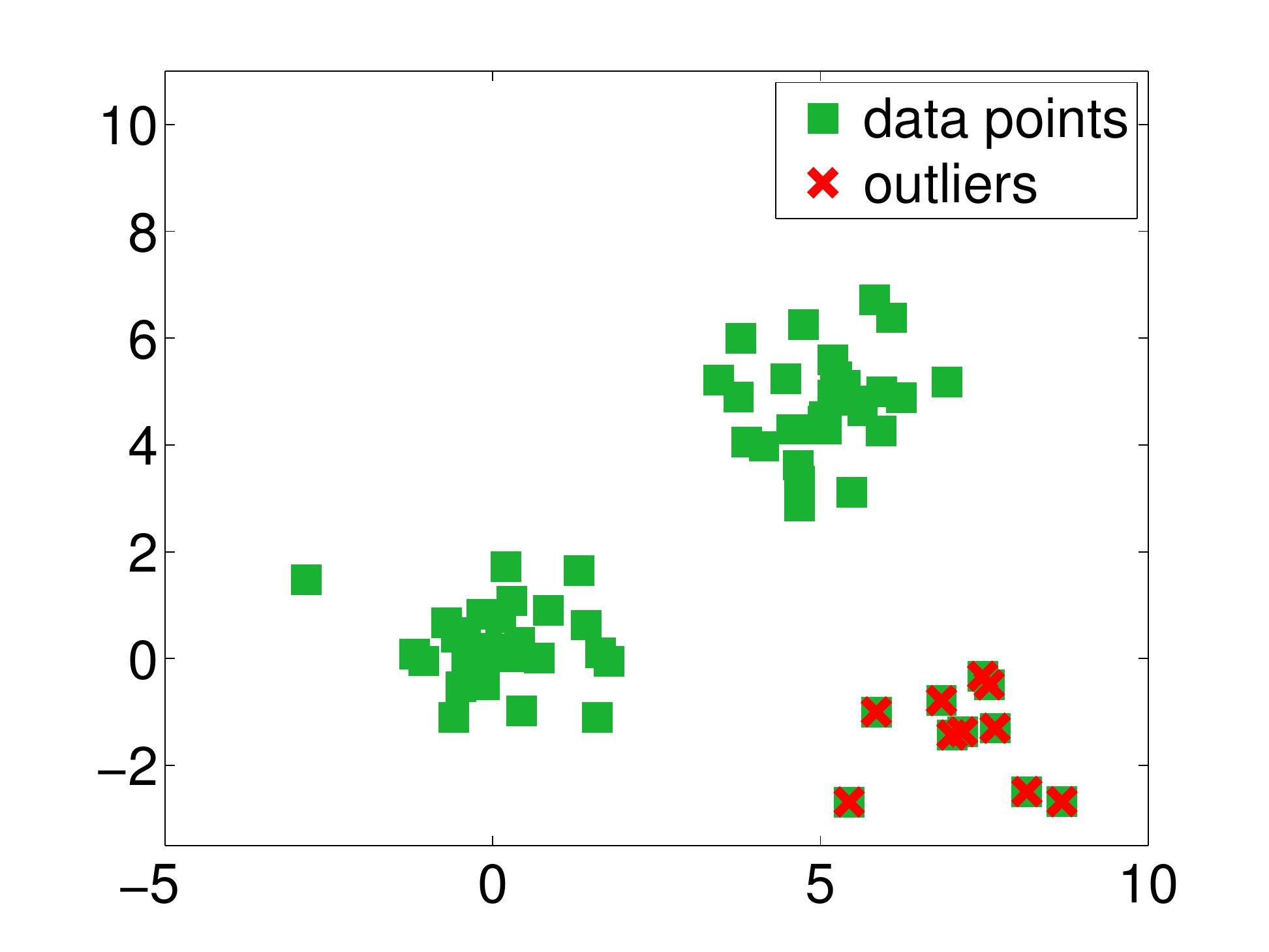}
\vspace{-4mm}
\caption{Target set}
\label{fig:outliers-t}
\end{subfigure}
\caption{\small{We generate a source set by drawing data points (blue circles) from a mixture of Gaussians with means $(0,0)$, $(5,5)$ and $(-1,7)$. We generate a target set by drawing data points (green squares) from a mixture of Gaussians with means $(0,0)$, $(5,5)$ and $(7,-1)$. Representatives (red pluses) of the source set and outliers (red crosses) of the target set found by our proposed optimization in \eqref{eq:tracerow-matrix-outlier1} with $w_i = 0.3$ are shown. Dissimilarity is the Euclidean distance between each source and target data point. Notice that we only select representatives from the two clusters with means $(0,0)$, $(5,5)$ that also appear in the target set. Our method finds the cluster with the mean $(7,-1)$ in the target set as outlier since there are no points in the source set efficiently encoding it.}}
\label{fig:outliers}
\end{figure}

In this section, we show that our framework can effectively deal with outliers in source and target sets.\footnote{In \cite{Elhamifar:NIPS12}, we showed that for the identical source and target sets, we can detect an outlier as an element that only represents itself. Here, we address the general setting where the source and target sets are different. However, our framework is also applicable to the scenario where the two sets are identical.} Notice that an outlier in the source set corresponds to an element that cannot effectively represent elements of the target set. Since our framework selects representatives, such outliers in $\X$ will not be selected, as shown in Figure \ref{fig:outliers-s}. In fact, this is one of the advantages of finding representatives, which, in addition to reducing a large set, helps to reject outliers in $\X$.

On the other hand, the target set, $\Y$, may contain outlier elements, which cannot be encoded efficiently by any element of $\X$. For example, when $\X$ and $\Y$ correspond, respectively, to sets of models and data points, some of the data may not be explained efficiently by any of the models, \eg, have a large representation error. Since the optimization program \eqref{eq:tracerow-matrix-1} requires every element of $\Y$ to be encoded, enforcing outliers to be represented by $\X$ often results in the selection of undesired representatives. In such cases, we would like to detect outliers and allow the optimization not to encode outliers via representatives.

To achieve this goal, we introduce a new optimization variable $e_j \in [0,1]$ associated with each $\y_j$, whose value indicates the probability of $\y_j$ being an outlier. We propose to solve 
\begin{equation}
\label{eq:tracerow-elements-outlier1}
\begin{split}
&\min_{\{ z_{ij} \}, \{ e_j\}} ~ \lambda \sum_{i=1}^{M}{\| \z_i \|_p} + \sum_ {j=1}^{N}{\sum_{i=1}^{M}{d_{ij} z_{ij}}} + \sum_{j=1}^{N} {w_j e_j} \\
&\st~~ \sum_{i=1}^{M}{z_{ij}} + e_j = 1,~\forall j;~z_{ij} \geq 0,~\forall i,j;~~e_j \geq 0,~\forall j.
\end{split}
\end{equation}
The constraints of the optimization above indicate that, for each $\y_j$, the probability of being an inlier, hence being encoded via $\X$, plus the probability of being an outlier must be one. When $e_j = 0$, we have $\sum_{i=1}^{M}{z_{ij}} = 1$. Hence, $\y_j$ is an inlier and must be encoded via $\X$. On the other hand, if $e_j = 1$, we have $\sum_{i=1}^{M}{z_{ij}} = 0$. Hence, $\y_j$ is an outlier and will not be encoded via $\X$. The weight $w_j > 0$ puts a penalty on the selection of $\y_j$ as an outlier. The smaller the value of $w_j$ is, the more likely $\y_j$ is an outlier. Notice that without such a penalization, \ie, when every $w_j$ is zero, we obtain the trivial solution of selecting all elements of $\Y$ as outliers, since by only penalizing $z_{ij}$ in the objective function, we obtain that every $z_{ij} = 0$ and every $e_j = 1$.

We can also rewrite the optimization \eqref{eq:tracerow-elements-outlier1} in the matrix form as 
\begin{equation}
\label{eq:tracerow-matrix-outlier1}
\begin{split}
&\min_{\Z, \e} \; \lambda \| \Z \|_{1,p} + \tr(\begin{bmatrix} \D \\ \w^\top \end{bmatrix}^\top \begin{bmatrix} \Z \\ \e^\top \end{bmatrix}) \\
&\st ~~ \1^\top \begin{bmatrix} \Z \\ \e^\top \end{bmatrix} = \1^{\top}, ~ \begin{bmatrix} \Z \\ \e^\top \end{bmatrix} \geq \0,
\end{split}
\end{equation}
where $\e = \begin{bmatrix} e_1 \!\!& \ldots \!\!& e_N \end{bmatrix}^\top \!\in\! \Re^N$ is the outlier indicator vector and $\w = \begin{bmatrix} w_1 \!\!& \ldots \!\!& w_N \end{bmatrix}^\top \! \in \! \Re^N$ is the corresponding weight vector. 

\begin{remark} 
Notice that comparing \eqref{eq:tracerow-matrix-outlier1} with \eqref{eq:tracerow-matrix-1}, we have augmented matrices $\Z$ and $\D$ with row vectors $\e^\top$ and $\w^\top$, respectively. This can be viewed as adding to $\X$ a new element, which acts as the representative of outliers in $\Y$ with the associated cost of $\w^\top \e$. At the same time, using $\| \Z \|_{1,p}$ in \eqref{eq:tracerow-matrix-outlier1}, we only penalize the number of representatives for the inliers in $\Y$.
\end{remark}

{One possible choice for the weights is to set $w_j = w$ for all $j$, which results in one additional regularization parameter with respect to \eqref{eq:tracerow-matrix-1}. Another choice for the outlier weights is to set
\begin{equation}
\label{eq:outlierweights}
w_j = \beta \, e^{- \frac{ \min_{i} d_{ij} }{ \tau }},
\end{equation}
for non-negative parameters $\beta$ and $\tau$. In other words, when there exists an element in the source set that can well represent $\y_j$, the likelihood of $\y_j$ being an outlier should decrease, \ie, $w_j$ should increase, and vice versa. The example in Figure \ref{fig:outliers-t} illustrates the effectiveness of our optimization in \eqref{eq:tracerow-matrix-outlier1} for dealing with outliers.

\subsection{Clustering via Representatives}
It is important to note that the optimal solution $\Z^*$ in \eqref{eq:tracerow-matrix-1} and \eqref{eq:tracerow-matrix-outlier1}, not only indicates the elements of $\X$ that are selected as representatives, but also contains information about the membership of elements of $\Y$ to representatives. More specifically, $\begin{bmatrix} z^*_{1j} \!& \ldots \!& z^*_{Mj} \end{bmatrix}^\top$ corresponds to the probability vector of $\y_j$ being represented by each element of $\X$. Hence, we obtain a soft assignment of $\y_j$ to representatives since $z^*_{ij} \in [0,1]$. 

We can also obtain a hard assignment, hence a clustering of $\Y$ using the solution of our optimization. More specifically, if $\{ \x_{\ell_1}, \ldots, \x_{\ell_K} \}$ denotes the set of representatives, then we can assign $\y_j$ to the representative $\x_{\delta_j}$ according to

\begin{equation}
\label{eq:hardassignment}
\delta_j = \argmin_{i \in \{\ell_1,\ldots, \ell_K\}}{d_{i j}}.
\end{equation}
Thus, we can obtain a partitioning of $\Y$ into $K$ groups corresponding to $K$ representatives. In Section \ref{sec:theory}, we show that when $\X$ and $\Y$ jointly partition into multiple groups based on dissimilarities (see Definition \ref{def:clustering} in Section \ref{sec:theory}), then elements of $\Y$ in each group select representatives from elements of $\X$ in the same group. 

\begin{remark}
\label{rem:biclustering}
While the number of clusters is determined by the number of representatives in \eqref{eq:hardassignment}, we can obtain a smaller number of clusters, if desired, by using co-clustering methods \cite{Dhillon:KDD01, Dhillon:KDD03, Dhillon:JMLR07} by jointly partitioning the bi-partite graph of similarities between representatives and $\Y$ into the desired number of groups. 
\end{remark}

\subsection{Alternative Formulation and Relationships to Integer Programming-Based Formulations}
The optimization in \eqref{eq:tracerow-matrix-1} does not directly enforce a specific number of representatives, instead it aims at balancing the encoding cost and number of representatives via $\lambda$. An alternative convex formulation (for $p \geq 1$), which is related to \eqref{eq:tracerow-matrix-1} via Lagrange multiplier, is
\begin{equation}
\label{eq:dsmrs-v2}
\min_{\Z} ~ \tr(\D^\top \Z) ~~ \st ~~ \1^\top \Z = \1^\top, ~ \Z \geq 0, ~ \| \Z \|_{1,p} \leq \tau,
\end{equation}
where $\tau > 0$ is the regularization parameter. In fact, \eqref{eq:dsmrs-v2} aims at minimizing the encoding cost given a representative `budget' $\tau$. For $p = \infty$, which typically results in $\{0, 1\}$ elements in the solution, $\tau$ corresponds to the desired number of representatives.

In fact, \eqref{eq:tracerow-matrix-1} and \eqref{eq:dsmrs-v2} can be thought of as generalization and relaxation of, respectively, uncapacitated and capacitated facility location problem \cite{Charikar:JCSS02}, where we relax the binary constraints $z_{ij} \in \{0, 1\}$ to $z_{ij} \in [0, 1]$ and use arbitrary $\ell_p$-norm, instead of the $\ell_{\infty}$-norm, on rows of $\Z$. Thanks to our formulations, as we show in Section \ref{sec:ds3implementation}, we can take advantage of fast methods to solve the convex optimization efficiently instead of solving an integer or a linear program. More importantly, our result in this paper and our earlier work \cite{Elhamifar:NIPS12} is the first showing the integrality of convex program for clustering, \ie, the solution of our algorithm is guaranteed to cluster the data in non-trivial situations, as we show in Section \ref{sec:theory}. More discussions and extension of \eqref{eq:dsmrs-v2} to dealing with outliers can be found in the supplementary materials.

\section{DS3 Implementation}
\label{sec:ds3implementation}
In this section, we consider an efficient implementation of the DS3 algorithm using the Alternating Direction Method of Multipliers (ADMM) framework \cite{Boyd:FTML10, Gabay:CMA76}. We show that our ADMM implementation results in computational complexity of $O(MN)$, where $M$ and $N$ are, respectively, the number of rows and columns of the dissimilarity matrix. {Moreover, we show that our proposed framework is highly parallelizable, hence, we can further reduce the computational time.} 

We consider the implementation of our proposed optimization in \eqref{eq:tracerow-matrix-1} using the ADMM approach (generalization to \eqref{eq:tracerow-matrix-outlier1} is similar and straightforward). To do so, we introduce an auxiliary matrix $\C \in \Re^{M \times N}$ and consider the optimization program

\begin{table*}[t!]
\caption{\footnotesize Average computational time (sec.) of CVX (Sedumi solver) and the proposed ADMM algorithm ($\mu = 0.1$) for $\lambda = 0.01 \, \lambda_{\max,p}$ over $100$ trials on randomly generated datasets of size $N \times N$. 
} 
\centering
\begin{small}
\begin{tabular}{|@{\;\,}c@{\;\,}|@{\;\,}c@{\;\,}|@{\;\,}c@{\;\,}|@{\;\,}c@{\;\,}|@{\;\,}c@{\;\,}|@{\;\,}c@{\;\,}|@{\;\,}c@{\;\,}|@{\;\,}c@{\;\,}|}
\hline
$N$ & $30$ & $50$ & $100$ & $200$ & $500$ & $1,000$ & $2,000$\\
\hline
\multicolumn{5}{l}{$p=2$}\\
\hline
CVX & $1.2 \times 10^0$ & $2.6 \times 10^0$ & $3.1 \times 10^1$ & $2.0 \times 10^2$ & $5.4 \times 10^3$ & --- & --- \\
ADMM &    $8.3 \times 10^{-3}$ & $7.5 \times 10^{-2}$ & $1.8 \times 10^{-1}$ & $2.5 \times 10^{0}$ & $3.6 \times 10^{0}$ & $2.4 \times 10^{1}$ & $8.3 \times 10^1$\\
\hline
\multicolumn{5}{l}{$p=\infty$}\\
\hline
CVX &  $4.3 \times 10^{0}$ & $1.5 \times 10^{1}$ & $2.5 \times 10^{2}$ & $9.1 \times 10^3$ & --- & --- & ---\\
ADMM & $4.0 \times 10^{-1}$ & $4.5 \times 10^{0}$ & $7.6 \times 10^{0}$ & $2.4 \times 10^{1}$ & $7.8 \times 10^{1}$ & $1.8 \times 10^{2}$ & $6.8 \times 10^{2}$ \\
\hline
\end{tabular}
\end{small}
\label{tab:compTime_muComparison}
\end{table*}
\begin{equation}
\label{eq:tracerow-matrix-equivalent}
\begin{split}
&\min_{\Z, \C} \; \lambda \| \Z \|_{1,p} + \tr(\D^{\top} \C) + \frac{\mu}{2}{\| \Z - \C \|_F^2} \\
&\st ~~ \1^{\top} \C = \1^{\top}, ~ \C \geq \0, ~ \Z = \C,
\end{split}
\end{equation}
where $\mu > 0$ is a penalty parameter. Notice that \eqref{eq:tracerow-matrix-1} and \eqref{eq:tracerow-matrix-equivalent} are equivalent, \ie, they find the same optimal solution for $\Z$. This comes from the fact that the last term in the objective function of \eqref{eq:tracerow-matrix-equivalent} vanishes for any feasible solution, since it satisfies $\Z = \C$. Augmenting the last equality constraint of \eqref{eq:tracerow-matrix-equivalent} to the objective function via the Lagrange multiplier matrix $\bLambda \in \Re^{M \times N}$, we can write the Lagrangian function \cite{BoydVandenberghe04} as
\begin{equation}
\label{eq:tracerow-Lagrangian1}
\begin{split}
\!\!\!\! \L &= \lambda \, \| \Z \|_{1,p} + \frac{\mu}{2}\,{\| \Z - (\C - \frac{\bLambda}{\mu}) \|_F^2} + h_1( \C, \bLambda)\\
    &= \sum_{i=1}^{M}{\!( \lambda \| \Z_{i*} \|_q \! + \frac{\mu}{2}\,{\| \Z_{i*} \! - \! (\C_{i*} \! - \! \frac{\bLambda_{i*}}{\mu}) \|_2^2} )} + h_1( \C, \bLambda),
\end{split}
\end{equation}
where $\A_{i*}$ denotes the $i$-th row of the matrix $\A$ and the term $h_1(\cdot)$ does not depend on $\Z$. We can rewrite the Lagrangian as
\begin{equation}
\label{eq:tracerow-Lagrangian2}
\begin{split}
\!\! \L &= \frac{\mu}{2} \, {\| \C - (\Z + \frac{\bLambda + \D}{\mu}) \|_F^2} + h_2( \Z, \bLambda)\\
    &= \sum_{i=1}^{N}{\frac{\mu}{2} \| \C_{*i} \! - (\Z_{*i} + \frac{\bLambda_{*i} + \D_{*i}}{\mu}) \|_2^2} + h_2( \Z, \bLambda)
\end{split}
\end{equation}
where $\A_{*i}$ denotes the $i$-th column of the matrix $\A$ and the term $h_2(\cdot)$ does not depend on $\C$. After initializing $\Z$, $\C$ and $\bLambda$, the ADMM iterations consist of 1) minimizing $\L$ with respect to $\Z$ while fixing other variables; 2) minimizing $\L$ with respect to $\C$ subject to the constraints $\{ \1^\top \C = \1^\top, \C \geq \0 \}$ while fixing other variables; 3) updating the Lagrange multiplier matrix $\bLambda$, having other variables fixed. Algorithm \ref{alg:ADMM} shows the steps of the ADMM implementation of the DS3 algorithm.\footnote{The infinite norm of a matrix, as used in the computation of the errors in Algorithm \ref{alg:ADMM}, is defined as the maximum absolute value of the elements of the matrix \ie, $\| \A \|_\infty = \max_{ij}{|a_{ij}}|$.}

\begin{algorithm}[t!]
\caption{\bf: DS3 Implementation using ADMM}
\label{alg:ADMM}
\textbf{Initialization:} Set $\mu = {10}^{-1}, \varepsilon = 10^{-7}$, maxIter = $10^5$. Initialize $k = 0, \Z^{(0)} = \C^{(0)} = \I, \boldsymbol{\Lambda}^{(0)} = \0$ and $\text{error}1 = \text{error}2 = 2 \varepsilon$.
\begin{algorithmic}[1]
\While{($\text{error}1 > \varepsilon$ or $\text{error}2 > \varepsilon$) and ($k < $ maxIter)}
\State Update $\Z$ and $\C$ by
\begin{equation*}
\Z^{(k+1)} = ~\argmin_{\Z} \; \frac{\lambda}{\mu} \| \Z \|_{1,p} + \frac{1}{2} \| \Z - (\C^{(k)} - \frac{\bLambda^{(k)}}{\mu}) \|_F^2;
\end{equation*}
\begin{equation*}
\begin{split}
\!\!\!\!\!\!\!\!\!\!\!\!\!\!\! \C^{(k+1)} =~&\argmin_{\C} \; \| \C - (\Z^{(k+1)} + \frac{\bLambda^{(k)} + \D}{\mu}) \|_F^2,\\
& ~~~ \st ~~~~ \1^\top \C = \1^\top,~\C \geq \0
\end{split}
\end{equation*}
\State Update the Lagrange multiplier matrix by 
\begin{equation*}
\bLambda^{(k+1)} = \bLambda^{(k)} + \mu \, (\Z^{(k+1)} - \C^{(k+1)});
\end{equation*}
\State Update errors by
\begin{equation*}
\begin{split}
\text{error}1 &= \| \Z^{(k+1)} - \C^{(k+1)} \|_\infty,\\
\text{error}2 &= \| \Z^{(k+1)} - \Z^{(k)} \|_\infty;
\end{split}
\end{equation*}
\State $k \, \leftarrow \, k + 1$;
\EndWhile
\end{algorithmic}
\textbf{Output:} Optimal solution $\Z^* = \Z^{(k)}$.
\end{algorithm}

Our implementation results in a memory and computational time complexity which are of the order of the number of elements in $\D$. In addition, it allows for parallel implementation, which can further reduce the computational time. More specifically,

\smallskip\noindent-- Minimizing the Lagrangian function in \eqref{eq:tracerow-Lagrangian1} with respect to $\Z$ can be done in $O(MN)$ computational time. We can obtain the solution in the case of $p=2$ via shrinkage and thresholding operation and in the case of $p=\infty$ via projection onto the $\ell_1$ ball \cite{Combettes:MMS05, Chaux:IP07}. Notice that we can perform the minimization in \eqref{eq:tracerow-Lagrangian1} via $M$ independent smaller optimization programs over the $M$ rows of $\Z$. Thus, having $P$ parallel processing resources, we can reduce the computational time to $O(\lceil {M}/{P} \rceil N)$.

\smallskip\noindent-- Minimizing the Lagrangian function in \eqref{eq:tracerow-Lagrangian2} with respect to $\C$ subject to the probability simplex constraints $\{ \1^\top \C = \1^\top, \; \C \geq \0 \}$ can be done using the algorithm in \cite{Duchi:ICML08} with $O(M \log(M) N)$ computational time ($O(M N)$ expected time using the randomized algorithm in \cite{Duchi:ICML08}). Notice that we can solve \eqref{eq:tracerow-Lagrangian2} via $N$ independent smaller optimization programs over the $N$ columns of $\C$. Thus, having $P$ parallel processing resources, we can reduce the computational time to $O(M  \log(M) \lceil {N}/{P} \rceil)$ (or $O(M \lceil {N}/{P} \rceil)$ expected time using the randomized algorithm in \cite{Duchi:ICML08}).

\smallskip\noindent-- The update on $\bLambda$ has $O(MN)$ computational time and can be performed, respectively, by $M$ or $N$ independent updates over rows or columns, hence having $O(\lceil {M}/{P} \rceil N)$ or $O(M \lceil {N}/{P} \rceil)$ computational time when using $P$ parallel processing resources.

As a result, the proposed ADMM implementation of our algorithm can be performed in $O(M \log(M) N)$ computational time, while we can reduce the computational time to $O(\lceil M N / P \rceil \log(M))$ using $P$ parallel resources. This provides significant improvement with respect to standard convex  solvers, such as CVX \cite{cvx}, which typically have cubic or higher complexity in the problem size.

{Table \ref{tab:compTime_muComparison} shows the average computational time of CVX (Sedumi solver) and our proposed ADMM-based framework (serial implementation) over $100$ randomly generated datasets of varying size on an X86--64 server with 1.2 GHz CPU and 132 GB memory.} Notice that for both $p=2$ and $p=\infty$, the ADMM approach is significantly faster than CVX. In fact, while for a dataset of size $N = 100$, CVX runs out of memory and time, our ADMM framework runs~efficiently.

\section{Theoretical Analysis}
\label{sec:theory}
In this section, we study theoretical guarantees of the DS3 algorithm. We consider our proposed optimization in \eqref{eq:tracerow-matrix-1} and, first, study the effect of the regularization parameter, $\lambda$, on the solution. Second, we show that when there exists a joint partitioning of $\X$ and $\Y$, based on dissimilarities, DS3 finds representatives from all partitions of $\X$ and, at the same time, reveals the clustering of the two sets. We also discuss the special yet important case where source and target sets are identical and discuss implications of our theoretical results for proving the integrality of convex program for clustering. It is important to mention that, motivated by our work, there has been a series of interesting results showing the integrality of the alternative formulation in \eqref{eq:dsmrs-v2} for clustering under specific assumptions on dissimilarities and $p$ \cite{Nellore:TechRep14, Awasthi:ITCS15}.

\begin{figure*}[t!]
\centering
\includegraphics[width=0.17\linewidth, trim = 5 0 3 0 , clip]{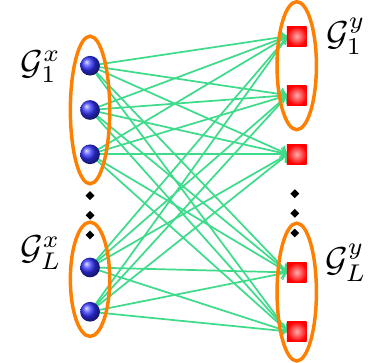}\hspace{17mm}
\includegraphics[width=0.17\linewidth, trim = 5 0 3 0 , clip]{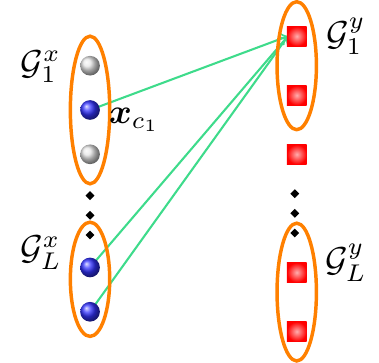}\hspace{17mm}
\includegraphics[width=0.17\linewidth, trim = 5 0 3 0 , clip]{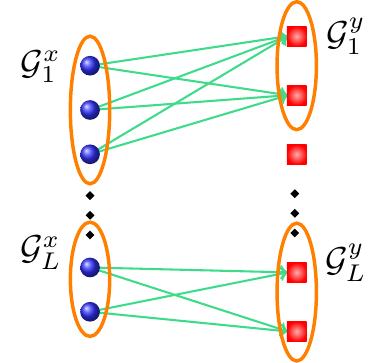}
\caption{\small{Illustration of our theoretical result for clustering. Left: we assume a joint partitioning of source and target sets into $L$ groups, $\{ (\G^x_k, \G^y_k) \}_{k=1}^{L}$. Middle: we assume that the medoid of each $\G^x_k$ better represents $\G^y_k$ than other partitions $\G^x_{k'}$ for $k' \neq k$. Right: Our optimization in \eqref{eq:tracerow-matrix-1} selects representatives from all source set partitions and each partition in the target set only get represented by the corresponding source~set~partition.}}
\label{fig:dsrsTheory}
\end{figure*}
%

\subsection{Regularization Parameter Effect}
The regularization parameter in \eqref{eq:tracerow-matrix-1} puts a trade-off between two opposing terms: the number of representatives and the encoding cost via representatives. In other words, we obtain a smaller encoding cost by selecting more  representatives and vice versa. As we increase the value of $\lambda$ in \eqref{eq:tracerow-matrix-1}, we put more emphasis on penalizing the number of representatives compared to the encoding cost, hence, we expect to obtain fewer representatives. In fact, we show that when $\lambda$ is larger than a certain threshold, which we determine using dissimilarities, we obtain only one representative. More specifically, we prove the following result (the proofs of all theoretical results are provided in the supplementary materials).
\vspace{1.5mm}
\begin{theorem}
\label{thm:max-lambda}
\emph{
Consider the optimization program \eqref{eq:tracerow-matrix-1}. Let $\ell^* \triangleq \argmin_{i} \1^{\top} \d_i$ and 
\begin{equation}
\label{eq:max-lambda}
\begin{split}
&\lambda_{\max,2} \;\; \triangleq \, \max_{i \neq \ell} \, \frac{\sqrt{N}}{2} \cdot \frac{\| \d_i - \d_{\ell^*} \|_2^2}{\1^{\top} (\d_i - \d_\ell^*)}, \\ &\lambda_{\max,\infty} \, \triangleq \, \max_{i \neq \ell} \, \frac{\| \d_i - \d_{\ell^*} \|_1}{2} .
\end{split}
\end{equation}
For $p \in \{2,\infty\}$, if $\lambda \geq \lambda_{\max,p}$, the solution of \eqref{eq:tracerow-matrix-1} is $\Z^* = \e_{\ell^*} \1^\top$, where $\e_{\ell^*}$ denotes a vector whose $\ell^*$-th element is one and other elements are zero. Thus, for $\lambda \geq \lambda_{\max,p}$, the solution of \eqref{eq:tracerow-matrix-1} corresponds to selecting $\x_{\ell^*}$ as the representative of $\Y$.
}
\end{theorem}
\vspace{1.5mm}
Notice that the threshold value $\lambda_{\max,p}$ is, in general, different for $p=2$ and $p = \infty$. However, in both cases, we obtain the same representative, $\x_{\ell^*}$, which is the element of $\X$ that has the smallest sum of dissimilarities to elements of $\Y$. For instance, when $\X = \Y$ correspond to data points and dissimilarities are computed using the Euclidean distance, the single representative corresponds to the data point that is closest to the geometric median \cite{Wesolowsky:LS93} of the dataset, as shown in the right plot of Figure \ref{fig:2G-PZ}.

As we decrease the value of $\lambda$ in \eqref{eq:tracerow-matrix-1}, we put more emphasis on minimizing the encoding cost of $\Y$ via representatives compared to the number of representatives. In the limiting case where $\lambda$ approaches an arbitrarily small nonnegative value, we obtain the minimum encoding cost in \eqref{eq:tracerow-matrix-1}, where for every $\y_j$ we have
\begin{equation}
z_{i^*_jj} = 1, \quad i^*_j \triangleq \argmin_i d_{ij}.
\end{equation}
In other words, each element of $\Y$ selects the closest element of $\X$ as its representative.

\begin{figure*}[t!]
\centering
\includegraphics[width=0.27\linewidth, trim = 5 0 0 0 , clip]{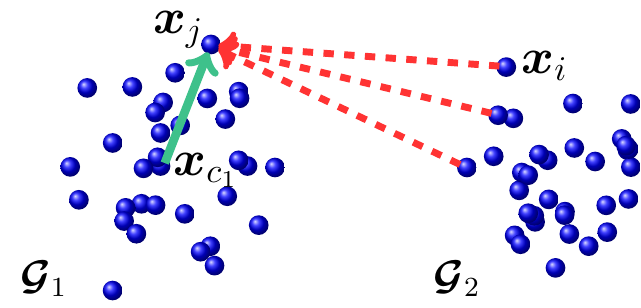}\ \hspace{8mm}
\includegraphics[width=0.27\linewidth, trim = 5 0 0 0 , clip]{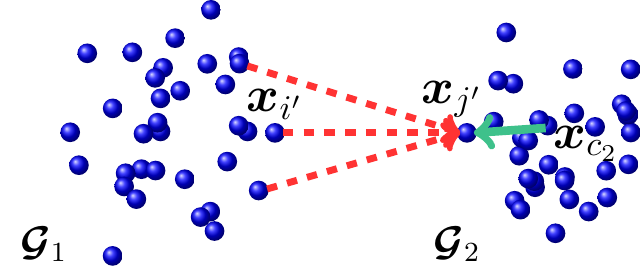}\ \hspace{10mm}
\includegraphics[width=0.27\linewidth, trim = 5 0 0 3 , clip]{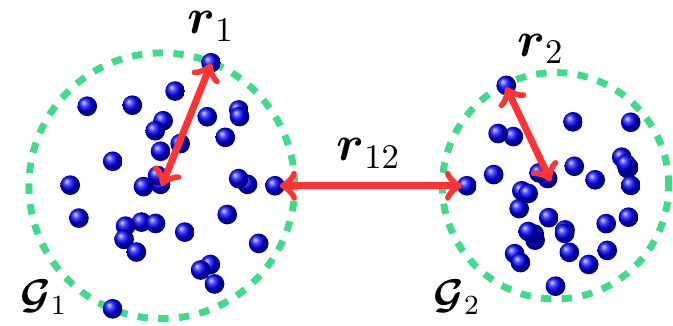}
\caption{\small{Left and middle plots: The dataset partitions into groups $\G_1$ and $\G_2$, according to Definition \ref{def:clustering}, if  1) for every $\x_j$ with $j$ in $\G_1$, the dissimilarity to $\x_{c_1}$ is smaller than the dissimilarity to any $\x_i$ with $i$ in $\G_2$ (left plot); 2) for every $\x_{j'}$ with $j'$ in $\G_2$, the distance to $\x_{c_2}$ is smaller than the distance to any $\x_{i'}$ with $i'$ in $\G_1$ (middle plot). In such a case, our proposed algorithm selects representatives from all $\G_i$'s and points in each group will be represented only by representatives from the same group. Right plot: A sufficient condition on the regularization parameter to reveal the clustering is to have $\lambda < r_{12} - \max\{ r_1, r_2 \}$.}}
\label{fig:2grps}
\end{figure*}
%

\subsection{Clustering Guarantees}
In this section, we investigate clustering guarantees of our proposed algorithm. We show that when $\X$ and $\Y$ jointly partition into multiple groups, in the solution of our proposed optimization in \eqref{eq:tracerow-matrix-1}, elements in each partition of $\Y$ select their representatives from the corresponding partition of $\X$. This has the important implication that all groups in $\X$ will be sampled. To better illustrate the notion of joint partitioning of $\X$ and $\Y$, we consider the following example.

\vspace{1.5mm}
\begin{example}
Let $\X = \{ \x_1, \ldots, \x_M \}$ be a set of models and $\Y = \{ \y_1, \ldots, \y_N \}$ be a set of data points. Assume $\G^x_1$ denotes indices of the first $q$ models, which efficiently represent the first $q'$ data points, indexed by $\G^y_1$, but have infinite dissimilarities to the rest of $N-q'$ data points, indexed by $\G^y_2$. Similarly, assume $\G^x_2$ denotes indices of the rest of $M-q'$ models, which efficiently represent data points indexed by $\G^y_2$, but have infinite dissimilarities to data points in $\G^y_1$. As a result, the solution of the optimization \eqref{eq:tracerow-matrix-1} will have the form
\begin{equation}
\Z^* = \begin{bmatrix} \Z_1^* & \0 \\ \0 & \Z_2^* \end{bmatrix},
\end{equation}
where $\Z_1^* \in \Re^{q \times q'}$ and $\Z_2^* \in \Re^{M-q \times N-q'}$ have a few nonzero rows. In this case, we say that $\X$ and $\Y$ jointly partition into two groups $( \G^x_1, \G^y_1 )$ and $( \G^x_2, \G^y_2 )$, where elements of $\Y$ indexed by $\G^y_k$, denoted by $\Y(\G^y_k)$, choose their representatives from elements of $\X$ indexed by $\G^x_k$, denoted by $\X(\G^x_k)$, for~$k = 1, 2$.
\end{example}
\vspace{1.5mm}

Formalizing the notion of the joint partitioning of $\X$ and $\Y$ into $L$ groups $\{ (\G^x_k, \G^y_k) \}_{k = 1}^{L}$, we prove that in the solution of \eqref{eq:tracerow-matrix-1}, each partition $\G^y_k$ selects representatives from the corresponding partition $\G^x_k$. To do so, we first introduce the notions of \emph{dissimilarity radius} of $( \G^x_k, \G^y_k )$ and the \emph{medoid} of $\G^x_k$.

\vspace{1.5mm}
%
\begin{definition}
\label{def:radius}
\emph{
Let $\G^x_k \subseteq \{1, \ldots, M\}$ and $\G^y_k \subseteq \{1, \ldots, N\}$.  
We define the dissimilarity-radius associated with $( \G^x_k, \G^y_k )$ as
\begin{equation}
r_k \, = \, r(\G^x_k, \G^y_k) \, \triangleq \, \min_{i \in \G^x_k} \; \max_{j \in \G^y_k}\;{d_{ij}}.
\end{equation}
We define the medoid of $\G^x_k$, denoted by $c_k$, as the element of $\G^x_k$ for which we obtain the dissimilarity radius, \ie,
\begin{equation}
c_k \, = \, c(\G^x_k, \G^y_k) \, \triangleq \, \argmin_{i \in \G^x_k} \; ( \, \max_{j \in \G^y_k}\;{d_{ij}} \, ).
\end{equation}
}
\end{definition}
%
\vspace{1.5mm}
In other words, $c_k$ corresponds to the element of $\X(\G^x_k)$ whose maximum dissimilarity to $\Y(\G^y_k)$ is minimum. Also, $r_k$ corresponds to the maximum dissimilarity of $c_k$ to $\Y(\G^y_k)$. Next, we define the notion of the joint partitioning of $\X$ and $\Y$.
\vspace{1.5mm}
\begin{definition}
\label{def:clustering}
\emph{
Given pairwise dissimilarities $\{ d_{ij} \}$ between $\X$ and $\Y$, we say that $\X$ and $\Y$ jointly partition into $L$ groups $\{ (\G^x_k,\G^y_k) \}_{k=1}^{L}$, if for each $\y_j$ with $j$ in $\G^y_k$, the dissimilarity between $\x_{c_k}$ and $\y_j$ is strictly smaller than the minimum dissimilarity between $\y_j$ and all partitions other than $\G^x_{k}$, \ie,
\begin{equation}
d_{c_k j} \, < \, \min_{k' \neq k} \, \min_{i \in \G^x_{k'}} \, d_{i j}, \quad \forall k=1, \ldots, L, \; \forall j \in \G_k^y.
\end{equation}
}
\end{definition}
\vspace{1.5mm}
Next, we show that if $\X$ and $\Y$ jointly partition into $L$ groups, then for a suitable range of the regularization parameter that we determine, $\Y(\G^y_k)$ selects its representatives from $\X(\G^x_k)$. Figure \ref{fig:dsrsTheory} illustrates our partitioning definition and theoretical results. 
\vspace{1.5mm}
%
\begin{theorem}
\label{thm:clustering-lambda}
\emph{
Given pairwise dissimilarities $\{ d_{ij} \}$, assume that $\X$ and $\Y$ jointly partition into $L$ groups $\{ (\G^x_k,\G^y_k) \}_{k=1}^{L}$ according to Definition \ref{def:clustering}. Let $\lambda_g$ be defined as
\begin{equation}
\label{eq:clustering-lambda1}
\lambda_g \, \triangleq \, \min_{k} \, \min_{j \in \G^y_{k}} \, (\min_{k' \neq k} \, \min_{i \in \G^x_{k'}} \, d_{i j} - d_{c_k j}).
\end{equation}
Then for $\lambda < \lambda_g\,$ in the optimization \eqref{eq:tracerow-matrix-1}, elements of $\Y(\G_k^y)$ select their representatives from $\X(\G_k^x)$, for all $k = 1, \ldots, L$.
}
\end{theorem}
%
\vspace{1.5mm}
\begin{remark}
\label{rem:1rep-cluster}
From Theorem \ref{thm:max-lambda} and \ref{thm:clustering-lambda} we can show that, under appropriate conditions, each $\Y(\G_k^y)$ will be represented via a single element from $\X(\G_k^x)$. More specifically, if $\lambda_{\max,p}(\G_k^x,\G_k^y)$ denotes the threshold value on $\lambda$ above which we obtain a single representative from $\X(\G_k^x)$ for $\Y(\G_k^y)$, then for $\max_{k} \lambda_{\max,p}(\G_k^x,\G_k^y) \leq \lambda < \lambda_{g}$, assuming such an interval is nonempty, each $\Y(\G_k^y)$ selects one representative from $\X(\G_k^x)$, corresponding to its medoid. Hence, we obtain $L$ representatives, which correctly cluster the data into the $L$ underlying groups.
\end{remark}
%

\subsection{Identical Source and Target Sets}
The case where source and target sets are identical forms an important special case of our formulation, which has also been the focus of state-of-the-art algorithms \cite{Taskar:ICML11, Frey:Science07, Taskar:AISTATS13, Nellore:TechRep14, Awasthi:ITCS15}. Here, one would like to find representatives of a dataset given pairwise relationships between points in the dataset.

\vspace{1.5mm}
%
\begin{assumption}
When $\X$ and $\Y$ are identical, we assume that $d_{jj} < d_{ij}$ for every $j$ and every $i \neq j$, \ie, we assume that each point is a better representative for itself than other points.
\end{assumption}
%
\vspace{1.5mm}
It is important to note that our theoretical analysis in the previous sections also applies to this specific setting. Hence, when there exists a grouping of the dataset, our convex formulation has clustering theoretical guarantees. In particular, the result in Theorem \ref{thm:clustering-lambda} provides clustering guarantees in the nontrivial regime where points from different groups may be closer to each other than points from the same group, \ie, in the regime where clustering by thresholding pairwise dissimilarities between points fails.

\vspace{1.5mm}
\begin{example} 
Consider the dataset shown in Figure \ref{fig:2grps}, where points are gathered around two clusters $\G_1$ and $\G_2$, with medoids $\x_{c_1}$ and $\x_{c_2}$, respectively. Let the dissimilarity between a pair of points be their Euclidean distance. In order for the dataset to partition into $\G_1$ and $\G_2$ according to Definition \ref{def:clustering}, for every $\x_j$ with $j$ in $\G_k$, the distance between $\x_j$ and $\x_{c_k}$ must be smaller than the distance between $\x_j$ and any $\x_i$ in $\G_{k'}$ with $k' \neq k$, as shown in the left and middle plots of Figure \ref{fig:2grps}. In this case, it is easy to verify that for $\lambda < r_{12} - \max\{r_1,r_2\}$, with $r_i$ being the radius of each cluster and $r_{ij}$ being the distance between two clusters as shown in the right plot of Figure \ref{fig:2grps}, the clustering condition and result of Theorem \ref{thm:clustering-lambda} holds.
\end{example}
\vspace{1.5mm}

In the case where $\X$ and $\Y$ are identical, when the regularization parameter in \eqref{eq:tracerow-matrix-1} becomes sufficiently small, each point becomes a representative of itself, \ie, $z_{ii} = 1$ for all $i$. In other words, each point forms its own cluster. In fact, using Theorem \ref{thm:clustering-lambda}, we obtain a threshold $\lambda_{\min}$ such that for $\lambda \leq \lambda_{\min}$, the optimal solution of \eqref{eq:tracerow-matrix-1} becomes the identity matrix.

\vspace{1.5mm}
%
\begin{corollary}
Assume $\X = \Y$ and define $\lambda_{\min} \triangleq \min_{j} (\min_{i \neq j}{d_{ij}} - d_{jj})$. 
For $\lambda \leq \lambda_{\min}$ and $p \in \{2,\infty\}$, the solution of the optimization program \eqref{eq:tracerow-matrix-1} is the identity matrix, \ie, each point becomes a representative of itself.
\end{corollary}
%
\vspace{1.5mm}
\begin{figure*}[t!]
\centering
\hspace{-.8mm}\includegraphics[width=0.103\linewidth, trim =  0 0 0 0, clip]{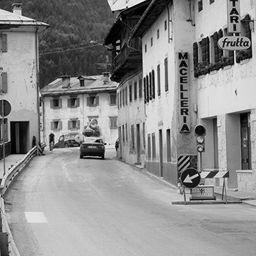} \hspace{-1mm}
\includegraphics[width=0.103\linewidth, trim =  0 0 0 0, clip]{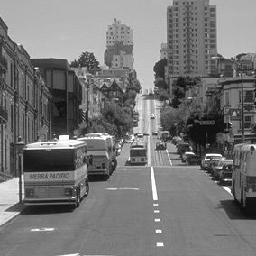} \hspace{-1mm}
\includegraphics[width=0.103\linewidth, trim =  0 0 0 0, clip]{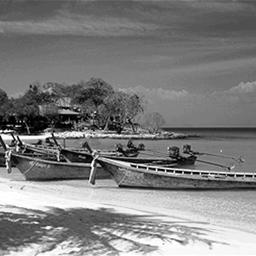} \hspace{-1mm}
\includegraphics[width=0.103\linewidth, trim =  0 0 0 0, clip]{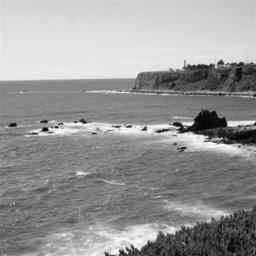} \hspace{-1mm}
\includegraphics[width=0.103\linewidth, trim =  0 0 0 0, clip]{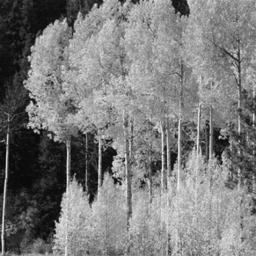} \hspace{-1mm}
\includegraphics[width=0.103\linewidth, trim =  0 0 0 0, clip]{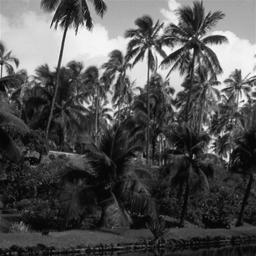} \hspace{-1mm}
\includegraphics[width=0.103\linewidth, trim =  0 0 0 0, clip]{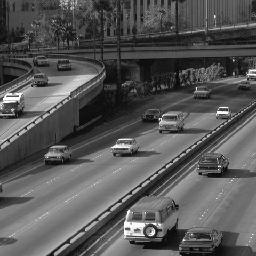} \hspace{-1mm}
\includegraphics[width=0.103\linewidth, trim =  0 0 0 0, clip]{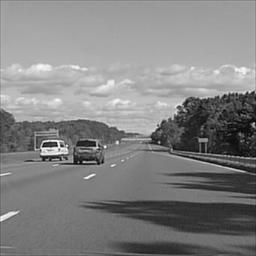} \hspace{-1mm}
\includegraphics[width=0.103\linewidth, trim =  0 0 0 0, clip]{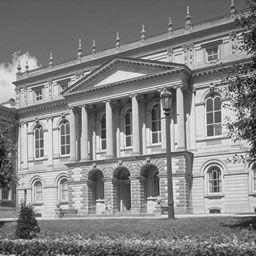}\\\vspace{1mm}
\hspace{.05mm}\includegraphics[width=0.103\linewidth, trim =  0 0 0 0, clip]{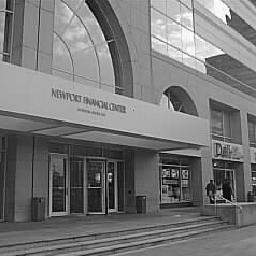} \hspace{-.9mm}
\includegraphics[width=0.103\linewidth, trim =  0 0 0 0, clip]{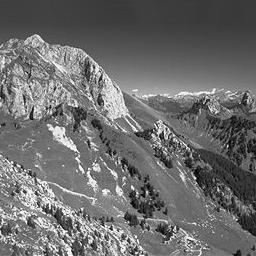} \hspace{-1mm}
\includegraphics[width=0.103\linewidth, trim =  0 0 0 0, clip]{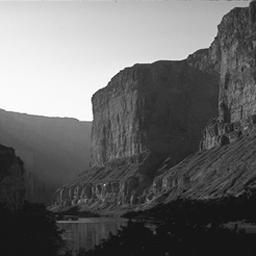} \hspace{-.9mm}
\includegraphics[width=0.103\linewidth, trim =  0 0 0 0, clip]{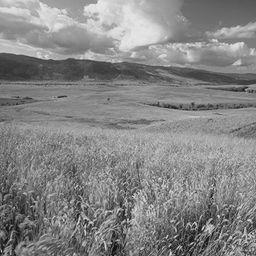} \hspace{-1mm}
\includegraphics[width=0.103\linewidth, trim =  0 0 0 0, clip]{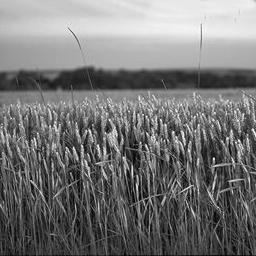} \hspace{-.9mm}
\includegraphics[width=0.103\linewidth, trim =  0 70 0 40, clip]{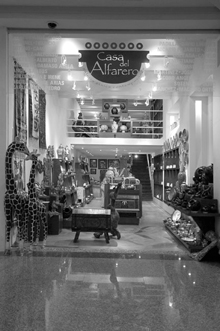} \hspace{-.9mm}
\includegraphics[width=0.103\linewidth, trim =  0 70 0 40, clip]{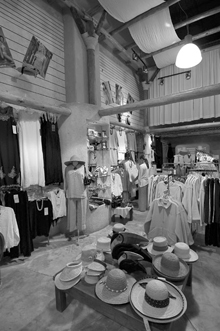} \hspace{-.9mm}
\includegraphics[width=0.103\linewidth, trim =  0 0 0 0, clip]{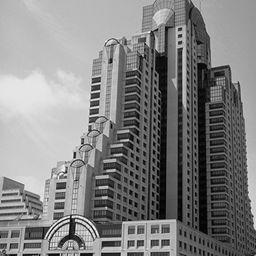} \hspace{-.9mm}
\includegraphics[width=0.103\linewidth, trim =  0 0 0 0, clip]{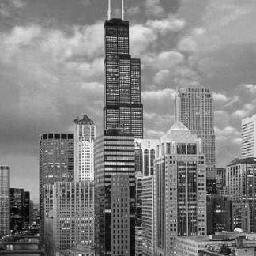} \hspace{-1.5mm}
\vspace{.1mm} 
\caption{\footnotesize{We demonstrate the effectiveness of our proposed framework on the problem of scene categorization via representatives. We use the Fifteen Scene Categories dataset \cite{Lazebnik:CVPR06}, a few of its images are shown. The dataset contains images from 15 different categories of street, coast, forest, highway, building, mountain, open country, store, tall building, office, bedroom, industrial, kitchen, living room, and suburb.}}
\label{fig:15scene-samples}
\vspace{0mm}
\end{figure*}

\vspace{0mm}
\section{Experiments}
\vspace{0mm}
\label{sec:experiments}
In this section, we evaluate the performance of our proposed algorithm for finding representatives. We consider the two problems of nearest neighbor classification using representative samples and dynamic data modeling and segmentation using representative models. We evaluate the performance of our algorithm on two real-world datasets and show that it significantly improves the state of the art and addresses several existing challenges. 

Regarding the implementation of DS3, since multiplying $\D$ and dividing $\lambda$ by the same scalar does not change the solution of \eqref{eq:tracerow-matrix-1}, in the experiments, we scale the dissimilarities to be in $[0,1]$ by dividing $\D$ by its largest entry. Unless stated otherwise, we typically set $\lambda = \alpha \lambda_{\max,p}$ with $\alpha \in [0.01,0.5]$, for which we obtain good results. We only report the result for $p=\infty$, since we obtain similar performance for $p = 2$.

\vspace{0mm}

\subsection{Classification using Representatives}

We consider the problem of finding prototypes for the nearest neighbor (NN) classification \cite{Duda:04}. {Finding representatives, which capture the distribution of data, not only helps to significantly reduce the computational cost and memory requirements of the NN classification at the testing time, but also, as demonstrated here, maintains or even improves the performance. }

\subsubsection{Scene Categorization}
To investigate the effectiveness of our proposed method for finding prototypes for classification, we consider the problem of scene categorization from images. We use the Fifteen Scene Categories dataset  \cite{Lazebnik:CVPR06} that consists of images from $K=15$ different classes, such as coasts, forests, highways, mountains, stores, and more, as shown in Figure \ref{fig:15scene-samples}. There are between $210$ and $410$ images in each class, making a total of $4,485$ images in the dataset. We randomly select $80\%$ of images in each class to form the training set and use the rest of the $20\%$ of images in each class for testing. We find representatives of the training data in each class and use them as a reduced training set to perform NN classification on the test data. We compare our proposed algorithm with AP \cite{Frey:Science07}, Kmedoids \cite{Kaufman:TR87}, and random selection of data points (Rand) as the baseline. Since Kmedoids depends on initialization, we run the algorithm 1,000 times with different random initializations and use the result that obtains the lowest energy. To have a fair comparison, we run all algorithms so that they obtain the same number of representatives. For each image, we compute the spatial pyramid histogram \cite{Lazebnik:CVPR06}, as the feature vector, using $3$ pyramid levels and $200$ bins.

After selecting $\eta$ fraction of training samples in each class using each algorithm, we compute the average NN classification accuracy on test samples, denoted by $\text{accuracy}(\eta)$, and report 

\begin{equation}
\label{eq:nnerror}
\text{err}(\eta) = \text{accuracy}(1) - \text{accuracy}(\eta),
\end{equation}
where $\text{accuracy}(1)$ is the NN classification accuracy using all training samples in each class.

Table \ref{tab:15Scene-errors-chi} show the performance of different algorithms on the dataset as we change the fraction of representatives, $\eta$, selected from each class for $\chi^2$ distance dissimilarities. 
As the results show, increasing the value of $\eta$, \ie, having more representatives from each class, improves the classification results as expected. Rand and Kmedoids do not perform well, with Kmedoids suffering from dependence on a good initialization. On the other hand, DS3, in general, performs better than other methods. Notice that AP relies on a message passing algorithm, which solves the problem approximately when the graph of pairwise relationships is not a tree \cite{Koller:book09}, including our problem. Notice also that by selecting only $35\%$ of the training samples in each class, the performance of DS3 is quite close to the case of using all training samples, only $2.9\%$ worse. 

It is important to notice that the performance of all methods depends on the choice of  dissimilarities. In other words, dissimilarities should capture the distribution of data in a way that points from the same group have smaller dissimilarities than points in different groups. In fact, using the $\chi^2$ dissimilarity instead of Euclidean distances results in improving the classification performance of all algorithms by about $16\%$, as shown in the supplementary materials.

\begin{table}[t!]
\caption{\footnotesize Errors ($\%$) of different algorithms, computed via \eqref{eq:nnerror}, as a function of the fraction of selected samples from each class ($\eta$) on the 15~Scene~Categories dataset using $\chi^2$ distances.} \centering
\begin{small}
\begin{tabular}{|@{\;\,}c@{\;\,}|@{\;\,}c@{\;\,}|@{\;\,}c@{\;\,}|@{\;\,}c@{\;\,}|@{\;\,}c@{\;\,}|}
\hline
Algorithm & \;\; Rand \;\; & \, Kmedoids \, & \;\;\; AP\;\;\; & \;\;\; DS3 \;\;\;  \\
\hline
\hline
$\eta = 0.05$ &    $22.12$ & $14.42$ & $\textcolor{black}{\textbf{11.59}}$ & $12.04$ \\
\hline
\hline
$\eta = 0.10$ &  $15.54$ & $11.30$ & $7.91$ & $\textcolor{black}{\textbf{5.69}}$\\
\hline
\hline
$\eta = 0.20$ &  $11.97$ & $12.19$ & $6.01$ & $\textcolor{black}{\textbf{3.35}}$\\
\hline
\hline
$\eta = 0.35$ &  $7.18$ & $7.51$ & $6.46$ & $\textcolor{black}{\textbf{2.90}}$\\
\hline
\end{tabular}
\end{small}
\label{tab:15Scene-errors-chi}
\vspace{0mm}
\end{table}
\begin{figure*}[t!]
\centering
\begin{subfigure}[b]{0.29\textwidth}
\includegraphics[width=\textwidth, trim =  138 3 204 50, clip]{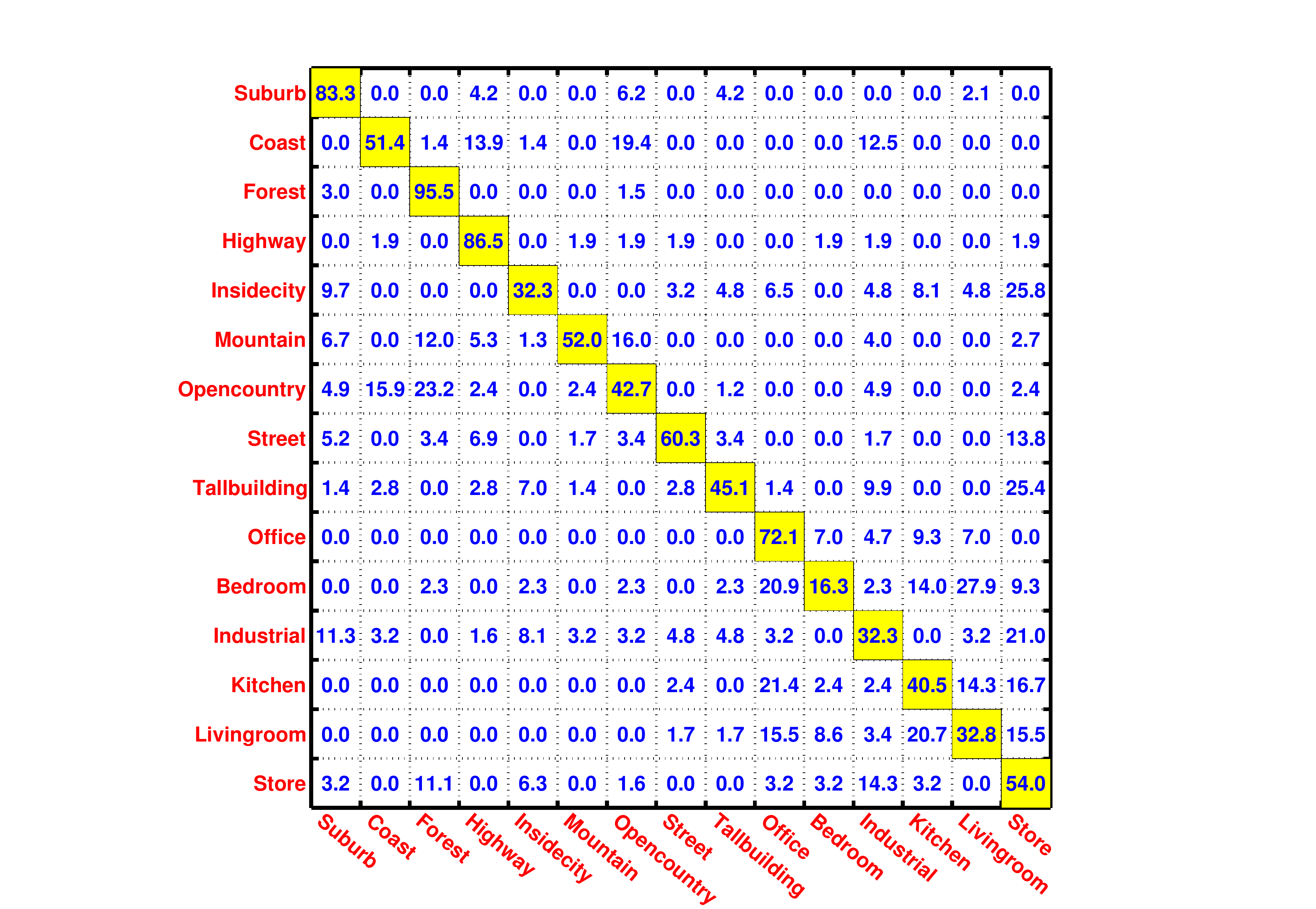}
\vspace{-4.5mm}
\caption{~$\eta = 0.05$}
\end{subfigure}
\hspace{2mm}
\begin{subfigure}[b]{0.29\textwidth}
\includegraphics[width=\textwidth, trim =  138 3 204 50, clip]{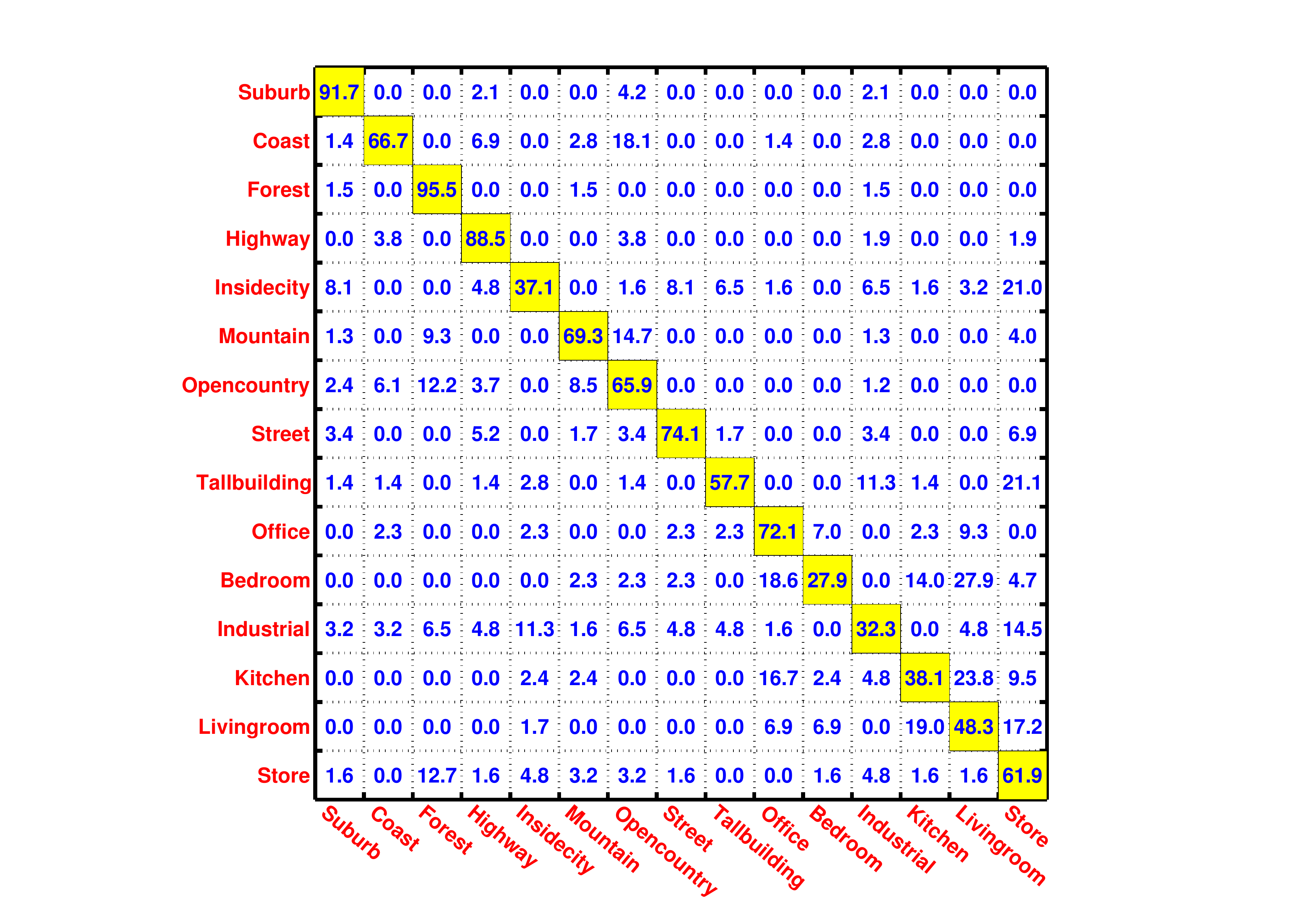}
\vspace{-4.5mm}
\caption{~$\eta = 0.35$}
\end{subfigure}
\hspace{2mm}
\begin{subfigure}[b]{0.29\textwidth}
\includegraphics[width=\textwidth, trim =  138 3 204 50, clip]{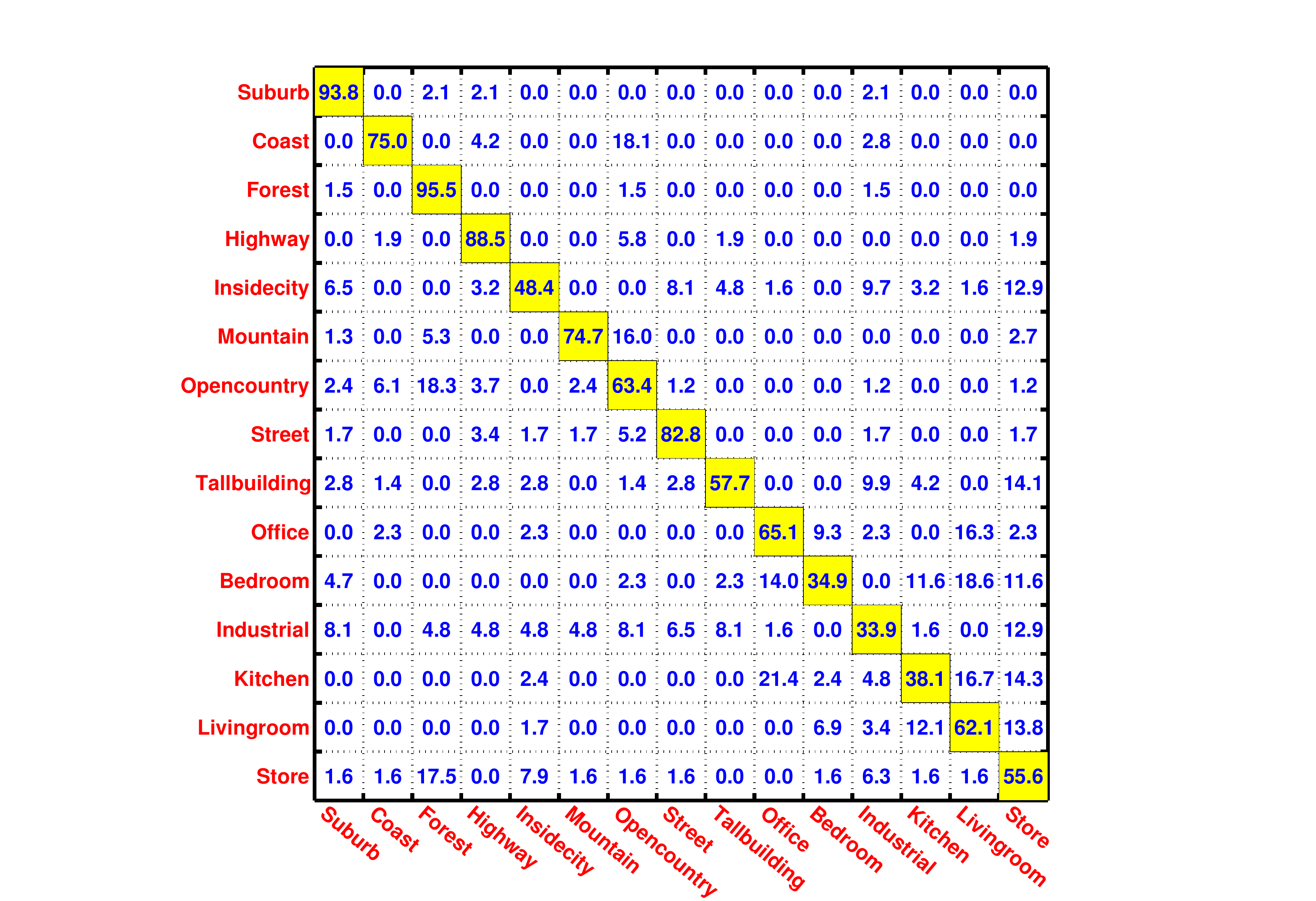}
\vspace{-4.5mm}
\caption{~$\eta = 1.00$}
\end{subfigure}
\vspace{-1mm} 
\caption{\footnotesize{Nearest Neighbor confusion matrix for the performance of the DS3 algorithm on the 15~Scene~Categories dataset for several values of the fraction of the training samples ($\eta$) selected from each class.}}
\label{fig:15scene-dsrmsConfMat}
\vspace{0mm}
\end{figure*}

Figure \ref{fig:15scene-dsrmsConfMat} shows the confusion matrix of the NN classifier using $\eta = 0.05$ and $\eta = 0.35$ of the training samples in each class obtained by DS3 (left and middle plots) and using $\eta = 1$ (right plot). Notice that as expected, increasing $\eta$, results in a closer confusion matrix to the case of using all training samples. More importantly, as the confusion matrices show, an important advantage of selecting prototypes is that the classification performance can even improve over the case of using all training samples. For instance, the recognition performance for the classes `store,' `office' and  `opencountry' improves when using representatives ($\eta = 0.35$). In particular, as the last row of the confusion matrices show, while using all training samples we obtain $55.6\%$ accuracy for classifying test images of the class `store,' we obtain $61.9\%$ accuracy using $\eta = 0.35$. This is due to the fact that by finding representatives, we remove samples that do not obey the distribution of the given class and are closer to other classes.

\subsubsection{Initializing Supervised Algorithms via DS3}
{It is important to notice that DS3 as well as AP and Kmedoids do not explicitly take advantage of the known labels of the training samples to minimize the classification error while choosing samples from each class. Extending DS3 to such a supervised setting is the subject of our current research. However, we show that using DS3 for initialization of one such supervised algorithm can improve the performance. More specifically, we use the Stochastic Neighbor Compression (SNC) algorithm \cite{Weinberger:ICML14}, where we initialize the method using $\eta$ fraction of samples chosen uniformly at random (SNC) versus chosen by DS3 (DS3 + SNC). As the results in Table \ref{tab:15Scene-errors-chi-2} show, running SNC with random initialization slightly outperforms DS3 due to its taking advantage of the known class labels. However, SNC initialized using the solution of DS3 not only performs better than SNC, but also achieves $2.53\%$ higher classification accuracy than NN using all training samples, demonstrating the importance of using representatives and also incorporating data distribution while minimizing the classification error of representatives.

\begin{table}[t!]
\caption{\footnotesize {Errors ($\%$) of DS3, SNC with random initialization and SNC initialized with solution of DS3, computed via \eqref{eq:nnerror}, as a function of the fraction of selected samples from each class ($\eta$) on the 15~Scene~Categories dataset.}} \centering
\begin{small}
\begin{tabular}{|@{\;\,}c@{\;\,}|@{\;\,}c@{\;\,}|@{\;\,}c@{\;\,}|@{\;\,}c@{\;\,}|@{\;\,}c@{\;\,}|}
\hline
Algorithm & \;\; $\eta = 0.05$ \;\; & $\eta = 0.10$ & \;\;\; $\eta = 0.20$ \;\;\; & \, $\eta = 0.35$ \\
\hline
\hline
DS3 &    $12.04$ & $5.69$ & $3.35$ & $2.90$ \\
\hline
\hline
SNC &  $10.01$ & $4.30$ & $3.21$ & $1.62$ \\
\hline
\hline
DS3 + SNC &  $\textcolor{black}{\textbf{8.52}}$ & $\textcolor{black}{\textbf{2.31}}$ & $\textcolor{black}{\textbf{0.05}}$ & $\textcolor{black}{\textbf{-2.53}}$ \\
\hline
\end{tabular}
\end{small}
\label{tab:15Scene-errors-chi-2}
\vspace{0mm}
\end{table}
\begin{figure*}[t!]
\centering
\includegraphics[width=0.1770\linewidth]{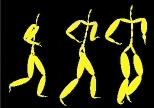} \hspace{-1.3mm}
\includegraphics[width=0.1395\linewidth]{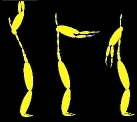} \hspace{-1.3mm}
\includegraphics[width=0.1970\linewidth]{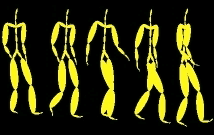} \hspace{-1.3mm}
\includegraphics[width=0.1855\linewidth]{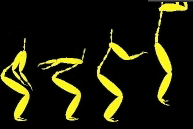} \hspace{-1.3mm}
\includegraphics[width=0.2710\linewidth]{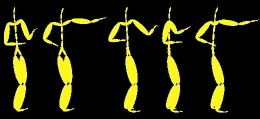} \hspace{-1.3mm}
\vspace{-.2mm} 
\caption{\footnotesize{We demonstrate the effectiveness of our proposed framework on the temporal segmentation of human activities. We use CMU motion capture dataset \cite{MoCap:12}. The dataset contains 149 subjects performing several activities. The motion capture system uses 42 markers per subject. We consider the data from subject 86 in the dataset, consisting of 14 different trials. Each trial comprises multiple activities such as `walk,' `squat,' `run,' `stand,' `arm-up,' `jump,' `drink,' `punch,' `stretch,' etc.}}
\label{fig:CMUmocap-samples}
\vspace{0mm}
\end{figure*}
%

\subsection{Modeling and Segmentation of Dynamic Data}
\label{sec:dynamicdataclustering}
In this section, we consider the problem of modeling and segmentation of time-series data generated by switching among dynamical systems. This problem has important applications, such as learning and segmentation of human activities in videos and motion capture data, learning nonlinear dynamic models and inverse modeling of complex motor control systems. We show that our framework can be used to robustly learn nonlinear dynamical systems and segment time-series data.

\subsubsection{Learning Switching Dynamical Models}
Assume that we have a time-series trajectory $\{ \q(t) \in \Re^p \}_{t=1}^{T}$ that is generated by a mixture of $K$ different models with parameters $\{ \bbeta_i \}_{i=1}^{K}$. We denote the switching variable by $\sigma_t \in \{1, \ldots, K\}$, where $K$ corresponds to the number of models. Two important instances of special interest are the state--space and the input/output switched models. In the state--space model, we have
\begin{equation}
\label{eq:pwa-ss}
\begin{split}
\z(t+1) &= \A_{\sigma_t} \z(t) + \g_{\sigma_t} + \v(t),\\
\q(t) &= \C_{\sigma_t} \z(t) + \h_{\sigma_t} + \bvarep(t),
\end{split}
\end{equation}
where $\z(t) \in \Re^n$ is the state of the system and $\v(t)$ and $\bvarep(t)$ denote the process and measurement errors, respectively. In this case, the model parameters are $\bbeta_i \triangleq \{ \A_i, \B_i, \g_i, \h_i \}$. In the input/output model, we have
\begin{equation}
\label{eq:pwa-io}
\q(t) = \btheta_{\sigma_t}^\top \begin{bmatrix} \r(t) \\ 1 \end{bmatrix} + \bvarep(t),
\end{equation}
where $\btheta_i$ is the parameter vector, $\bvarep(t)$ denotes the measurement error and, given a model order $m$, the regressor $\r(t)$ is defined~as
\begin{equation}
\label{eq:pwa-io-regressor}
\r(t) = \begin{bmatrix} \q(t-1)^\top \!\! & \cdots \! & \q(t-m)^\top \end{bmatrix}^\top \in \Re^{pm}.
\end{equation}
Given time-series data, $\{ \q(t) \}_{t=1}^{T}$, our goal is to recover the underlying model parameters, $\{ \bbeta_i \}_{i=1}^{K}$, and estimate the switching variable at each time instant, $\sigma_t$, hence recover the segmentation of the data. This problem  corresponds to the identification of hybrid dynamical systems \cite{Paoletti:EJC07}. 

\begin{table*}[t!]
\caption{\footnotesize The top rows show the sequence identifier, number of frames and activities for each of the 14 sequences in the CMU MoCap~dataset. The bottom rows show the clustering error ($\%$) of Spectral Clustering (SC), Spectral BiClustering (SBiC), Kmedoids, Affinity Propagation (AP) and our propose algorithm, DS3.} \centering
\begin{small}
\begin{tabular}{|@{\;}c@{\;}|@{\;}c@{\;}|@{\;}c@{\;}|@{\;}c@{\;}|@{\;}c@{\;}|@{\;}c@{\;}|@{\;}c@{\;}|@{\;}c@{\;}|@{\;}c@{\;}|@{\;}c@{\;}|@{\;}c@{\;}|@{\;}c@{\;}|@{\;}c@{\;}|@{\;}c@{\;}|@{\;}c@{\;}|}
\hline
Sequence number & 1 & 2 & 3 & 4 & 5 & 6 & 7 & 8 & 9 & 10 & 11 & 12 & 13 & 14 \\
\hline
\hline
$\#$ frames &    $865$ & $2,115$ & $1,668$ & $2,016$ & $1,638$ & $1,964$ & $1,708$ & $1,808$ & $931$ & $1,514$ & $1,102$ & $1,738$ & $1,164$ & $1,204$\\
$\#$ activities & $4$  &   $8$  &   $7$  &   $7$  &   $7$  &  $10$  &   $6$  &   $9$   &  $4$  &   $4$  &   $4$  &   $7$   &  $6$  &  $4$\\
\hline
\hline
SC error ($\%$) &  $23.86$ & $30.61$ & $19.02$ & $40.60$ & $26.43$ & $47.77$ & $14.85$ & $38.09$ & $\textcolor{black}{\textbf{9.02}}$ & $8.31$ & $\textcolor{black}{\textbf{13.26}}$ & $\textcolor{black}{\textbf{3.47}}$ & $27.61$ & $49.46$\\
SBiC error ($\%$) &  $22.77$ & $22.08$ & $18.94$ & $28.40$ & $29.85$ & $30.96$ & $30.50$ & $24.78$ & $13.03$ & $12.68$ & $28.34$ & $23.68$ & $35.14$ & $40.86$\\
Kmedoids error ($\%$) &  $18.26$ & $46.26$ & $49.89$ & $51.99$ & $37.07$ & $54.75$ & $29.81$ & $49.53$ & $9.71$ & $33.50$ & $35.35$ & $33.80$ & $40.41$ & $48.39$\\
AP error ($\%$) &  $22.93$ & $41.22$ & $49.66$ & $54.56$ & $37.87$ & $50.19$ & $37.84$ & $48.37$ & $9.71$ & $26.05$ & $36.17$ & $23.84$ & $37.75$ & $54.53$\\
DS3 error ($\%$) &   $\textcolor{black}{\textbf{5.33}}$ & $\textcolor{black}{\textbf{9.90}}$ & $\textcolor{black}{\textbf{12.27}}$ & $\textcolor{black}{\textbf{19.64}}$ & $\textcolor{black}{\textbf{16.55}}$ & $\textcolor{black}{\textbf{14.66}}$ & $\textcolor{black}{\textbf{12.56}}$ & $\textcolor{black}{\textbf{11.73}}$ & $11.18$ & $\textcolor{black}{\textbf{3.32}}$ & $22.97$ & $6.18$ & $\textcolor{black}{\textbf{24.45}}$ & $\textcolor{black}{\textbf{28.92}}$\\
\hline
\end{tabular}
\end{small}
\label{tab:CMUmocap-errors}
\end{table*}

To address the problem, we propose to first estimate a set of local models with parameters $\{ \hbbeta_i \}_{i=1}^{M}$ for the time-series data $\{ \q(t) \}_{t=1}^{T}$. We do this by taking $M$ snippets of length $\Delta$ from the time-series trajectory and estimating a dynamical system, in the form \eqref{eq:pwa-ss} or \eqref{eq:pwa-io} or other forms, for each snippet using standard system identification techniques. Once local models are learned, we form the source set, $\X$, by collecting the $M$ learned models and from the target set, $\Y$, by taking snippets at different time instants. We compute dissimilarities by $d_{ij} = \ell(\q(j);\hbbeta_i)$, where $\ell(\q(j);\hbbeta_i)$ denotes the error of representing the snippet ending at $\q(j)$ using the $j$-th model with parameters $\hbbeta_i$. We then run the DS3 algorithm whose output will be a few representative models that explain the data efficiently along with the segmentation of data according to memberships to selected models. 

\begin{remark}
Our proposed method has several advantages over the state-of-the-art switched system identification methods \cite{Paoletti:EJC07, Ferrari:Automatica03, Vidal:CDC03}. First, we are not restricted to a particular class of models, such as linear versus nonlinear or state--space versus input/output models. In fact, as long as we can estimate local models using standard identification procedures, we can deal with all the aforementioned models. Second, we overcome the non-convexity of the switched system identification, due to both $\{ \bbeta_i \}_{i=1}^{K}$ and $\sigma_t$ being unknown, by using a large set of candidate models $\{ \hat{\bbeta_i} \}_{i=1}^{K}$ and selecting a few of them in a convex programming framework. Moreover, since both arguments in $\ell(\q(j);\hbbeta_i)$ are known, we can use arbitrary loss function in our algorithm.
\end{remark}

\subsubsection{Segmentation of human activities}
\label{sec:humanactivityclustering}
To examine the performance of our proposed framework, we consider modeling and segmentation of human activities in motion capture data. We use the Carnegie Mellon Motion Capture dataset \cite{MoCap:12}, which consists of time-series data of different subjects, each performing several activities. The motion capture system uses 42 markers per subject and records measurements at multiple joints of the human body captured at different time instants $t \in [1,T]$. Similar to \cite{Barbic:GI04} and \cite{Delatorre:PAMI13}, we use the 14 most informative joints. For each time instant $t$, we form a data point $\q(t) = \begin{bmatrix} \q_{1}(t)^\top \!\!& \cdots \!\!& \q_{14}(t)^\top  \end{bmatrix}^\top \in \Re^{42}$, where $\q_{i}(t) \in \mathbb{S}^3$ is the complex form of the quaternion for the $i$-th joint at the time $t$. We consider overlapping snippets of length $\Delta$ and estimate a discrete--time state--space model of the form \eqref{eq:pwa-ss} for each snippet using the subspace identification method \cite{Overschee:96}. We set the loss function $\ell(\q(j);\hbbeta_i)$ to be the Euclidean norm of the representation error of the snippet ending at $\q(j)$ using the $i$-th estimated model, $\hbbeta_i$. We use all $14$ trials from subject $86$ in the dataset, where each trial is a combination of multiple activities, such as jumping, squatting, walking, drinking, etc, as shown in Figure \ref{fig:CMUmocap-samples}.

For DS3, we use snippets of length $\Delta = 100$ to estimate local models. Since Kmedoids and AP deal with a single dataset, we use Euclidean distances between pairs of data points as dissimilarities. We also evaluate the Spectral Clustering (SC) performance \cite{Ng:NIPS01, Shi-Malik:PAMI00}, where we compute the similarity between a data point and each of its $\kappa$ nearest neighbors as $\exp{(-\| \q(i) - \q(j) \|_2 / \gamma)}$. We use $\kappa=10$ and $\gamma = 6$, which result in the best performance for SC. We also run the Spectral Bi-Clustering (SBiC) algorithm \cite{Dhillon:KDD01}, which similar to DS3 can work with pairwise relationships between models and data. However, the goal of SBiC is graph partitioning rather than finding representatives. We use $\exp{(- \ell(\q(j);\hbbeta_i) / \gamma)}$ as the edge weights between models and data and set $\gamma = 0.0215$, which gives the best performance for SBiC. We assume the number of activities, $K$, in each time-series trajectory is known and run all methods to obtain $K$ clusters. 

Table \ref{tab:CMUmocap-errors} shows the results, from which we make the following conclusions:
\newline\noindent\textbf{--} Kmedoids and AP generally have large errors on the dataset. This comes from the fact that they try to cluster the time-series trajectory by choosing $K$ representative data points and assigning other points to them. However, a data point itself may have a large distance to other points generated by the same model. While one can also try dissimilarities between models, computing distances between models is a challenging problem \cite{Afsari:CVPR12}. 
\newline\noindent\textbf{--} SC and SBiC obtain smaller errors than Kmedoids and AP, yet large errors, in general. This comes from the fact that they try to cluster data by minimizing the cut criterion, hence are effective only when nodes from different classes are sufficiently~dissimilar.  
\newline\noindent\textbf{--} DS3 obtains small error on the dataset. This is due the the fact that not only DS3 allows for different source and target sets, which results in finding a few models underlying the dynamics of data, but also, based on our theory, it can cluster datasets when dissimilarities between some elements within the same class are higher than dissimilarities between different classes, \ie, it succeeds in cases where graph partitioning can fail.

Figure \ref{tab:CMUmocap-kntSize} shows the segmentation error of DS3 as we change the length, $\Delta$, of snippets to estimate local models. For each value of $\Delta$, we show segmentation errors on all trials by different color bars and the average error over trials by a black horizontal line. Notice that the results do not change much by changing the value of $\Delta$. This comes from the fact that, for each snippet length, among local model estimates, there exist models that well represent each of the underlying activities. However, if $\Delta$ is very large, snippets will contain data from different models/activities, hence, local estimated models cannot well represent underlying models/activities in the time-series trajectory.

\begin{figure}[t!]
\centering
\includegraphics[width=0.70\linewidth,  trim =  105 25 120 43, clip]{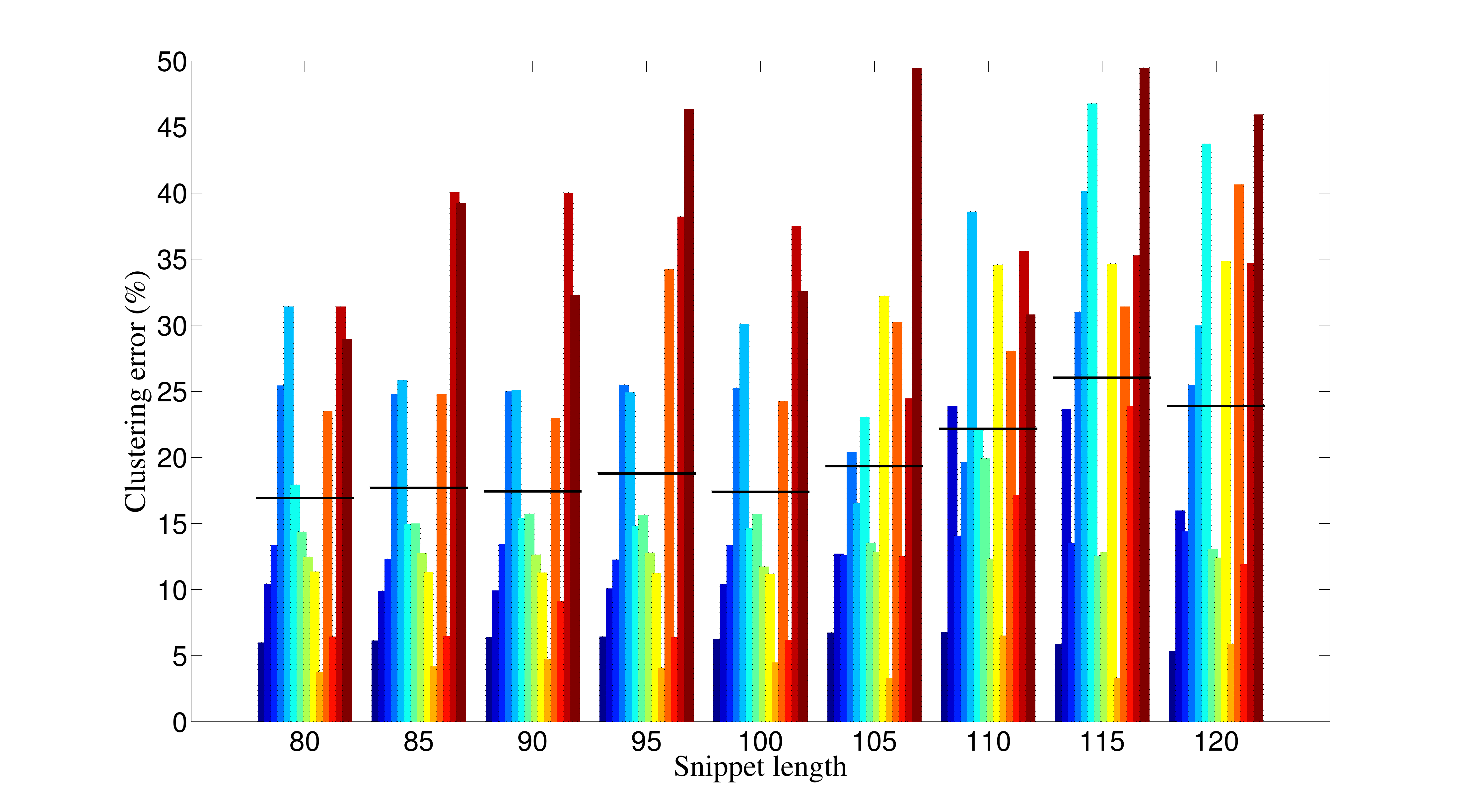}
\vspace{-.8mm} 
\caption{\footnotesize{Clustering error ($\%$) of DS3 on the $14$ sequences (sequence 1: dark blue---sequence 14: dark red) in the CMU~MoCap dataset as a function of the length of snippets used to estimate dynamical systems. The horizontal black line shows the average clustering error for each snippet length over all 14 sequences.}}
\label{tab:CMUmocap-kntSize}
\vspace{0mm}
\end{figure}

{\subsection{Dealing with outliers}
In this section, we examine the performance of our algorithm, formulated in Section \ref{sec:outlier}, for dealing with outliers. To do so, we consider the problem of model selection and segmentation of dynamic data using DS3, which we studied in Section \ref{sec:dynamicdataclustering}, and introduce outliers to the target set. More specifically, we take the motion capture data corresponding to human activities, which we considered in Section \ref{sec:humanactivityclustering}, and exclude one of the activities present in the time series to learn ensemble of dynamical models. Thus, learned models would provide good representatives for all except one of the activities. We apply the optimization in \eqref{eq:tracerow-matrix-outlier1} where we set the weights $w_j$ according to \eqref{eq:outlierweights} with varying values of $\beta > 0$ and $\tau \in \{0.1, 1\}$ and compute the False Positive Rate (FPR) and True Positive Rate (TPR) for $\beta \in [0.1,150]$. Figure \ref{fig:roc_outliers} shows ROC curves obtained by DS3. Notice that our method achieves a high TPR at a low FPR. More precisely, with $\tau = 0.1$, for `Walk' we obtain $95.2\%$ TPR at $12.2\%$ FPR, for `Jump' we obtain $90.6\%$ TPR at $1.5\%$ FPR, and for `Punch' we obtain $90.67\%$ TPR at $5.94\%$ FPR. As a result, we can effectively detect and reject outlying activities in times series data.
}

\begin{figure}[h!]
\centering
\begin{subfigure}[b]{0.486\linewidth}
\includegraphics[width=\linewidth, trim =  17 0 56 20, clip]{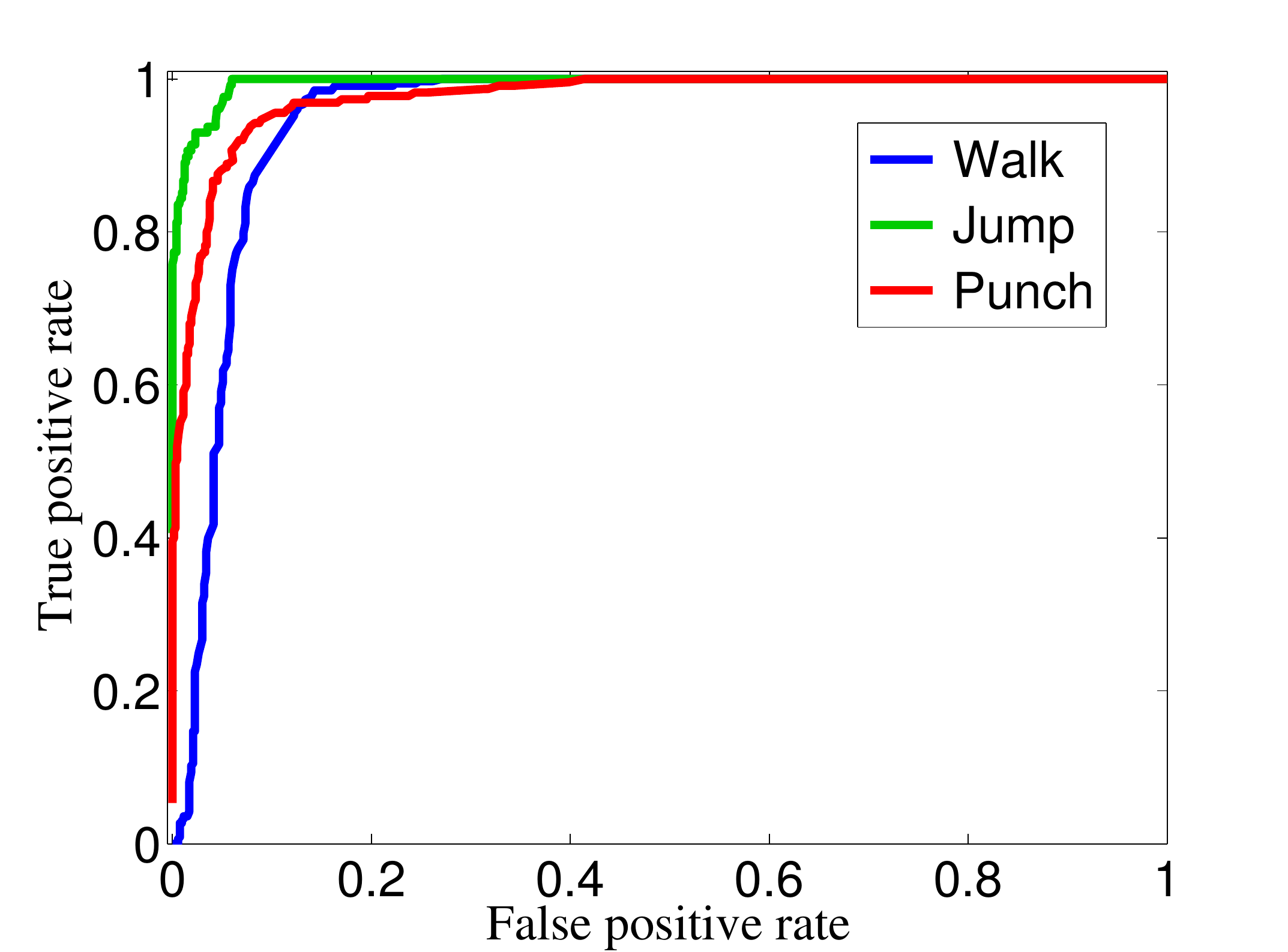}
\caption{~$\tau = 0.1$}
\end{subfigure}
\hspace{0mm}
\begin{subfigure}[b]{0.486\linewidth}
\includegraphics[width=\linewidth, trim =  17 0 56 20, clip]{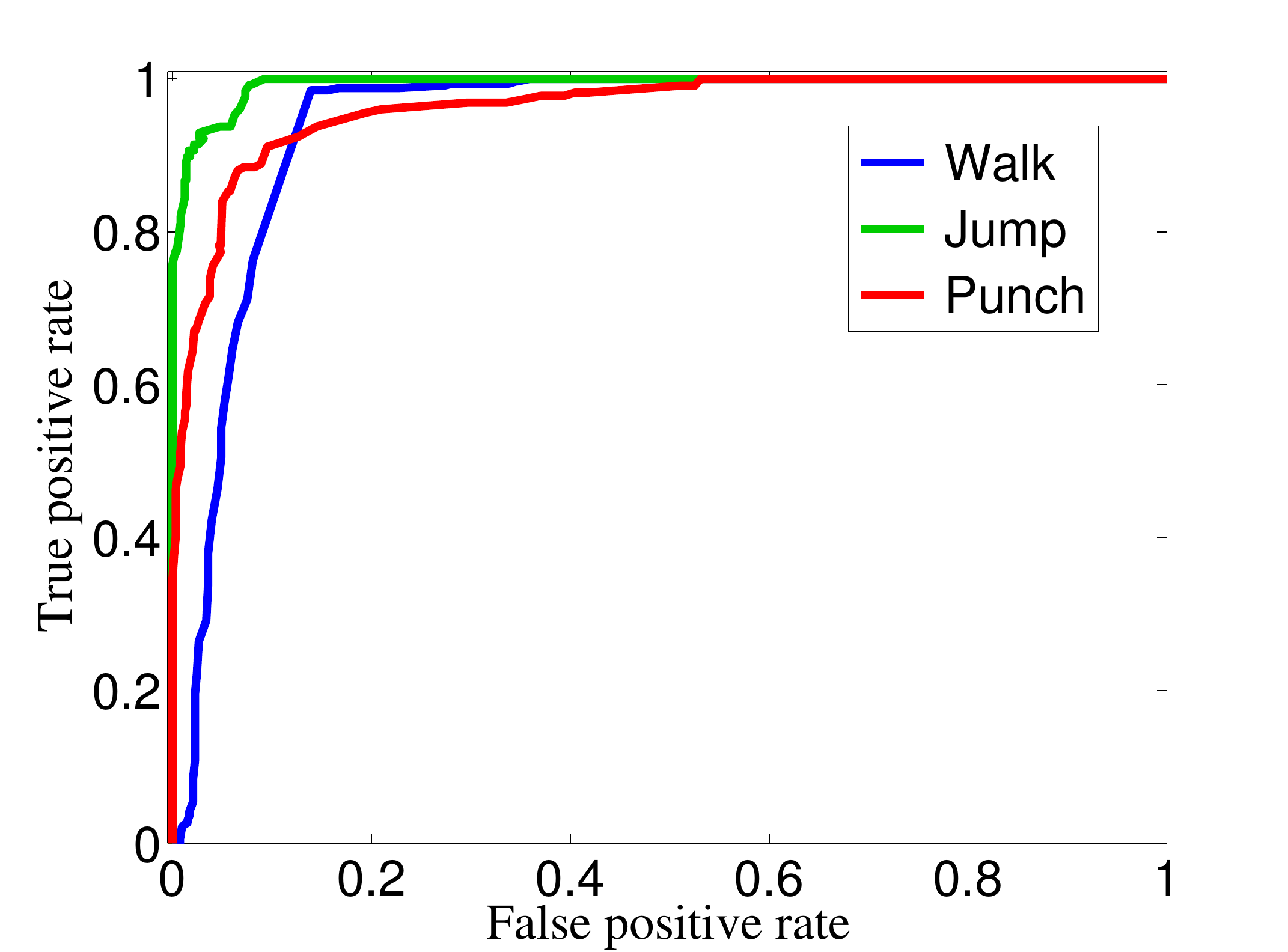}
\caption{~$\tau = 1.0$}
\end{subfigure}
\vspace{-6mm} 
\caption{\footnotesize{{ROC curves for sequence 1 in the CMU MoCap dataset for two values of $\tau$ in \eqref{eq:outlierweights}. We exclude one of the activities in \{ `Walk', `Jump', `Punch' \} at the time of estimating an ensemble of linear dynamical systems from a trajectory.}} }
\label{fig:roc_outliers}
\vspace{0mm}
\end{figure}
%

\section{Conclusion}

Given pairwise dissimilarities between a source and a target set, we considered the problem of finding representatives from the source set that can efficiently encode the target set. We proposed a row-sparsity regularized trace minimization formulation, which can be solved efficiently using convex programming. We showed that our algorithm has theoretical guarantees in that when there is a joint grouping of sets, our method finds representatives from all groups and reveals the clustering of the sets. We also investigated the effect of the regularization parameter on properties of the obtained solution. We provided an efficient implementation of our algorithm using an ADMM approach and showed that our implementation is highly parallelizable, hence further reducing the computational time. Finally, by experiments on real datasets, we showed that our algorithm improves the state of the art on the problems of scene categorization using representative images and modeling and segmentation of time-series data using representative models. Our ongoing research work includes scaling the DS3 algorithm to very large datasets, investigating theoretical guarantees of our algorithm in high-dimensional statistical settings and a more in-depth study of the properties of DS3 when dealing with outliers.


\ifCLASSOPTIONcaptionsoff
  \newpage
\fi

{
\bibliographystyle{IEEEtran}
\bibliography{biblio/subsetselection,biblio/math,biblio/sparse,biblio/vidal,biblio/learning,biblio/vision,biblio/recognition,biblio/segmentation,biblio/control}

\begin{thebibliography}{10}
\providecommand{\url}[1]{#1}
\csname url@samestyle\endcsname
\providecommand{\newblock}{\relax}
\providecommand{\bibinfo}[2]{#2}
\providecommand{\BIBentrySTDinterwordspacing}{\spaceskip=0pt\relax}
\providecommand{\BIBentryALTinterwordstretchfactor}{4}
\providecommand{\BIBentryALTinterwordspacing}{\spaceskip=\fontdimen2\font plus
\BIBentryALTinterwordstretchfactor\fontdimen3\font minus
  \fontdimen4\font\relax}
\providecommand{\BIBforeignlanguage}[2]{{%
\expandafter\ifx\csname l@#1\endcsname\relax
\typeout{** WARNING: IEEEtran.bst: No hyphenation pattern has been}%
\typeout{** loaded for the language `#1'. Using the pattern for}%
\typeout{** the default language instead.}%
\else
\language=\csname l@#1\endcsname
\fi
#2}}
\providecommand{\BIBdecl}{\relax}
\BIBdecl

\bibitem{Simon:ICCV07}
I.~Simon, N.~Snavely, and S.~M. Seitz, ``Scene summarization for online image
  collections,'' \emph{ICCV}, 2007.

\bibitem{Elhamifar:CVPR12}
E.~Elhamifar, G.~Sapiro, and R.~Vidal, ``See all by looking at a few: Sparse
  modeling for finding representative objects,'' \emph{CVPR}, 2012.

\bibitem{Elhamifar:NIPS12}
------, ``Finding exemplars from pairwise dissimilarities via simultaneous
  sparse recovery,'' \emph{NIPS}, 2012.

\bibitem{Taskar:ICML11}
A.~Kulesza and B.~Taskar, ``k-dpps: Fixed-size determinantal point processes,''
  \emph{ICML}, 2011.

\bibitem{Yang:CVPR13}
B.~M. Smith, L.~Zhang, J.~Brandt, Z.~Lin, and J.~Yang, ``Exemplar-based face
  parsing,'' \emph{CVPR}, 2013.

\bibitem{Sha:NIPS14}
B.~Gong, W.~Chao, K.~Grauman, and F.~Sha, ``Diverse sequential subset selection
  for supervised video summarization,'' \emph{NIPS}, 2014.

\bibitem{Bilmes:ARSU09}
H.~Lin, J.~Bilmes, and S.~Xie, ``Graph-based submodular selection for
  extractive summarization,'' \emph{IEEE Automatic Speech Recognition and
  Understanding}, 2009.

\bibitem{Esser:TIP12}
E.~Esser, M.~Moller, S.~Osher, G.~Sapiro, and J.~Xin, ``A convex model for
  non-negative matrix factorization and dimensionality reduction on physical
  space,'' \emph{IEEE Trans. on Image Processing}, 2012.

\bibitem{Frey:Science07}
B.~J. Frey and D.~Dueck, ``Clustering by passing messages between data
  points,'' \emph{Science}, 2007.

\bibitem{Bien:AAStats11}
J.~Bien and R.~Tibshirani, ``Prototype selection for interpretable
  classification,'' \emph{The Annals of Applied Statistics}, 2011.

\bibitem{Gillenwater:NIPS14}
J.~Gillenwater, A.~Kulesza, E.~Fox, and B.~Taskar, ``Expectation-maximization
  for learning determinantal point processes,'' \emph{NIPS}, 2014.

\bibitem{Hartline:WWW08}
J.~Hartline, V.~S. Mirrokni, and M.~Sundararajan, ``Optimal marketing
  strategies over social networks,'' \emph{World Wide Web Conference}, 2008.

\bibitem{Elhamifar:IFAC14}
E.~Elhamifar, S.~Burden, and S.~S. Sastry, ``Adaptive piecewise-affine inverse
  modeling of hybrid dynamical systems,'' \emph{IFAC}, 2014.

\bibitem{Mahoney:NAS09}
M.~W. Mahoney and P.~Drineasp, ``Cur matrix decompositions for improved data
  analysis,'' \emph{Proc. Natl. Acad. Sci.}, 2009.

\bibitem{Garcia:PAMI12}
S.~Garcia, J.~Derrac, J.~R. Cano, and F.~Herrera, ``Prototype selection for
  nearest neighbor classification: Taxonomy and empirical study,'' \emph{IEEE
  Trans. PAMI}, 2012.

\bibitem{Elhamifar:ICCV13}
E.~Elhamifar, G.~Sapiro, A.~Yang, and S.~S. Sastry, ``A convex optimization
  framework for active learning,'' \emph{ICCV}, 2013.

\bibitem{Grauman:CVPR13}
Z.~Lu and K.~Grauman, ``Story-driven summarization for egocentric video,''
  \emph{CVPR}, 2013.

\bibitem{Hebert:WACV14}
I.~Misra, A.~Shrivastava, and M.~Hebert, ``Data-driven exemplar model
  selection,'' \emph{WACV}, 2014.

\bibitem{Grauman:ECCV12}
S.~Vijayanarasimhan and K.~Grauman, ``Active frame selection for label
  propagation in videos,'' \emph{ECCV},
  2012.

\bibitem{Li:CVPR05}
F.~Li and P.~Perona, ``A {B}ayesian hierarchical model for learning natural
  scene categories,'' \emph{CVPR}, 2005.

\bibitem{Lowe:IJCV04}
D.~Lowe, ``Distinctive image features from scale-invariant keypoints,''
  \emph{IJCV}, 2004. 

\bibitem{Kaufman:TR87}
L.~Kaufman and P.~Rousseeuw, ``Clustering by means of medoids,'' \emph{Y.
  Dodge (Ed.), Statistical Data Analysis based on the L1 Norm}, 1987.

\bibitem{Gu:SIAM96}
M.~Gu and S.~C. Eisenstat, ``Efficient algorithms for computing a strong
  rank-revealing qr factorization,'' \emph{SIAM Journal on Scientific
  Computing}, 1996.

\bibitem{Tropp:SODA09}
J.~A. Tropp, ``Column subset selection, matrix factorization, and eigenvalue
  optimization,'' \emph{ACM-SIAM Symp. Discrete Algorithms (SODA)}, 2009.

\bibitem{Boutsidis:SODA09}
C.~Boutsidis, M.~W. Mahoney, and P.~Drineas, ``An improved approximation
  algorithm for the column subset selection problem,'' \emph{ACM-SIAM Symp. Discrete Algorithms (SODA)}, 2009.

\bibitem{Golland:NIPS07}
D.~Lashkari and P.~Golland, ``Convex clustering with exemplar-based models,''
  \emph{NIPS}, 2007.

\bibitem{Chan:LAA87}
T.~Chan, ``Rank revealing qr factorizations,'' \emph{Lin. Alg. and its Appl.}, 1987.

\bibitem{Balzano:WNIPS10}
L.~Balzano, R.~Nowak, and W.~Bajwa, ``Column subset selection with missing
  data,'' in \emph{NIPS Workshop on Low-Rank Methods for Large-Scale Machine
  Learning}, 2010.

\bibitem{Bien:NIPS10}
J.~Bien, Y.~Xu, and M.~W. Mahoney, ``Cur from a sparse optimization
  viewpoint,'' \emph{NIPS}, 2010.

\bibitem{Charikar:JCSS02}
M.~Charikar, S.~Guha, A.~Tardos, and D.~B. Shmoys, ``A constant-factor
  approximation algorithm for the k-median problem,'' \emph{Journal of Computer
  System Sciences}, 2002.

\bibitem{Frey:NIPS06}
B.~J. Frey and D.~Dueck, ``Mixture modeling by affinity propagation,''
  \emph{NIPS}, 2006.

\bibitem{Frey:UAI11}
I.~E. Givoni, C.~Chung, and B.~J. Frey, ``Hierarchical affinity propagation,''
  \emph{UAI}, 2011.

\bibitem{Duda:04}
R.~Duda, P.~Hart, and D.~Stork, \emph{Pattern Classification}.\hskip 1em plus
  0.5em minus 0.4em\relax Wiley-Interscience, 2004.

\bibitem{Frey:ICCV07}
D.~Dueck and B.~J. Frey, ``Non-metric affinity propagation for unsupervised
  image categorization,'' \emph{ICCV},
  2007.

\bibitem{Macchi:AAP75}
M.~Macchi, ``The coincidence approach to stochastic point processes,''
  \emph{Advances in Applied Probability}, 1975.

\bibitem{Borodin:Arxiv09}
A.~Borodin, ``Determinantal point processes,''
  \emph{http://arxiv.org/abs/0911.1153}, 2009.

\bibitem{Taskar:AISTATS13}
R.~H. Affandi, A.~Kulesza, E.~B. Fox, and B.~Taskar, ``Nystrom approximation
  for large-scale determinantal processes,'' \emph{ICML}, 2013.

\bibitem{Bilmes:INTERSPEECH09}
H.~Lin and J.~A. Bilmes, ``How to select a good training-data subset for
  transcription: Submodular active selection for sequences,'' \emph{Annual
  Conference of the International Speech Communication Association}, 2009.

\bibitem{Krause:JMLR08}
A.~Krause, H.~B. McMahan, C.~Guestrin, and A.~Gupta, ``Robust submodular
  observation selection,'' \emph{JMLR}, 2008.

\bibitem{Shmoys:TheoryComp12}
D.~B. Shmoys, E.~Tardos, and K.~Aardal, ``Approximation algorithms for facility
  location problems,'' \emph{ACM Symposium on Theory of Computing}, 1997.

\bibitem{Li:InfoComp12}
S.~Li, ``A 1.488 approximation algorithm for the uncapacitated facility
  location problem,'' \emph{Information and Computation}, 2012.

\bibitem{Li:TheoryComp12}
S.~Li and O.~Svensson, ``Approximating k-median via pseudo-approximation,''
  \emph{ACM Symposium on Theory of Computing}, 2013.

\bibitem{Tropp:SP06}
J.~A. Tropp., ``Algorithms for simultaneous sparse approximation. part ii:
  Convex relaxation,'' \emph{Signal Processing, special issue "Sparse
  approximations in signal and image processing"}, 2006.

\bibitem{Jenatton:JMLR11}
R.~Jenatton, J.~Y. Audibert, and F.~Bach, ``Structured variable selection with
  sparsity-inducing norms,'' \emph{JMLR}, 2011.

\bibitem{Afsari:CVPR12}
B.~Afsari, R.~Chaudhry, A.~Ravichandran, and R.~Vidal, ``Group action induced
  distances for averaging and clustering linear dynamical systems with
  applications to the analysis of dynamic scenes,'' \emph{CVPR}, 2012.

\bibitem{cvx}
M.~Grant and S.~Boyd, ``{CVX}: Matlab software for disciplined convex
  programming,'' \url{http://cvxr.com/cvx}.

\bibitem{Boyd:FTML10}
S.~Boyd, N.~Parikh, E.~Chu, B.~Peleato, and J.~Eckstein, ``Distributed
  optimization and statistical learning via the alternating direction method of
  multipliers,'' \emph{Foundations and Trends in Machine Learning}, 2010.

\bibitem{Gabay:CMA76}
D.~Gabay and B.~Mercier, ``A dual algorithm for the solution of nonlinear
  variational problems via finite-element approximations,'' \emph{Comp. Math.
  Appl.}, 1976.

\bibitem{Xing:NIPS02}
E.~P. Xing, A.~Y. Ng, M.~I. Jordan, and S.~Russell, ``Distance metric learning,
  with application to clustering with side-information,'' \emph{NIPS}, 2002.

\bibitem{Dhillon:ICML07}
J.~V. Davis, B.~Kulis, P.~Jain, S.~Sra, and I.~S. Dhillon,
  ``Information-theoretic metric learning,'' \emph{ICML}, 2007.

\bibitem{Dhillon:KDD01}
I.~S. Dhillon, ``Co-clustering documents and words using bipartite spectral
  graph partitioning,'' \emph{ACM SIGKDD International Conference on Knowledge
  Discovery and Data Mining}, 2001.

\bibitem{Dhillon:KDD03}
I.~S. Dhillon, S.~Mallela, and D.~S. Modha, ``Information-theoretic
  co-clustering,'' \emph{ACM SIGKDD International Conference on Knowledge
  Discovery and Data Mining}, 2003.

\bibitem{Dhillon:JMLR07}
A.~Banerjee, I.~S. Dhillon, J.~Ghosh, S.~Merugu, and D.~S. Modha, ``A
  generalized maximum entropy approach to bregman co-clustering and matrix
  approximation,'' \emph{JMLR}, 2007.

\bibitem{BoydVandenberghe04}
S.~Boyd and L.~Vandenberghe, \emph{Convex Optimization}.\hskip 1em plus 0.5em
  minus 0.4em\relax Cambridge University Press, 2004.

\bibitem{Combettes:MMS05}
P.~Combettes and V.~Wajs, ``Signal recovery by proximal forward-backward
  splitting,'' \emph{SIAM Journal on Multiscale Modeling and Simulation}, 2005.

\bibitem{Chaux:IP07}
C.~Chaux, P.~Combettes, J.~C. Pesquet, and V.~Wajs, ``A variational formulation
  for frame-based inverse problems,'' \emph{Inverse Problems}, 2007.

\bibitem{Duchi:ICML08}
J.~Duchi, S.~Shalev-Shwartz, Y.~Singer, and T.~Chandra, ``Efficient projections
  onto the l1-ball for learning in high dimensions,'' \emph{ICML}, 2008.

\bibitem{Nellore:TechRep14}
A.~Nellore and R.~Ward, ``Recovery guarantees for exemplar-based clustering,''
  \emph{arXiv:1309.3256}, 2014.

\bibitem{Awasthi:ITCS15}
P.~Awasthi, A.~S. Bandeira, M.~Charikar, R.~Krishnaswamy, S.~Villar, and
  R.~Ward, ``Relax, no need to round: Integrality of clustering formulations,''
  in \emph{Conference on Innovations in Theoretical Computer Science},
  2015.

\bibitem{Wesolowsky:LS93}
G.~Wesolowsky, ``The weber problem: History and perspective,'' \emph{Location
  Science}, 1993.

\bibitem{Lazebnik:CVPR06}
S.~Lazebnik, C.Schmid, and J.~Ponce, ``Beyond bags of features: Spatial pyramid
  matching for recognizing natural scene categories,'' \emph{CVPR}, 2006.

\bibitem{Koller:book09}
D.~Koller and N.~Friedman, \emph{Probabilistic Graphical Models: Principles and
  Techniques}.\hskip 1em plus 0.5em minus 0.4em\relax New York: MIT Press,
  2009.

\bibitem{Weinberger:ICML14}
M.~Kusner, S.~Tyree, K.~Weinberger, and K.~Agrawal, ``Stochastic neighbor
  compression,'' \emph{ICML}, 2014.

\bibitem{MoCap:12}
``Carnegie mellon university motion capture database,''
  http://mocap.cs.cmu.edu, 2012.

\bibitem{Paoletti:EJC07}
S.~Paoletti, A.~Juloski, G.~Ferrari-Trecate, and R.~Vidal, ``Identification of
  hybrid systems: A tutorial,'' \emph{European Journal of Control}, 2007.

\bibitem{Ferrari:Automatica03}
G.~Ferrari-Trecate, M.~Muselli, D.~Liberati, and M.~Morari, ``A clustering
  technique for the identification of piecewise affine systems,''
  \emph{Automatica}, 2003.

\bibitem{Vidal:CDC03}
R.~Vidal, S.~Soatto, Y.~Ma, and S.~Sastry, ``An algebraic geometric approach to
  the identification of a class of linear hybrid systems,'' \emph{CDC}, 2003.

\bibitem{Barbic:GI04}
J.~Barbic, A.~Safonova, J.~Y. Pan, C.~Faloutsos, J.~K. Hodgins, and N.~S.
  Pollard, ``Segmenting motion capture data into distinct behaviors,''
  \emph{Graphics Interface}, 2004.

\bibitem{Delatorre:PAMI13}
F.~Zhou, F.~D. Torre, and J.~K. Hodgins, ``Hierarchical aligned cluster
  analysis for temporal clustering of human motion,'' \emph{IEEE Trans.
  PAMI}, 2013.

\bibitem{Overschee:96}
P.~V. Overschee and B.~D. Moor, \emph{Subspace Identification For Linear
  Systems: Theory, Implementation, Applications}.\hskip 1em plus 0.5em minus
  0.4em\relax Kluwer Academic Publishers, 1996.

\bibitem{Ng:NIPS01}
A.~Ng, Y.~Weiss, and M.~Jordan, ``On spectral clustering: analysis and an
  algorithm,'' in \emph{NIPS}, 2001.

\bibitem{Shi-Malik:PAMI00}
J.~Shi and J.~Malik, ``Normalized cuts and image segmentation,'' \emph{{IEEE}
  Trans. PAMI}, 2000.

\end{thebibliography}
}

\appendix

\section*{Proofs of Theoretical Results}

In this section, we prove the theoretical results in the paper for our proposed optimization program in \eqref{eq:tracerow-matrix-1}. To do so, we make use of the following Lemmas, which are standard results from convex analysis and can be found in \cite{BoydVandenberghe04}.

\vspace{1.5mm}
\begin{lemma}
\label{lem:subgrad_q2}
For a vector $\z \in \Re^N$, the subgradients of $\| \z \|_{2}\,$ at $\z = \0\,$ and $\z = \1$ are given by 
\begin{eqnarray}
\partial_{\z = \0} \| \z \|_{2} &=& \{ \u \in \Re^N : \| \u \|_2 \leq 1 \}, \label{eq:subgrad_q2_z0}\\
\partial_{\z = \1} \| \z \|_{2} &=& \{ \u \in \Re^N: \u = \frac{1}{\sqrt{N} } \1 \} \label{eq:subgrad_q2_z1}.
\end{eqnarray}
\end{lemma}
\vspace{1.5mm}
\begin{lemma}
\label{lem:subgrad_qinf}
For a vector $\z \in \Re^N$, the subgradients of $\| \z \|_{\infty}\,$ at $\z = \0\,$ and $\z = \1$ are given by 
\begin{eqnarray}
\partial_{\z = \0} \| \z \|_{\infty} &=& \{ \u \in \Re^N : \| \u \|_1 \leq 1 \}, \label{eq:subgrad_qinf_z0}\\
\partial_{\z = \1} \| \z \|_{\infty} &=& \{ \u \in \Re^N: \1^{\top} \u = 1, \; \u \geq \0 \} \label{eq:subgrad_qinf_z1}.
\end{eqnarray}
\end{lemma}
\vspace{1.5mm}
We also make use of the following Lemma, which we prove next.
\vspace{1.5mm}
\begin{lemma}
\label{lem:subgrad3}
The sets $\S_1$ and $\S_2$ defined as
\begin{eqnarray}
\S_1 &\!\!\!\! \triangleq &\!\!\!\! \{ \u - \v \in \Re^N\!\!: \| \u \|_1 \leq 1, \, \1^{\top} \v = 1, \, \v \geq \0 \},\label{eq:s1}\\
\S_2 &\!\!\!\! \triangleq &\!\!\!\! \{ \bdelta \in \Re^N\!\!: \| \bdelta \|_1 \leq 2, \, \1^{\top} \bdelta \leq 0 \}.\label{eq:s2}
\end{eqnarray}
are equal, \ie, $\S_1 = \S_2$.
\end{lemma}
\vspace{1.5mm}
\begin{proof}
In order to prove $\S_1 = \S_2$, we need to show that $\S_1 \subseteq \S_2$ and $\S_2 \subseteq \S_1$. First, we show that $\S_1 \subseteq \S_2$. Take any $\x \in \S_1$. Using \eqref{eq:s1}, we can write $\x$ as $\x = \u - \v$, where $\| \u \|_1 \leq 1$, $\v \geq \0$ and $\1^{\top} \v = 1$. Since
\begin{equation}
\label{eq:cond1}
\| \x \|_1 = \| \u - \v \|_1 \leq \| \u \|_1 + \| \v \|_1 \leq 2,
\end{equation}
and 
\begin{equation}
\label{eq:cond2}
\1^{\top} \x = \1^{\top} \u - \1^{\top} \v = \1^{\top} \u  - 1 \leq \| \u \|_1 - 1 \leq 0,
\end{equation}
from \eqref{eq:s2} we have that $\x \in \S_2$. Thus, $\S_1 \subseteq \S_2$. Next, we show that $\S_2 \subseteq \S_1$. Take any $\bdelta \in \S_2$. From \eqref{eq:s2}, we have $\| \bdelta \|_1 \leq 2$ and $\1^{\top} \bdelta \leq 0$. Without loss of generality, let 
\begin{equation}
\bdelta = \begin{bmatrix} \bdelta_{+} \\ -\bdelta_{-} \end{bmatrix},
\end{equation}
where $\bdelta_{+}$ and $\bdelta_{-}$ denote, respectively, nonnegative and negative elements of $\bdelta$, hence, $\bdelta_{+} \geq 0$ and $\bdelta_{-} > 0$. Notice that we have 
\begin{equation}
\| \bdelta \|_1 = \1^{\top} \bdelta_{+} + \1^{\top} \bdelta_{-} \leq 2,
\end{equation}
and 
\begin{equation}
\1^{\top} \bdelta = \1^{\top} \bdelta_{+} - \1^{\top} \bdelta_{-} \leq 0.
\end{equation}
The two inequalities above imply that 
\begin{equation}
\1^{\top} \bdelta_{+} \leq 1.
\end{equation}
In order to show $\bdelta \in \S_1$, we consider three cases on the value of $\1^{\top} \bdelta_{-}$.

\smallskip\noindent \emph{Case 1}: Assume $\1^{\top} \bdelta_{-} = 1$. Let $\u = \begin{bmatrix} \bdelta_{+} \\ \0 \end{bmatrix}$ and $\v = \begin{bmatrix} \0 \\ \bdelta_{-} \end{bmatrix}$. We can write $\bdelta = \u - \v$, where $\| \u \|_1 = \1^{\top} \bdelta_{+} \leq 1$, $\v \geq \0$ and $\1^{\top} \v = \1^{\top} \bdelta_{-} = 1$. Thus, according to \eqref{eq:s1}, we have $\bdelta \in \S_1$.

\smallskip\noindent \emph{Case 2}: Assume $\1^{\top} \bdelta_{-} > 1$. We can write
\begin{equation}
\bdelta = \underbrace{\begin{bmatrix} \bdelta_{+} \\ - \bdelta_{-} (1 - \frac{1}{\1^{\top} \bdelta_{-}}) \end{bmatrix}}_{\triangleq u} - \underbrace{\begin{bmatrix} \0 \\ \bdelta_{-} (\frac{1}{\1^{\top} \bdelta_{-}}) \end{bmatrix}}_{\triangleq v}.
\end{equation}
We have $\| \u \|_1 \leq 1$, since
\begin{equation}
\begin{split}
\label{eq:cond3}
\| \u \|_1 &= \1^{\top} \bdelta_{+} + \1^{\top} \bdelta_{-} (1 - \frac{1}{\1^{\top} \bdelta_{-}}) \\ &= \1^{\top} \bdelta_{+} + \1^{\top} \bdelta_{-} - 1 = \| \bdelta \|_1 - 1 \; \leq \; 1.
\end{split}
\end{equation}
We also have
\begin{equation}
\label{eq:cond4}
\1^{\top} \v = \1^{\top} \bdelta_{-} / (\1^{\top} \bdelta) = 1.
\end{equation}
Notice that equations \eqref{eq:cond3} and \eqref{eq:cond4} and the fact that $\v \geq \0$ imply $\bdelta \in \S_1$.

\smallskip\noindent \emph{Case 3}: Assume $\1^{\top} \bdelta_{-} < 1$. Similar to the previous case, let
\begin{equation}
\bdelta = \underbrace{\begin{bmatrix} \bdelta_{+} \\ - \bdelta_{-} (1 - \frac{1}{\1^{\top} \bdelta_{-}}) \end{bmatrix}}_{\triangleq u} - \underbrace{\begin{bmatrix} \0 \\ \bdelta_{-} (\frac{1}{\1^{\top} \bdelta_{-}}) \end{bmatrix}}_{\triangleq v}.
\end{equation}
As a result, we have $\| \u \|_1 \leq 1$, since
\begin{equation}
\begin{split}
\label{eq:cond5}
\| \u \|_1 &= \1^{\top} \bdelta_{+} - \1^{\top} \bdelta_{-} (1 - \frac{1}{\1^{\top} \bdelta_{-}}) \\ &= \1^{\top} \bdelta_{+} - \1^{\top} \bdelta_{-} + 1 = \1^{\top} \bdelta + 1 \; \leq \; 1,
\end{split}
\end{equation}
where we used the fact that $\1^{\top} \bdelta \leq 0$, since $\bdelta \in \S_2$. We also have
\begin{equation}
\label{eq:cond6}
\1^{\top} \v = \1^{\top} \bdelta_{-} (\frac{1}{\bdelta_{-}}) = 1.
\end{equation}
Equations \eqref{eq:cond5} and \eqref{eq:cond6} together with $\v \geq \0$ imply that $\bdelta \in \S_1$.
\end{proof}
\vspace{1.5mm}
We are ready now to prove the result of Theorem \ref{thm:max-lambda} in the paper.
\vspace{1.5mm}

\noindent \textbf{Proof of Theorem \ref{thm:max-lambda}}:
Denote the objective function of \eqref{eq:tracerow-matrix-1} by $J$. In order to prove the result, we consider the cases of $p=2$ and $p=\infty$ separately.

\smallskip\noindent \emph{Case of $p=2$.} First, we incorporate the affine constraints $\sum_{i=1}^{M}{z_{ij}} = 1$ into the objective function of \eqref{eq:tracerow-matrix-1} by rewriting $z_{Mj}$ in terms of other variables as $z_{Mj} = 1 - \sum_{i=1}^{M-1}{z_{ij}}$. Hence, we can rewrite the objective function of \eqref{eq:tracerow-matrix-1} as 
\begin{equation}
\begin{split}
\!\!\!J = \sum_{i=1}^{M-1}{\d_i^{\top} \z_i} &+ \d_M^{\top} \begin{bmatrix} 1-z_{1,1}-\cdots-z_{M-1,1} \\ \vdots \\ 1-z_{1,N}-\cdots-z_{M-1,N} \end{bmatrix} \\&+ \lambda \sum_{i=1}^{M-1}{\sqrt{z_{i,1}^2+z_{i,2}^2+ \cdots+z_{i,N}^2}} \\&+ \lambda \sqrt{ \sum_{i=1}^{N}(1-z_{1,i}- \cdots-z_{M-1,i})^2}.
\end{split}
\end{equation}
Without loss of generality, we assume that in the solution of \eqref{eq:tracerow-matrix-1}, all rows of $\Z$ except the last one are zero (later, we will show which row is the only nonzero vector in the solution $\Z$). From the optimality of the solution, for every $i = 1,\ldots, M-1$, we have
\begin{equation}
\label{eq:optim1}
\0 \in \partial_{\z_i}{J} = \d_i - \d_M + \lambda \, \partial_{\z_i = \0} \| \z_{i} \|_2 - \frac{\lambda}{\sqrt{N}} \1.
\end{equation}
From Lemma \ref{lem:subgrad_q2}, the subgradient of $\| \z_i \|_2$ at $\0$ is a vector $\u \in \Re^N$ which satisfies $\| \u \|_2 \leq 1$. Thus, we can rewrite \eqref{eq:optim1} as 
\begin{equation}
\frac{1}{\sqrt{N}} \, \1 + \frac{\d_M - \d_i}{\lambda} \in \{ \u \in \Re^N : \| \u \|_2 \leq 1 \},
\end{equation}
which implies that
\begin{equation}
\left \| \frac{1}{\sqrt{N}} + \frac{\d_M - \d_i}{\lambda} \right \|_2^2 \leq 1.
\end{equation}
Expanding the left-hand-side of the above inequality, we obtain
\begin{equation}
\label{eq:lambdamax}
\frac{2\lambda}{\sqrt{N}} \1^{\top} (\d_M - \d_i) + \| \d_M - \d_i \|_2^2 \leq 0.
\end{equation}
Since $\| \d_i-\d_M \|_2$ in the above equation is always nonnegative, the first term must be nonpositive, \ie,
\begin{equation}
\1^{\top} (\d_M - \d_i) \leq 0.
\end{equation}
As a result, the index of the nonzero row of the optimal solution corresponds to the one for which $\1^{\top} \d_i$ is minimum (here, without loss of generality, we have assumed $\d_M$ is the row with the minimum dissimilarity sum). Finally, from \eqref{eq:lambdamax}, we obtain
\begin{equation}
\lambda \; \geq \; \frac{\sqrt{N}}{2} \frac{\| \d_i - \d_M \|_2^2}{ \1^{\top} (\d_i - \d_M)}, ~~ \forall \, i \neq M.
\end{equation}
Thus, the threshold value on the regularization parameter beyond which we obtain only one nonzero row in the optimal solution of \eqref{eq:tracerow-matrix-1} is given by
\begin{equation}
\lambda_{\max,2} \triangleq \max_{i \neq N} \frac{\sqrt{N}}{2} \frac{\| \d_i - \d_M \|_2^2}{ \1^{\top} (\d_i - \d_M)}.
\end{equation}
\smallskip\noindent \emph{Case of $p=\infty$.} Similar to the previous case, we incorporate the affine constraints $\sum_{i=1}^{M}{z_{ij}} = 1$ into the objective function of \eqref{eq:tracerow-matrix-1} and rewrite it as 
\begin{equation}
\begin{split}
J &= \sum_{i=1}^{M-1}{\d_i^{\top} \z_i} + \d_M^{\top} \begin{bmatrix} 1-z_{1,1}-\cdots-z_{M-1,1} \\ \vdots \\ 1-z_{1,N}-\cdots-z_{M-1,N} \end{bmatrix} \\&+ \lambda \sum_{i=1}^{M-1}{\| \z_i \|_{\infty}} + \lambda \left \| \begin{bmatrix} 1-z_{1,1}-\cdots-z_{M-1,1} \\ \vdots \\ 1-z_{1,N}-\cdots-z_{M-1,N} \end{bmatrix} \right \|_{\infty}\!\!.
\end{split}
\end{equation}
Without loss of generality, we assume that in the solution of \eqref{eq:tracerow-matrix-1} all rows of $\Z$ except the last one are zero. From the optimality of the solution, for every $i = 1,\ldots, N-1$, we have
\begin{equation}
\begin{split}
\label{eq:subgradinf}
\!\!\0 \in \partial_{\z_i}{J} &= \d_i - \d_M + \lambda \, \partial_{\z_i = \0} \| \z_{i} \|_{\infty} \\&+ \lambda \, \partial \left \| \begin{bmatrix} 1-z_{1,1}-\cdots-z_{M-1,1} \\ \vdots \\ 1-z_{1,N}-\cdots-z_{M-1,N} \end{bmatrix} \right \|_{\infty}.
\end{split}
\end{equation}
From Lemma \ref{lem:subgrad_qinf} we have 
\begin{equation}
\label{eq:subdiffinf1}
\partial_{\z_i = \0} \| \z_i \|_{\infty} \in \{ \u \in \Re^N : \| \u \|_1 \leq 1 \},
\end{equation}
and
\begin{equation}
\label{eq:subdiffinf2}
\!\partial \left \| \! \begin{bmatrix} 1- \sum_{i=1}^{M-1}{z_{i,1}} \\ \!\!\!\!\!\!\!\! \vdots \\ 1- \sum_{i=1}^{M-1}{z_{i,N}} \end{bmatrix} \! \right \|_{\infty} \!\!\!\! \in \{ \v \in \Re^N \!\! : \1^{\top} \v = -1, \, \v \geq \0 \}.
\end{equation}
Substituting \eqref{eq:subdiffinf1} and \eqref{eq:subdiffinf2} in \eqref{eq:subgradinf}, we obtain 
\begin{equation}
\label{eq:subgradinf3}
\frac{\d_i - \d_M}{\lambda} \in \{ \u - \v : \| \u \|_1 \leq 1, \v \leq \0, \1^{\top} \v = -\1 \}.
\end{equation}
From Lemma \ref{lem:subgrad3}, the set on the right-hand-side of \eqref{eq:subgradinf3}, \ie, $\S_1$, is equal to $\S_2$, hence
\begin{equation}
\label{eq:subgradinf4}
\frac{\d_i - \d_N}{\lambda} \in \{ \boldsymbol{\delta} : \| \boldsymbol{\delta} \|_1 \leq 2, \1^{\top} \boldsymbol{\delta} \leq \0 \}.
\end{equation}
The constraint $\1^{\top}(\frac{\d_i - \d_M}{\lambda}) \leq \0$ implies that for every $i$ we must have $\1^{\top} \d_M \leq \1^{\top} \d_i$. In other words, the index of the nonzero row of $\Z$ is given by the row of $\D$ for which $\1^{\top} \d_i$ is minimum (here, without loss of generality, we have assumed $\d_M$ is the row with the minimum dissimilarity sum). From \eqref{eq:subgradinf4}, we also have
\begin{equation}
\frac{\| \d_i - \d_M \|_1}{\lambda} \leq 2, ~~ \forall \, i \neq M,
\end{equation}
from which we obtain
\begin{equation}
\lambda \; \geq \; \frac{\| \d_i - \d_M \|_1}{2}, ~~ \forall \, i \neq M.
\end{equation}
Thus, the threshold value on the regularization parameter beyond which we obtain only one nonzero row in the optimal solution of \eqref{eq:tracerow-matrix-1} is given by
\begin{equation}
\lambda \; \geq \; \lambda_{\max,\infty} \, \triangleq \, \max_{i \neq N} \frac{\| \d_i - \d_M \|_1}{2}.
\end{equation}
\hfill\QED

\vspace{1.5mm}

\noindent \textbf{Proof of Theorem \ref{thm:clustering-lambda}}:
\vspace{1.5mm}
Without loss of generality, assume that elements in $\X$ are ordered so that the first several elements are indexed by $\G^x_1$, followed by elements indexed by $\G^x_2$ and so on. Similarly, without loss of generality, assume that elements in $\Y$ are ordered so that the first several elements are indexed by $\G^y_1$, followed by elements indexed by $\G^y_2$ and so on. Thus, we can write $\D$ and $\Z$ as
\begin{equation}
\label{eq:D}
\D = \begin{bmatrix} \bd_{1,1}^{\top} &  \cdots & \bd_{1,L}^{\top} \\ & \bd_1^\top &  \\  \bd_{2,1}^{\top} &  \cdots  & \bd_{2,L}^{\top}  \\  & \bd_2^\top & \\ & \vdots & \end{bmatrix},
\end{equation}
\begin{equation}
\label{eq:Z}
\Z = \begin{bmatrix} \bz_{1,1}^{\top} &  \cdots & \bz_{1,n}^{\top} \\ & \bz_1^\top &  \\  \bz_{2,1}^{\top} &  \cdots  & \bz_{2,n}^{\top}  \\  & \bz_2^\top & \\ & \vdots & \end{bmatrix},
\end{equation}
where $\bd_{i,j}$ denotes dissimilarities between the first element of $\G^x_i$ and all elements of $\G^y_j$ for $i, j \in \{ 1, \ldots, L \}$. $\bd_i$ denotes dissimilarities between all elements of $\G^x_i$ except its first element and $\Y$. Similarly, we define vectors $\bz_{i,j}$ and matrices $\bz_i$ for assignment variables. 

To prove the result, we use contradiction. Without loss of generality, assume that in the optimal solution of \eqref{eq:tracerow-matrix-1}, $\Z^*$, some elements of $\G^y_j$ for $j > 2$, select some elements of $\G^x_1$ including its first element as their representatives, \ie, $\bz_{1,j} \neq 0$ for some $j > 1$. We show that we can construct a feasible solution which achieves a smaller objective function than $\Z^*$, hence arriving at contradiction. Let
\begin{equation}
\label{eq:Zp}
\Z' = \begin{bmatrix} \bz_{1,1}^{\top} &  \0 & \cdots & \0 \\ & & \hspace{-13mm} \bz_1^\top &  \\  \bz_{2,1}^{\top} & \bz_{2,2}^{\top} + \bz_{1,2}^{\top} &  \cdots  & \bz_{2,n}^{\top}  \\  & & \hspace{-13mm} \bz_2^\top & \\ & \vdots & \\ \bz_{n,1}^{\top} & \bz_{n,2}^{\top} &  \cdots  & \bz_{n,n}^{\top} + \bz_{1,n}^{\top} \\  & & \hspace{-13mm} \bz_n^\top & \end{bmatrix}.
\end{equation}
For $\Z'$, we can write the objective function of \eqref{eq:tracerow-matrix-1} as
\begin{equation}
\begin{split}
&J(\Z') = \lambda \| \bz_{1,1} \|_p + \lambda \left \| \begin{bmatrix} \bz_{2,1} \\ \bz_{2,2} + \bz_{1,2} \\ \vdots \\ \bz_{2,L} \end{bmatrix} \right \|_p + \cdots \\ &+ \lambda \left \| \! \begin{bmatrix} \bz_{L,1} \\ \bz_{L,2} \\ \vdots \\ \bz_{L,L}+ \bz_{1,L} \end{bmatrix} \! \right \|_p \!\!+ \d_{2,1}^\top \z_{1,2} + \cdots +  \d_{L,1}^\top \z_{1,L} + R,
\end{split}
\end{equation}
where $R$ denotes the other terms involved in computing the objective function. Using the triangle inequality for the $\ell_p$-norm, we can write
\begin{equation}
\label{eq:Jzp}
\begin{split}
J(\Z') &\leq \lambda \| \bz_{1,1} \|_p + \lambda \| \bz_{1,2} \|_p + \cdots + \lambda \| \bz_{1,L} \|_p \\ &+ \lambda \left \| \begin{bmatrix} \bz_{2,1} \\ \bz_{2,2} \\ \vdots \\ \bz_{2,L} \end{bmatrix} \right \|_p + \cdots + \lambda \left \| \begin{bmatrix} \bz_{L,1} \\ \bz_{L,2} \\ \vdots \\ \bz_{L,L} \end{bmatrix} \right \|_p \\&+ \bd_{2,1}^\top \bz_{1,2} + \cdots +  \bd_{L,1}^\top \bz_{1,L} + R.
\end{split}
\end{equation}
On the other hand, for the objective function of \eqref{eq:tracerow-matrix-1} evaluated at $\Z^*$, we can write
\begin{equation}
\label{eq:Jzo}
\begin{split}
J(\Z^*) &= \lambda \left \| \! \begin{bmatrix} \bz_{1,1} \\ \bz_{1,2} \\ \vdots \\ \bz_{1,L} \end{bmatrix} \! \right \|_p \!\!+ \lambda \left \| \! \begin{bmatrix} \bz_{2,1} \\ \bz_{2,2} \\ \vdots \\ \bz_{2,L} \end{bmatrix} \! \right \|_p \!\!+ \cdots \!+ \lambda \left \| \! \begin{bmatrix} \bz_{L,1} \\ \bz_{L,2} \\ \vdots \\ \bz_{L,L} \end{bmatrix} \! \right \|_p \\&+ \bd_{1,2}^\top \bz_{1,2} + \cdots +  \bd_{1,L}^\top \bz_{1,L} + R \\ & \geq \lambda \| \bz_{1,1} \|_p + \lambda \left \| \! \begin{bmatrix} \bz_{2,1} \\ \bz_{2,2} \\ \vdots \\ \bz_{2,L} \end{bmatrix} \! \right \|_p \!\!+ \cdots \!+ \lambda \left \| \! \begin{bmatrix} \bz_{L,1} \\ \bz_{L,2} \\ \vdots \\ \bz_{L,L} \end{bmatrix} \! \right \|_p \\&+ \bd_{1,2}^\top \bz_{1,2} + \cdots +  \bd_{1,L}^\top \bz_{1,L} + R.
\end{split}
\end{equation}
If we can show that 
\begin{multline}
\label{eq:desiredineq}
\lambda \| \bz_{1,2} \|_p + \cdots + \lambda \| \bz_{1,L} \|_p < (\bd_{1,2} - \bd_{2,2})^\top \bz_{1,2} \\+ \cdots + (\bd_{1,L} - \bd_{L,L})^\top \bz_{1,L},
\end{multline}
then from \eqref{eq:Jzp} and \eqref{eq:Jzo}, we have $J(\Z') < J(\Z^*)$, hence obtaining contradiction. Notice that for a vector $\a$ and $p \in \{ 2, \infty \}$, we have $\| \a \|_p \leq \| \a \|_1 = \1^\top \a$. Thus, from \eqref{eq:desiredineq}, if we can show that
\begin{multline}
\label{eq:desiredineq2}
\lambda \, \1^\top \bz_{1,2} + \cdots + \lambda \, \1^\top \bz_{1,L} < (\bd_{1,2} - \bd_{2,2})^\top \bz_{1,2} \\+ \cdots + (\bd_{1,L} - \bd_{L,L})^\top \bz_{1,L},
\end{multline}
or equivalently,
\begin{multline}
\label{eq:desiredineq3}
0 < (\bd_{1,2} - \bd_{2,2} - \lambda \1)^\top \bz_{1,2} \\+ \cdots + (\bd_{1,L} - \bd_{L,L} - \lambda \1)^\top \bz_{1,L},
\end{multline}
we obtain contradiction. Since the choice of the first element of $\G^x_j$ for $j > 2$ is arbitrary, we can choose the centroid of $\G^x_j$ as its first element. This, together with the definition of $\lambda_g$ in \eqref{eq:clustering-lambda1} and the assumption that $\tilde{z}_{1,j}  > 0$ for some $j > 2$, implies that the inequality in \eqref{eq:desiredineq3} holds, hence obtaining contradiction.
\hfill\QED
\section*{Results for $p = 2$}
Figure \ref{fig:RepModel_p2} shows the results of running our proposed algorithm using $p = 2$, for approximating the nonlinear manifold presented in the paper. Similarly, Figure \ref{fig:2G-PZ_p2} shows the results of DS3 using $p=2$ for the example of the dataset drawn from a mixture of three Gaussians presented in the paper. Notice that in general, the performance of $p=\infty$ and $p=2$ are quite similar. As mentioned in the paper, the main difference is that $p = 2$ promotes probabilities in the range $[0,1]$, while $p = \infty$ promotes probabilities in $\{0, 1\}$.

\begin{figure*}[t!]
\centering
\begin{subfigure}[b]{0.29\textwidth}
\includegraphics[width=.9\textwidth, trim = 78 68 58 43 , clip]{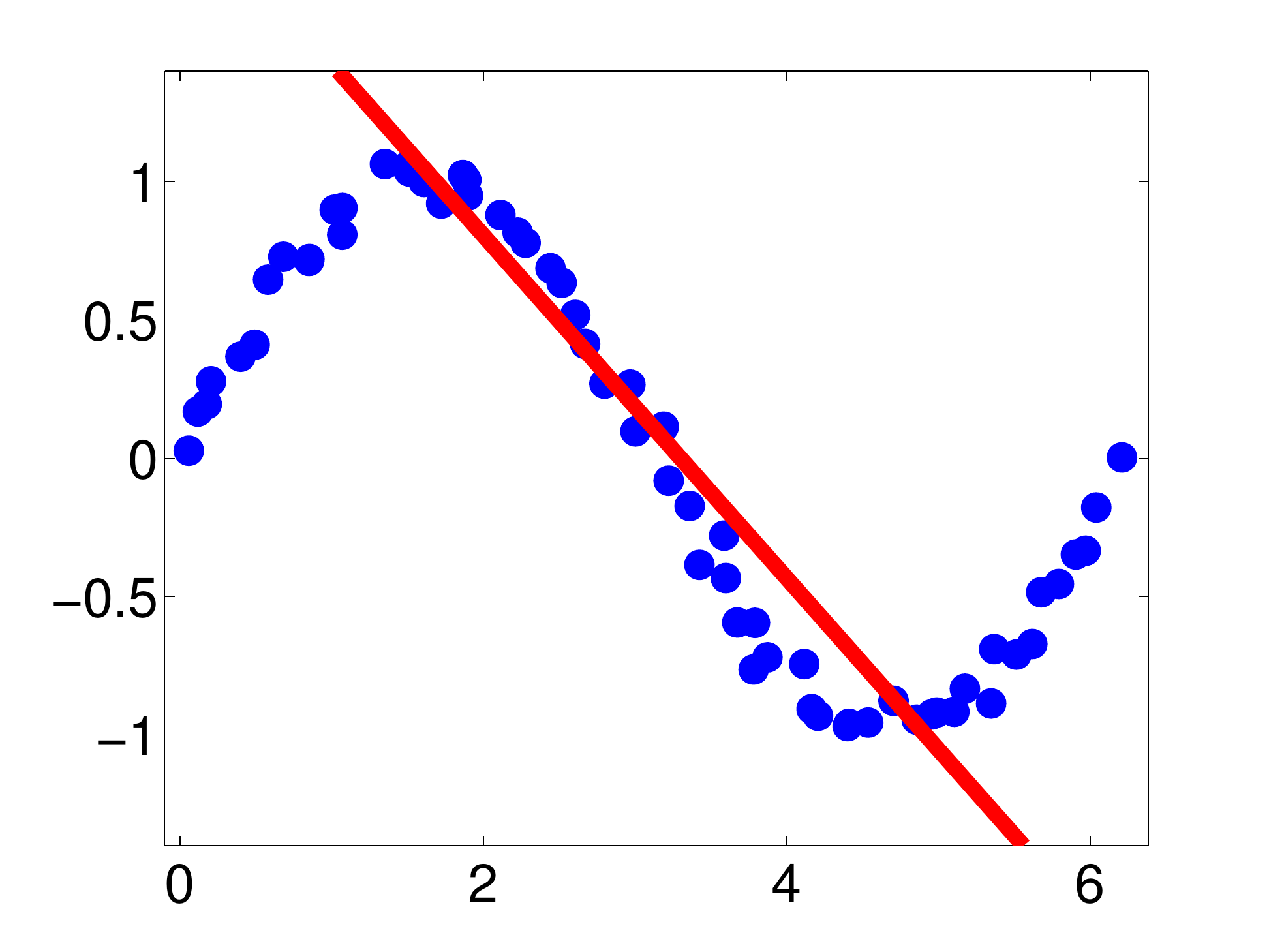}
\vspace{-1mm}
\caption{$\lambda = \lambda_{\max,2}$}
\end{subfigure}
\hspace{4mm}
\begin{subfigure}[b]{0.29\textwidth}
\includegraphics[width=.9\textwidth, trim = 78 68 58 43 , clip]{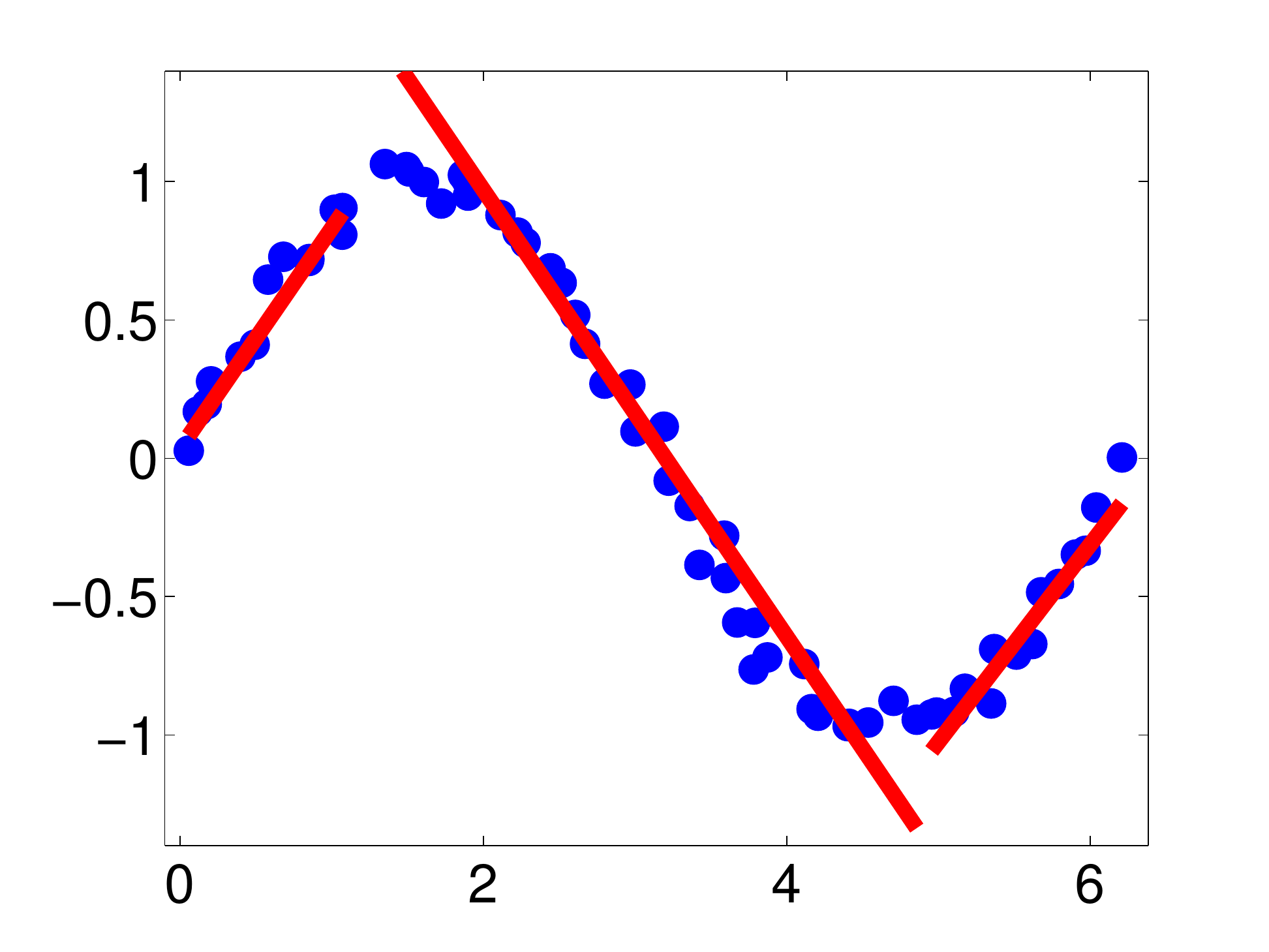}
\vspace{-1mm}
\caption{$\lambda = 0.1 \, \lambda_{\max,2}$}
\end{subfigure}
\hspace{4mm}
\begin{subfigure}[b]{0.29\textwidth}
\includegraphics[width=.9\textwidth, trim = 78 68 58 43 , clip]{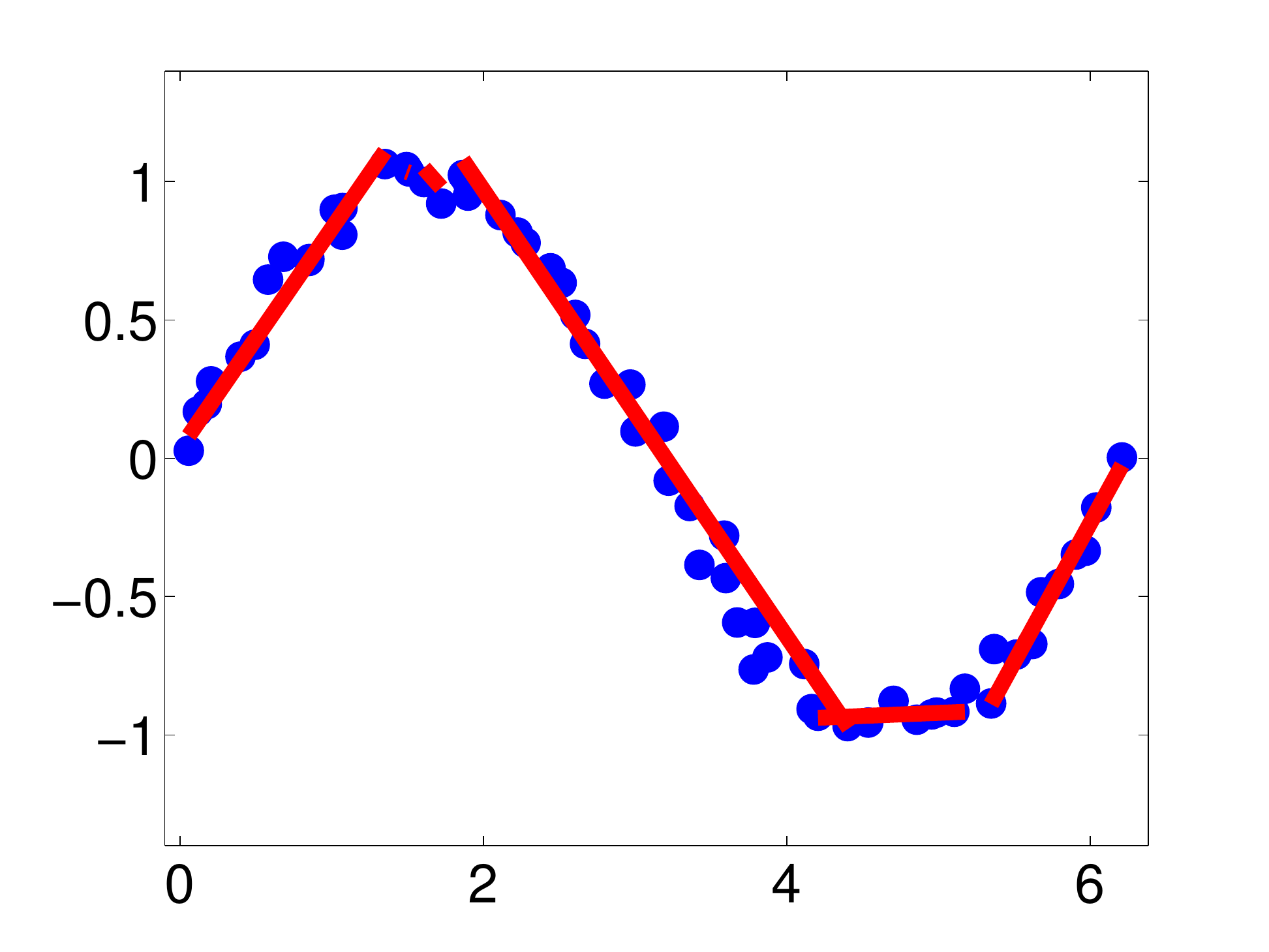}
\vspace{-1mm}
\caption{$\lambda = 0.01 \, \lambda_{\max,2}$}
\end{subfigure}
\caption{\small{Finding representative models for noisy data points on a nonlinear manifold. For each data point and its $K = 4$ nearest neighbors, we learn a one-dimensional affine model fitting the data. Once all models are learned, we compute the dissimilarity between each model and a data point by the absolute value of the representation error. Representative models found by our proposed optimization program in \eqref{eq:tracerow-matrix-1} for several values of $\lambda$, with $\lambda_{\max,2}$ defined in \eqref{eq:max-lambda}, are shown by red lines. Notice that as we decrease $\lambda$, we obtain a larger number of representative models, which more accurately approximate the nonlinear manifold.}}
\label{fig:RepModel_p2}
\end{figure*}
\begin{figure*}[t!]
\centering
\begin{subfigure}[b]{0.23\textwidth}
\includegraphics[width=\textwidth, trim = 56 23 46 40 , clip]{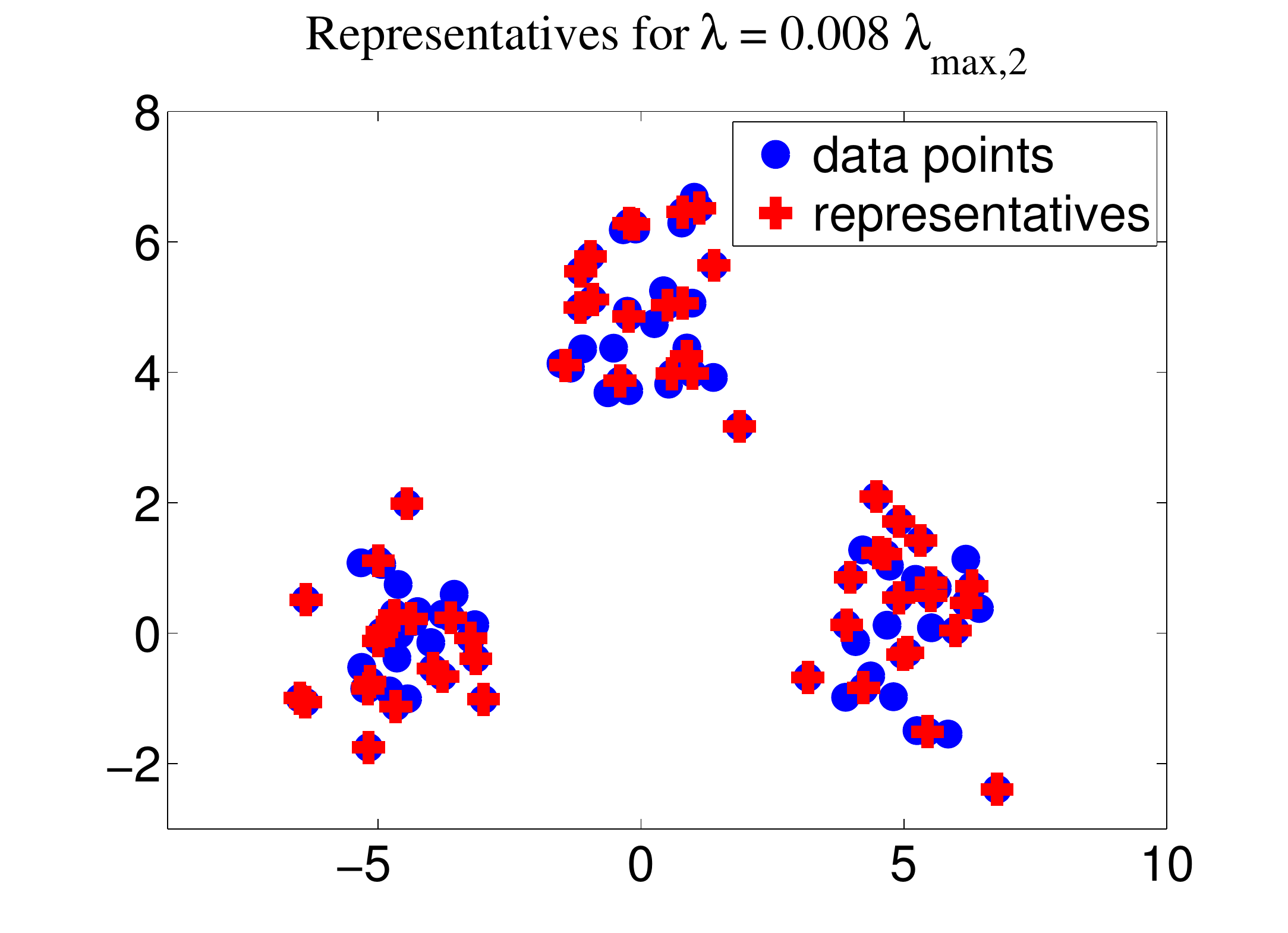}\\
\vspace{-1mm}
\includegraphics[width=\textwidth, trim = 33 20 34 40 , clip]{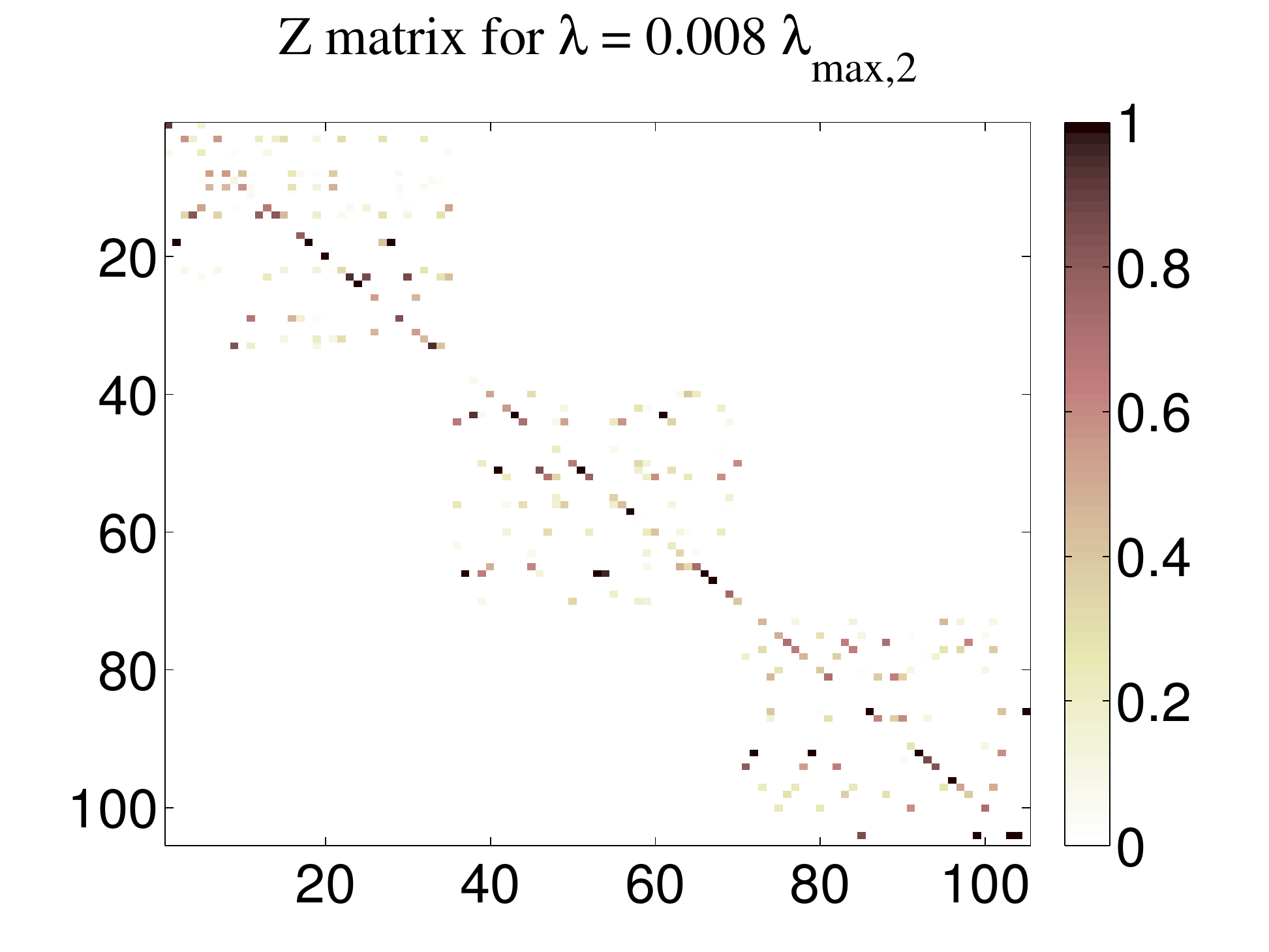}
\caption{$\lambda = 0.008 \, \lambda_{\max,2}$}
\end{subfigure}
\hspace{1mm}
\begin{subfigure}[b]{0.23\textwidth}
\includegraphics[width=\textwidth, trim = 56 23 46 40 , clip]{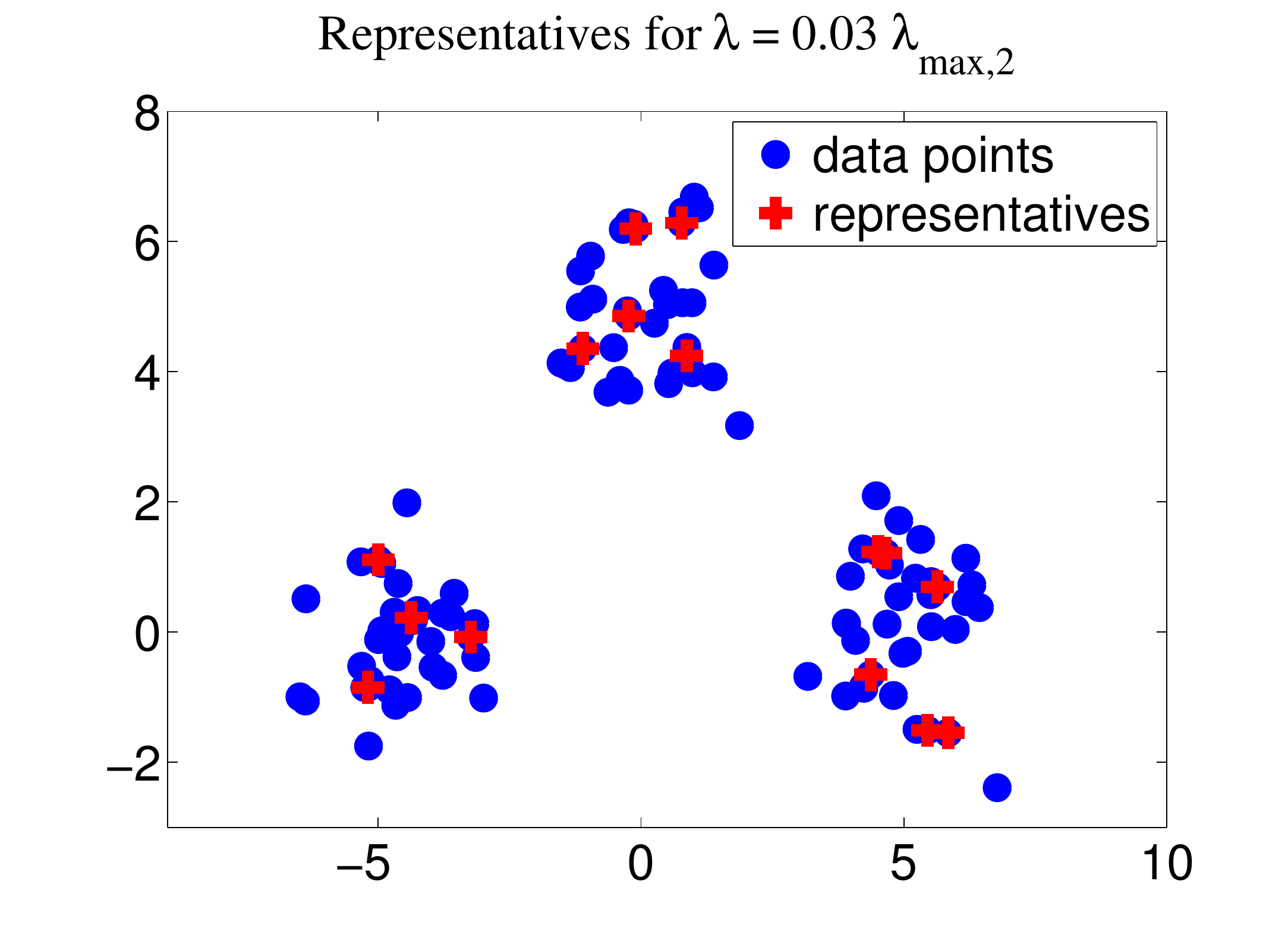}\\
\vspace{-1mm}
\includegraphics[width=\textwidth, trim = 33 20 34 40 , clip]{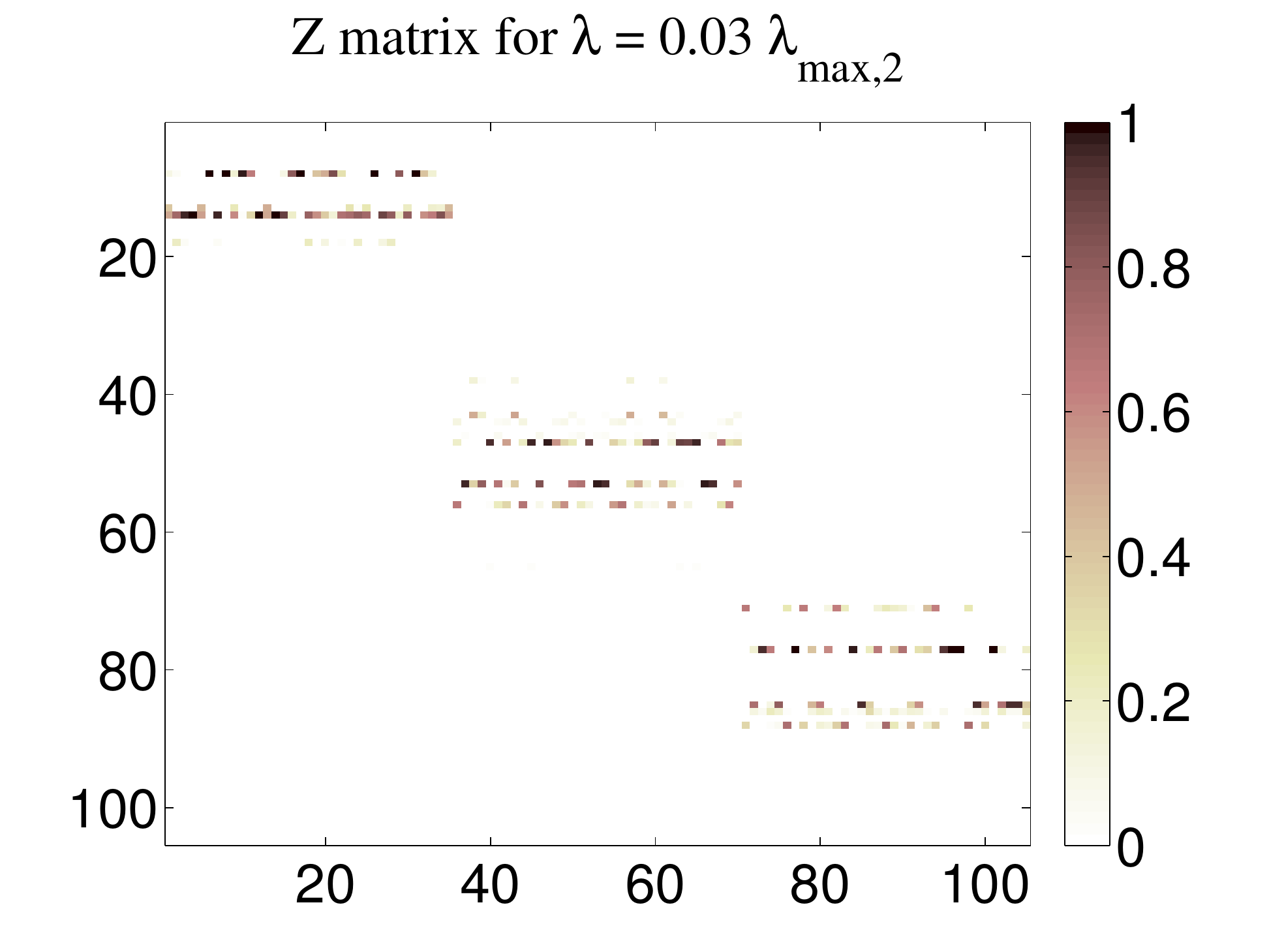}
\caption{$\lambda = 0.03 \, \lambda_{\max,2}$}
\end{subfigure}
\hspace{1mm}
\begin{subfigure}[b]{0.23\textwidth}
\includegraphics[width=\textwidth, trim = 56 23 46 40 , clip]{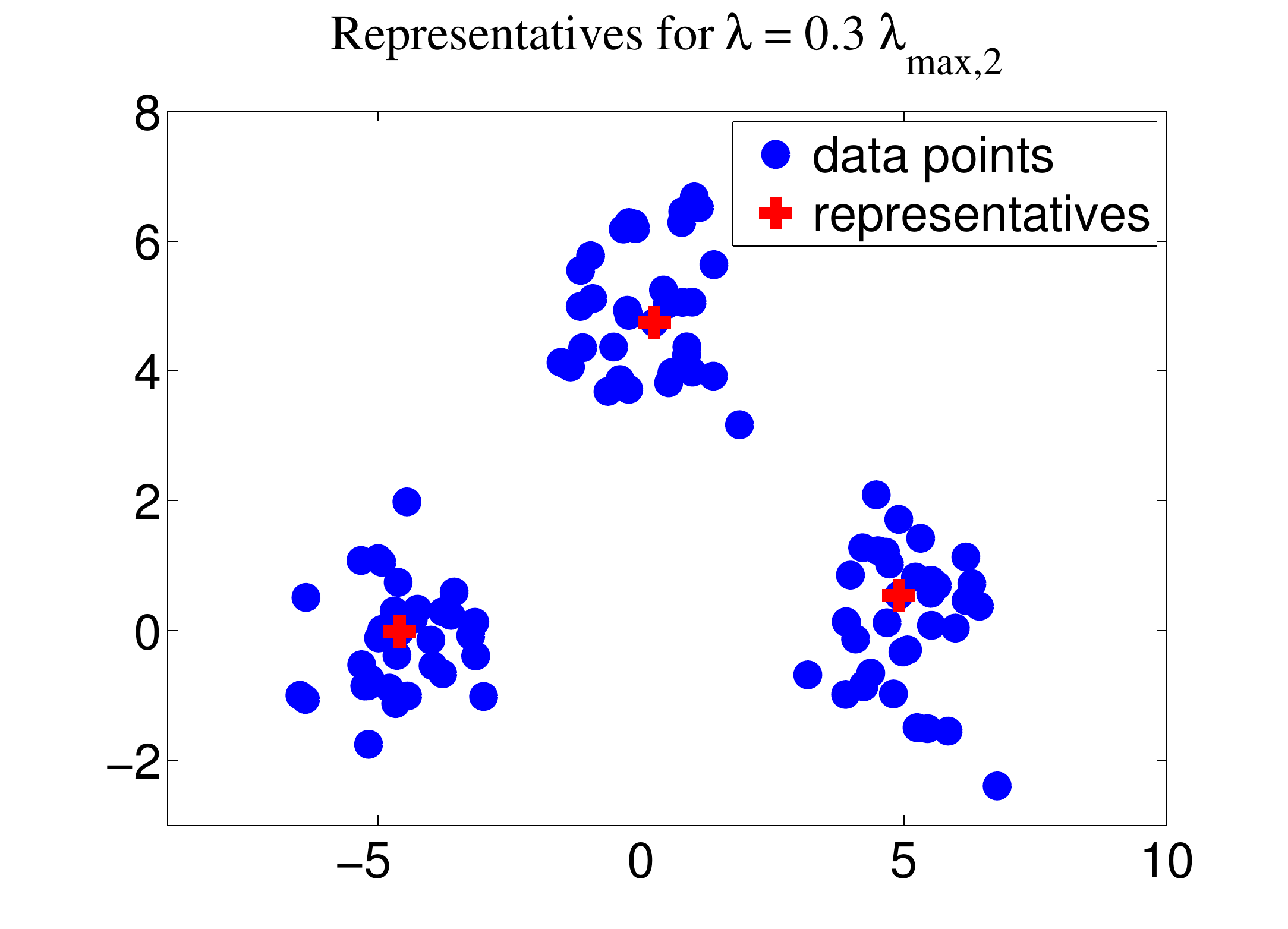}\\
\vspace{-1mm}
\includegraphics[width=\textwidth, trim = 33 20 34 40 , clip]{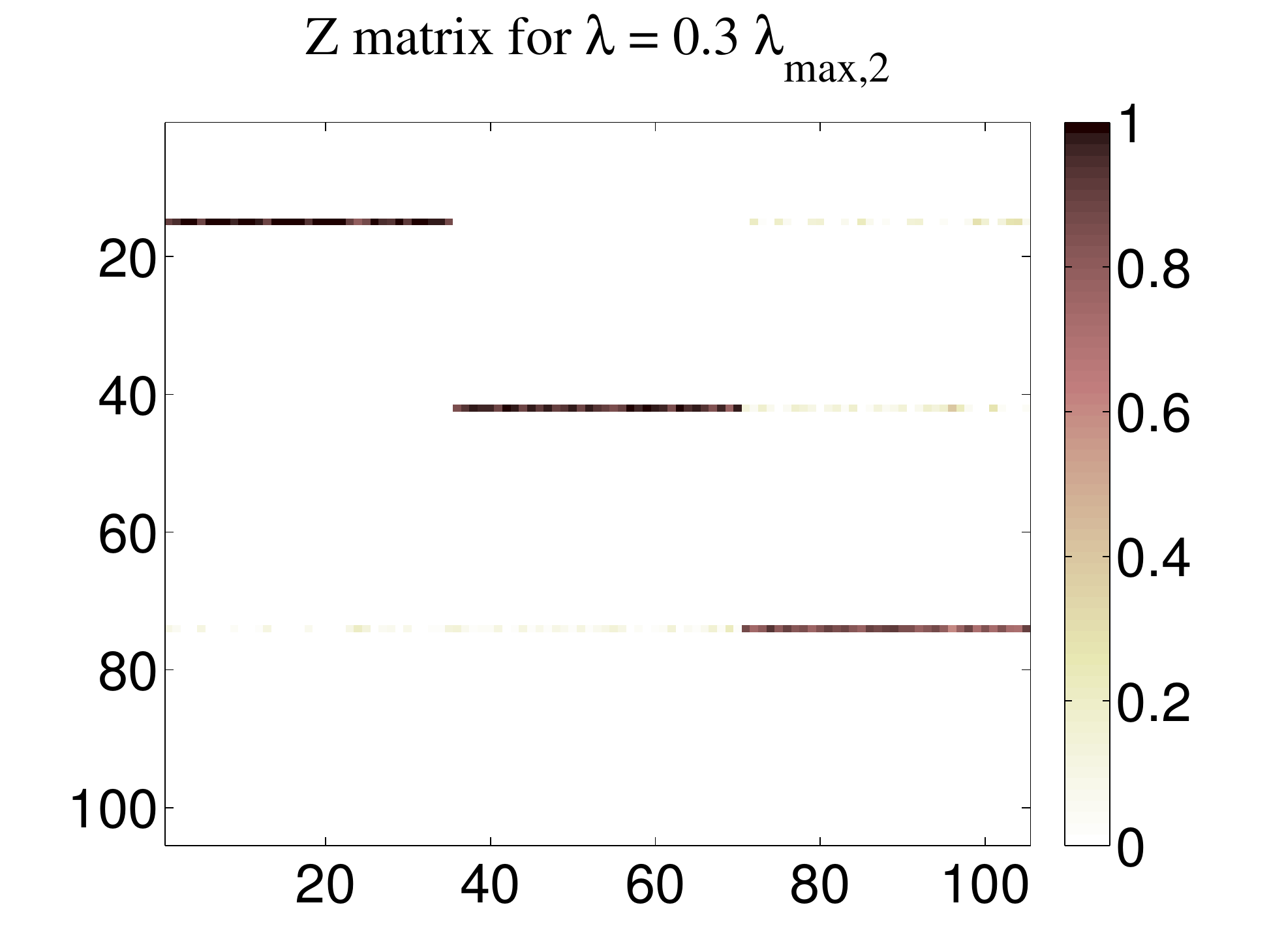}
\caption{$\lambda = 0.3 \, \lambda_{\max,2}$}
\end{subfigure}
\hspace{1mm}
\begin{subfigure}[b]{0.23\textwidth}
\includegraphics[width=\textwidth, trim = 56 23 46 40 , clip]{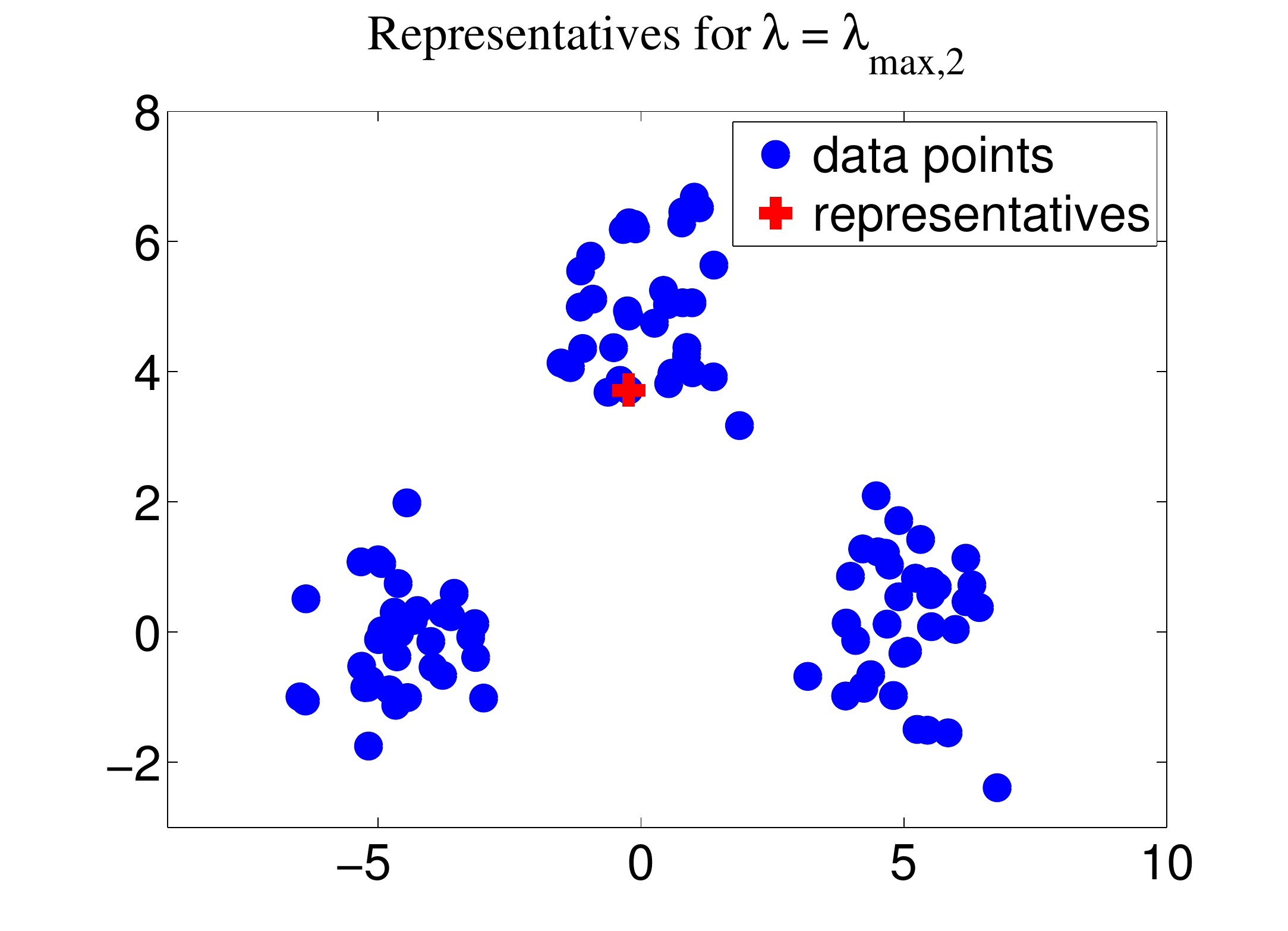}\\
\vspace{-1mm}
\includegraphics[width=\textwidth, trim = 33 20 34 40 , clip]{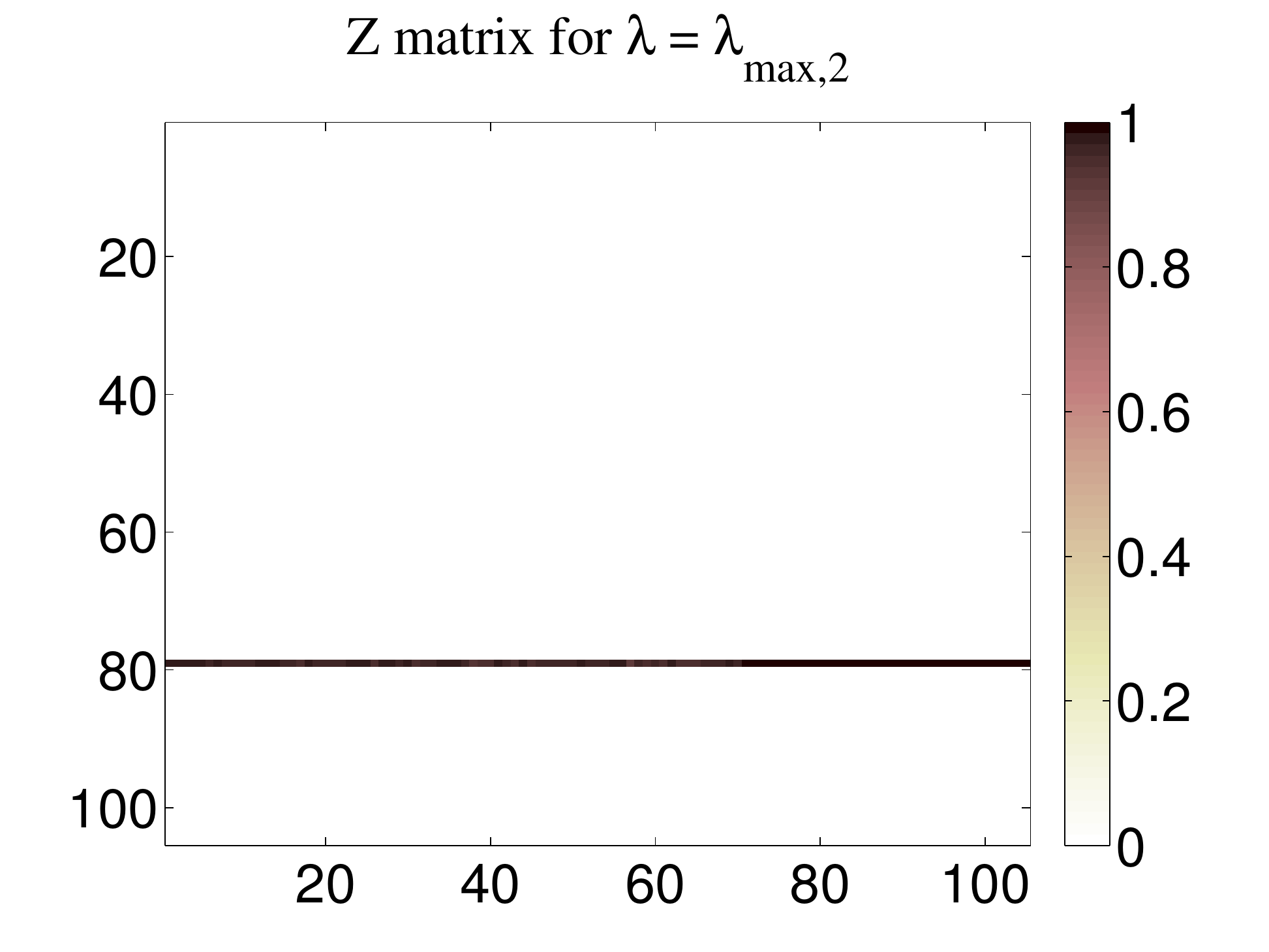}
\caption{$\lambda = \lambda_{\max,2}$}
\end{subfigure}
\caption{\small{Top: Data points (blue circles) drawn from a mixture of three Gaussians and the representatives (red pluses) found by our proposed optimization program in \eqref{eq:tracerow-matrix-1} for several values of $\lambda$, with $\lambda_{\max,2}$ defined in \eqref{eq:max-lambda}. Dissimilarity is chosen to be the Euclidean distance between each pair of data points. As we increase $\lambda$, the number of representatives decreases. Bottom: the matrix $\Z$ obtained by our proposed optimization program in \eqref{eq:tracerow-matrix-1} for several values of $\lambda$. The nonzero rows of $\Z$ indicate indices of the representatives. In addition, entries of $\Z$ provide information about the association probability of each data point with each representative.}}
\label{fig:2G-PZ_p2}
\end{figure*}
\begin{table}[t!]
\caption{\footnotesize Errors ($\%$) of different algorithms, computed via \eqref{eq:nnerror}, as a function of the fraction of selected samples from each class ($\eta$) on the 15~Scene~Categories dataset using $\chi^2$ distances.} \centering
\begin{small}
\begin{tabular}{|@{\;\,}c@{\;\,}|@{\;\,}c@{\;\,}|@{\;\,}c@{\;\,}|@{\;\,}c@{\;\,}|@{\;\,}c@{\;\,}|}
\hline
Algorithm & \;\; Rand \;\; & \, Kmedoids \, & \;\;\; AP\;\;\; & \;\;\; DS3 \;\;\;  \\
\hline
\hline
$\eta = 0.05$ &    $22.12$ & $14.42$ & $\textcolor{black}{\textbf{11.59}}$ & $12.04$ \\
\hline
\hline
$\eta = 0.10$ &  $15.54$ & $11.30$ & $7.91$ & $\textcolor{black}{\textbf{5.69}}$\\
\hline
\hline
$\eta = 0.20$ &  $11.97$ & $12.19$ & $6.01$ & $\textcolor{black}{\textbf{3.35}}$\\
\hline
\hline
$\eta = 0.35$ &  $7.18$ & $7.51$ & $6.46$ & $\textcolor{black}{\textbf{2.90}}$\\
\hline
\end{tabular}
\end{small}
\label{tab:15Scene-errors-chi}
\vspace{0mm}
\end{table}
\begin{table}[t!]
\caption{\footnotesize Errors ($\%$) of different algorithms, computed via \eqref{eq:nnerror}, as a function of the fraction of selected samples from each class ($\eta$) on the 15~Scene~Categories dataset using Euclidean distances.} \centering
\begin{small}
\begin{tabular}{|@{\;\,}c@{\;\,}|@{\;\,}c@{\;\,}|@{\;\,}c@{\;\,}|@{\;\,}c@{\;\,}|@{\;\,}c@{\;\,}|}
\hline
Algorithm & \;\; Rand \;\; & Kmedoids & \;\;\; AP \;\;\; & \, DS3 \, \\
\hline
\hline
$\eta = 0.05$ &    $15.61$ & $10.48$ & $\textcolor{black}{\textbf{7.58}}$ & $8.03$\\
\hline
\hline
$\eta = 0.10$ &  $11.82$ & $9.70$ & $7.07$ & $\textcolor{black}{\textbf{6.58}}$\\
\hline
\hline
$\eta = 0.20$ &  $9.92$ & $7.80$ & $6.13$ & $\textcolor{black}{\textbf{5.58}}$\\
\hline
\hline
$\eta = 0.35$ &  $7.69$ & $6.47$ & $5.24$ & $\textcolor{black}{\textbf{3.24}}$\\
\hline
\end{tabular}
\end{small}
\label{tab:15Scene-errors-euc}
\vspace{-2mm}
\end{table}
\section*{Classification using Representatives}
Table \ref{tab:15Scene-errors-chi} and Table \ref{tab:15Scene-errors-euc} show the NN classification error of different algorithms on the dataset as we change the fraction of representatives, $\eta$, selected from each class for $\chi^2$ distance and Euclidean distance dissimilarities, respectively. More specifically, after selecting $\eta$ fraction of training samples in each class using each algorithm, we compute the average NN classification accuracy on test samples, denoted by $\text{accuracy}(\eta)$, and report 
\begin{equation}
\label{eq:nnerror}
\text{err}(\eta) = \text{accuracy}(1) - \text{accuracy}(\eta),
\end{equation}
where $\text{accuracy}(1)$ is the NN classification accuracy using all training samples in each class.
As the results show, increasing the value of $\eta$ results in obtaining more representatives from each class, hence improving the classification results as expected. Rand performs worse than other methods, followed by Kmedoids, which suffers from dependence on a good initialization. On the other hand, DS3, in general, performs better than other methods, including AP. This comes from the fact that AP relies on a message passing algorithm, which results in an approximate solution when the moral graph \cite{Koller:book09} of pairwise relationships is not a tree, including our problem. Notice also that by selecting only $35\%$ of the training samples in each class, the performance of DS3 is quite close to the case of using all training samples. More specifically, using $\chi^2$ distances, the performance of DS3 is $2.90\%$ lower than the performance using all samples, while using Euclidean distances the difference is $3.24\%$. Also, it is important to notice that the performances of all methods depend on the choice of  dissimilarities. More specifically, good dissimilarities should capture the distribution of the data in a way that points from the same group have smaller dissimilarities than points in different groups. In fact, in the experiment above, using the $\chi^2$ dissimilarity results in improving the classification performance of all algorithms by about $16\%$.

\end{document}